%% file: main.tex
\documentclass{article} %
\usepackage{iclr2027_conference,times}
\input{math_commands.tex}

\usepackage{hyperref}
\hypersetup{hidelinks,pdftitle={Rewriting Scaling Laws for Synthetic Data with Compute and Data Constraints},pdfauthor={Nicholas Roberts, Ahmed Ahmed, Suhas Kotha, Percy Liang, Yu Wang, Alex Renda, Nitya Mani},pdfsubject={Compute-optimal mixtures of natural and synthetically rewritten pretraining data},pdfkeywords={scaling laws, synthetic data, pretraining, data rewriting}}
\usepackage{url}
\usepackage{graphicx}
\usepackage{tcolorbox}
\usepackage{enumitem}
\usepackage{booktabs}
\usepackage{multirow}
\usepackage{xcolor}
\usepackage{listings}
\usepackage{needspace}
\usepackage{marginalia}

\newcommand{\Rhat}{\widehat{R}^*}  %

\title{Rewriting Scaling Laws for Synthetic Data with Compute and Data Constraints}

\author{%
\makebox[\textwidth][c]{%
\begin{tabular}{cccc}
\textbf{Nicholas Roberts}\thanks{Equal contribution, order determined by coin flip. Work done during an internship at Jane Street.\\Correspondence to \emph{nick11roberts@princeton.edu} and \emph{ahmedah@cs.stanford.edu}} &
\textbf{Ahmed Ahmed}\footnotemark[1] &
\textbf{Suhas Kotha} &
\textbf{Percy Liang} \\
\normalfont Princeton University & \normalfont Stanford University & \normalfont Stanford University & \normalfont Stanford University
\end{tabular}}\\[10pt]
\makebox[\textwidth][c]{%
\begin{tabular}{ccc}
\textbf{Yu Wang} & \textbf{Alex Renda} & \textbf{Nitya Mani} \\
\normalfont Jane Street & \normalfont Jane Street & \normalfont Jane Street
\end{tabular}}%
}

\iclrfinalcopy %
\begin{document}

\maketitle
\lhead{Preprint}

\input{sections/abstract}

\input{sections/introduction}

\input{sections/methodology}

\input{sections/experiments}

\input{sections/related}
\input{sections/conclusion}

\input{sections/ai_use}

\input{sections/acks}

\bibliography{ref}
\bibliographystyle{iclr2027_conference}

\newpage
\appendix
\input{sections/appendix}

\end{document}

%% file: math_commands.tex
\usepackage{amsmath,amsfonts,bm}
\usepackage{mathtools}

\def\eqref#1{equation~\ref{#1}}

\def\1{\bm{1}}

\DeclareMathAlphabet{\mathsfit}{\encodingdefault}{\sfdefault}{m}{sl}
\SetMathAlphabet{\mathsfit}{bold}{\encodingdefault}{\sfdefault}{bx}{n}

\DeclareMathOperator*{\argmin}{arg\,min}

%% file: sections/abstract.tex
\begin{abstract}

Synthetic rewrites, where a language model rewrites a natural document e.g. into a target style, are now a standard ingredient of pretraining corpora. 
However, critical aspects of their use remain poorly understood, such as how much compute should be spent rewriting training data and whether a model trained on rewrites can outperform the rewriter. 
We study these two questions by proposing a rewriting scaling law that predicts loss from model size, training tokens, the fraction of training tokens that are rewrites, and the number of epochs over the natural data, charging the compute spent generating rewrites. 
We fit this scaling law in two settings. 
In the compute-limited setting, natural data is plentiful and rewrites come from a third-party rewriter trained with more compute than our largest run. 
We find that the estimated optimal rewrite fraction increases with compute, from under $10\%$ at $3\times10^{18}$ FLOPs to $32\%$ at $10^{22}$ FLOPs. 
Our scaling forecast extrapolates $33\times$ in FLOPs to train a 22B-parameter model, whose final loss is within $1.1\%$ of our prediction. 
This model achieves a better BPB than its rewriter on several web domains at one-tenth the rewriter's training compute. 
In the data-limited setting, we assume we only have a fixed seed set of natural data and that our rewriter is a model of the same size as the target, pretrained on that seed set. 
Again, we find the optimal rewrite fraction increases with the compute budget. 
We also demonstrate that a model trained on rewrites of its own pretraining data, generated by a rewriter of the same size, beats the best run on natural data alone at every model size up to 3B. 
Our results show that synthetic rewriting can be studied through the scaling law lens: that optimal methodological choices are predictable and scaling does not necessarily break down beyond the rewriter's capabilities.

\end{abstract}

%% file: sections/introduction.tex
\section{Introduction}
Large language models have a host of remarkable capabilities, many of which improve smoothly during pretraining and grow as model parameters and training tokens increase, in a manner that is predictable via scaling laws \citep{kaplan2020scalinglawsneurallanguage,hoffmann2022trainingcomputeoptimallargelanguage}. 
Early language models were pretrained using corpora of text that were primarily comprised of \textit{natural data}, text generated either manually by a human or generated programmatically. 
Common sources of such tokens included web pages collected from internet crawls, GitHub repositories, and academic publications~\citep{gao2020pile800gbdatasetdiverse}. 
However, the amount of compute available for training has grown much faster than the amount of high quality, natural pretraining data \citep{villalobos2022runoutdatalimits, kim2025pretraininginfinitecompute}.

These constraints on the availability of large-scale natural corpora of text have motivated using \textit{synthetic rewrites}, documents generated by prompting a language model with a natural document and an instruction to write a different version of the document. 
Common instructions for synthetic rewrites include
rewriting in the manner of a Wikipedia article~\citep{maini2024rephrasingwebrecipecompute},
converting math web pages into synthetic math textbook-style exposition~\citep{karimimahabadi2025nemotronccmath},
synthesizing or summarizing lengthy source material~\citep{maini2025beyondweblessonsscalingsynthetic},
and 
transforming a lower quality source into a higher quality output using the model's knowledge~\citep{su2024nemotroncctransformingcommoncrawl,nguyen2025recyclingwebmethodenhance}.

Synthetic rewrites of natural documents have become a standard ingredient in many modern pretraining corpora \citep{maini2025beyondweblessonsscalingsynthetic,su2024nemotroncctransformingcommoncrawl,nguyen2025recyclingwebmethodenhance,niklaus2026synthesizehighqualitypretrainingdata}. 
However, their use does have several drawbacks.
First, synthetic rewrites are costly to generate: for a given model size, rewriting a given document involves a significant fraction of the compute of training on that document.
Second, synthetic rewrites are necessarily generated by a model with a given amount of complexity and capability, and it is unclear the extent to which that bounds the capabilities of a model trained on these synthetic rewrites. 

In this paper we develop a \emph{rewriting scaling law} that takes as input the \emph{rewrite fraction}, the fraction of training tokens that are synthetic rewrites, and predicts the expected loss of a model trained at that rewrite fraction.
When extrapolating this scaling law to predict the loss-minimizing fraction at a given compute budget, we account for the \emph{cost-adjusted training FLOPs}, the amount of FLOPs available for training after accounting for the cost of generating synthetic rewrites.

With this scaling law, we investigate two specific research questions:
\begin{enumerate}[label=\textbf{RQ\arabic*}, leftmargin=*, itemsep=1pt, topsep=3pt, parsep=0pt]
    \item %
    \emph{Does our scaling law accurately predict the optimal rewrite fraction and its achieved loss?}
    \item %
    \emph{Should we include rewrites when targeting capabilities that surpass those of the rewriter?}
\end{enumerate}

\begin{figure}[t!]
\begin{center}
\includegraphics[width=0.49\textwidth]{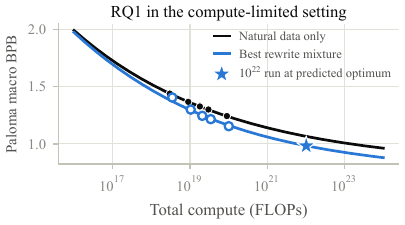}%
\hfill
\includegraphics[width=0.49\textwidth]{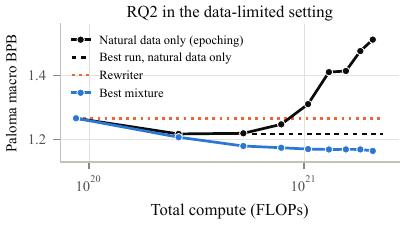}
\end{center}
\caption{
    \emph{(Left; compute-limited) Effective rewriting recipes are predictable via scaling laws:} best possible Paloma BPB (lower is better; allowing third-party rewrites) vs FLOPs (accounting for rewrite compute costs) as compared to using natural data only. Unfilled blue points are imputed via linear interpolations of the loss of the closest rewrite fractions that we evaluate, whereas solid points represent observed training runs. \emph{(Right; data-limited) The trained model surpasses its rewriter:} final  Paloma BPB vs FLOP for a $3$B target model trained using only natural data or synthetic rewrites coming from a fixed $5$B-token natural data corpus. Using natural data only, which necessitates epoching, results in worse loss after $3$ epochs. The best proportion of rewrites from a bootstrapped rewriter improves over epoching on the natural seed set alone at every budget. 
}
\label{fig:intro}
\end{figure}
We study these questions in distinct settings characteristic of language model training paradigms in which the primary constraint is either compute~\citep{hoffmann2022trainingcomputeoptimallargelanguage} or data~\citep{kim2025pretraininginfinitecompute}.

In the first setting, which we term \textit{compute-limited}, we focus on a scenario where natural data is unbounded. 
We assume an \emph{external rewriter}, a rewriting model that is trained with more total compute than our largest training run. 
This setting exemplifies the small-scale scientific workflow.

In the second setting, which we term the \textit{data-limited} setting, we focus on a scenario in which natural data is bounded. 
We also assume that we only have access to a \emph{bootstrapped rewriter} which is trained on a subset of the available data and for fewer FLOPs than a given candidate training run. 
This setting exemplifies the frontier-scale workflow in which the objective is to train a model more capable than any previous model.

We find the following results:
\begin{itemize}[leftmargin=*, itemsep=1pt, topsep=3pt, parsep=0pt]
    \item 
    \emph{Our scaling law accurately predicts the optimal rewrite fraction and achieved loss at large budgets.}
    In the compute-limited setting, we fit our rewriting scaling law 
    at low budgets
    and extrapolate to a set of $10^{22}$-FLOP runs.
    We find that the predicted optimal rewrite fraction results in the lowest held-out loss among tested models at the $10^{22}$-FLOP scale, achieving loss within $1.1\%$ of the predicted loss. 
    In the data-limited setting, a scaling law fit to low budgets predicts a rewrite fraction at large held-out budgets whose loss is within $1.14\%$ of that of the best observed rewrite fraction.
    In both settings, the optimal rewrite fractions increase with scale and are task-specific, differing between math, code, question-answering, and general~web~text. 
    \item \emph{Models can achieve and exceed their rewriter model's performance.} 
    In the compute-limited setting, pretraining a model for $10^{22}$ FLOPs---where $32\%$ of its training data is rewritten---outperforms the external rewriter (SmolLM2-1.7B-Instruct) on 10 out of 16 general web text sources  (using roughly one-tenth of the rewriter's training compute). 
    In the data-limited setting, where the bootstrapped rewriter is always trained with fewer FLOPs than the target model, models trained on a mixture containing rewrites consistently outperform the rewriter. 
\end{itemize}

Our results give evidence that synthetic rewriting can be studied through the scaling law lens, showing both that optimal methodological choices are predictable and that scaling does not necessarily break down beyond the capabilities of the rewriter.

%% file: sections/methodology.tex
\section{Methodology}
\label{sec:scaling_model}

In this section we first present the rewriting scaling law we use to predict the expected loss of a model trained with a given amount of synthetic rewrites in the training set.
We then describe how we practically fit and use this scaling law to predict optimal rewrite fractions, including how we account for the cost of compute to generate the synthetic rewrites.

\subsection{A Rewriting Scaling Law}

In a given experiment in both the compute- and data-limited settings, we pretrain an $N$-parameter \emph{target model} on a \emph{corpus} with $D$ total tokens composed of a combination of \emph{natural tokens} and \emph{synthetic rewrites} thereof.
The set of natural tokens in a given corpus is drawn from a \emph{seed set} of $T$ natural tokens, potentially with replacement (epoching).
The synthetic rewrite tokens are rewrites of the natural tokens selected for the corpus, without epoching.
We define the \emph{rewrite fraction} $R \in [0, 1]$ for a given corpus as the fraction of tokens in the corpus that~are~synthetic~rewrites.

Our \emph{rewriting scaling law} predicts that an $N$-parameter target model trained on a $D$-token corpus with seed set size $T$ and rewrite fraction $R$ has held-out loss $L$ as follows:
\begin{equation}
    L(N, D, R, T) = \underbrace{\frac{A}{N^{\alpha}} + \frac{B}{D^{\beta}}}_{\text{Chinchilla terms}}
    + \underbrace{E(R)}_{\substack{\text{rewrite-fraction-dependent}\\ \text{irreducible loss}}}
    \hspace{0.6em} + \hspace{1.4em} \underbrace{F(R, D, T)}_{\mathclap{\text{repetition penalty}}}.
\label{eq:rewriting_scaling_law}
\end{equation}

The terms $A/N^{\alpha}+B/D^{\beta}$ are the same as in the classic Chinchilla-style scaling law of~\citet{hoffmann2022trainingcomputeoptimallargelanguage}, which model the loss contributions of using a certain parameter count and dataset size. 

The term $E(R)$ represents the irreducible loss as a function of the rewrite fraction $R$.
Appendix~\ref{scaling_law_details} defines this as an approximation of the KL divergence between the training and evaluation distributions. 
The specific functional form for this \emph{rewrite-fraction-dependent irreducible loss} is given as $E(R) = E_0 - \lambda_n \log\!\left(1-(1-\rho_n)R\right) -\lambda_s \log\!\left(\rho_s+(1-\rho_s)R\right)$, where $\lambda_n$, $\lambda_s$, $\rho_n$, $\rho_s$, and $E_0$ are all fitted parameters. Additional details can be found in Appendix~\ref{scaling_law_details}. 

The term $F(R, D, T)$ represents the cost of epoching natural data multiple times, and is drawn from Equation (7) in~\citet{lovelace2026prescriptivescalinglawsdata}.
This \emph{repetition penalty} term is defined as $F(R, D, T) = P\,(1 - R)^{s}\,\bigl((1-R)D/T - 1\bigr)_+,$
where $P$ and $s$ are fitted parameters. 
The repetition penalty is only nonzero when the natural tokens are epoched, so it appears only in the data-limited setting.

\subsection{Predicting Optimal Rewrite Fractions at a Compute Budget}

Once we fit our rewriting scaling law, our goal is to choose the rewrite fraction that minimizes loss for a fixed compute budget. 
Since synthetic rewrites are not free, the compute budget must account for both the cost of training the target model and the cost of generating synthetic rewrites. 

We approximate the FLOPs for training a target model with $N$ parameters on a corpus of size $D$ with rewrite fraction $R$ using synthetic data generated by a rewriter with $N_r$ parameters with:
\begin{equation}
    C_{\mathrm{total}} :=
    \underbrace{6ND}_{\text{target-model training}}
    + \underbrace{2N_rDR}_{\text{source prefill}}
    + \underbrace{2N_rDR}_{\text{rewrite decoding}}.     
\label{eq:total_compute}
\end{equation}

This $C_{\mathrm{total}}$ approximation makes several assumptions.
First, we use the standard assumptions that forward and backward passes are respectively $2N$ and $4N$ FLOPs per token~\citep{kaplan2020scalinglawsneurallanguage}, giving $6ND$ target-model training FLOPs.
Second, we assume that the length of a synthetic rewrite is the same as the length of the input document, and that these lengths are large relative to other elements of the prompt; this gives $2N_rDR$ FLOPs for forward passes (prefill) on the natural tokens used for rewriting, and an additional $2N_rDR$ FLOPs for forward passes (decode) when generating the synthetic rewrites.\footnote{For FLOPs to be a reasonable proxy for the cost of generating synthetic rewrites, we must also assume that generation is FLOP-bound and not memory-bandwidth-bound; this often holds with sufficiently large batch sizes and short context lengths~\citep{pope2023efficiently}, which is tractable when generating synthetic data.}
We also assume no amortization of the cost of generating rewrites across multiple different target model training runs.

The \emph{optimal rewrite fraction} $R^*$ minimizes the true loss at a given compute budget. We denote our estimate of it as $\Rhat$, which is the minimizer of our fitted scaling law:
\begin{equation}
    \Rhat = \argmin_{R \in [0,1]} \; \min_{N,\, D} \; L(N, D, R, T)
    \quad \text{s.t.} \quad
    6ND + 4N_rRD = C_{\mathrm{total}}.
\label{eq:optimal_rewrite_fraction}
\end{equation}

%% file: sections/experiments.tex
\section{Experiments}
We study our research questions in two different settings.

First in the \textit{compute-limited} setting, we assume a fixed FLOP budget, more natural data in our seed set than our largest target model requires ($T \gg D$), and an \emph{external rewriter} trained for more FLOPs than any member of our scaling ladder on a potentially different seed set of data.
This setting is motivated by the small-scale scientific workflow. 

Next in the \textit{data-limited} setting, we assume a fixed FLOP budget, a limited seed set ($T \ll D$), and for each target model a corresponding \emph{bootstrapped rewriter} trained for fewer FLOPs than that target model on the same seed set.
This setting is motivated by the frontier-scale workflow in which the objective is to train a model more capable than any previous model.

Consistent across both settings we find:
\textbf{(RQ1)} scaling law based extrapolations well predict the optimal fraction of rewrites and the corresponding achieved loss; and
\textbf{(RQ2)}, in both the compute-limited and data-limited settings, the scaling law predicts and we empirically confirm that it is optimal to include rewrites (i.e., $\Rhat > 0$) for target models that surpass capabilities of the rewriter.

\subsection{Experimental Setup}
\label{sec:recipe}

\paragraph{Architecture and training.}
All runs use Qwen3 dense models~\citep{yang2025qwen3technicalreport} with the Llama~3 tokenizer~\citep{grattafiori2024llama3herdmodels}, trained with Adam and Hyperball~\citep{wen2026fantasticpretrainingoptimizersii} under a linear warm-up and linear-decay schedule.
Appendix~\ref{model_arch} gives the width-to-depth rule and Appendix~\ref{completedp} the hyperparameter tuning and transfer recipe.

\paragraph{Data.}
In the compute-limited setting, we train on FineWeb-Edu~\citep{penedo2024finewebdatasetsdecantingweb} with FinePhrase rewrites~\citep{niklaus2026synthesizehighqualitypretrainingdata}, which are FineWeb-Edu rewrites from SmolLM2-1.7B~\citep{benallal2025smollm2smolgoesbig}. 
In the data-limited setting, we train on a $5$B-token seed set from FineWeb-Edu and rewrites of that seed set using a model that we train ourselves, also on the seed set. %
We retain only documents from web-archive (WARC) files dated before 2023 to reduce LM-generated text in the natural corpus (Appendix~\ref{fwedu_data}).
For mixtures with rewrite fraction $0<R<1$ we interleave natural and rewritten documents in proportion to $R$ throughout training (Appendix~\ref{mixture_construction}).

\paragraph{Rewriting protocol.}
Following FinePhrase~\citep{niklaus2026synthesizehighqualitypretrainingdata}, every rewrite converts one natural document into a math exposition, an FAQ, a table, or a tutorial. 

\paragraph{Evaluation.}
Our evaluations include Paloma~\citep{magnusson2024paloma} and the set of aggregations in the Olmo 3 base easy evaluation suite~\citep{olmo2025olmo3}: QA (ARC, HellaSwag, and MMLU), Math (Minerva MATH, with GSM8K-Platinum and TheoremQA added), and Code (HumanEval, MBPP, and multilingual MBPP) \citep{clark2018thinksolvedquestionanswering,zellers2019hellaswag,hendrycks2021measuringmassivemultitask,cobbe2021trainingverifierssolvemath,vendrow2025largelanguagemodelbenchmarks,chen2023theoremqa,hendrycks2021measuringmathematicalproblem,lewkowycz2022solvingquantitativereasoning,chen2021evaluatinglargelanguagemodels,austin2021programsynthesislargelanguage,cassano2023multiple}.
We list the suite membership and aggregation for all evaluations in Appendix~\ref{evals}. 

On all evaluations, we report bits-per-byte (BPB) at the trained $4096$-token context.
For Paloma, we report the macro-averaged BPB, BPB averaged within each Paloma domain then averaged across domains (equal-weighting all domains).
For the Olmo 3 suites we report gold-answer BPB.

\subsection{Compute-Limited Experiments}
\label{sec:compute_eff_exp}

In the \textit{compute-limited} setting, we assume a fixed FLOP budget, more natural data in our seed set than our largest target model requires ($T \gg D$), and an \emph{external rewriter} trained for more FLOPs than any member of our scaling ladder on a potentially different seed set of data.

\paragraph{Scaling ladder methodology.}

We train \textit{IsoFLOP ladders}---sets of models with varying size and token counts, each trained at one of a range of fixed training budgets---on FineWeb-Edu at eleven rewrite fractions $R\in\{0,0.1,\ldots,1.0\}$. 
SmolLM2-1.7B \citep{benallal2025smollm2smolgoesbig} generates rewrites, which we draw from FinePhrase \citep{niklaus2026synthesizehighqualitypretrainingdata}.
We train target models with training budgets spanning $3\times10^{18}$ to $9\times10^{19}$ FLOPs at all eleven rewrite fractions, and natural-data-only ($R=0$) ladders at $1.8\times10^{20}$ and $3\times10^{20}$ FLOPs.
We fit our scaling laws with these target models against performance on each of the QA, Math, Code, and Paloma evals.
We validate our rewriting scaling law's predictions with 
validation runs at $10^{21}$ FLOPs across all eleven rewrite fractions and a set of $10^{22}$ runs including a run at the predicted $\Rhat$ optimum (and corresponding $N$ and $D$ optima) for the Paloma eval.

\paragraph{Rewriting methodology.}
In the compute-limited setting, we use the released FinePhrase SmolLM2-1.7B rewrites.
SmolLM2-1.7B was trained for roughly $10^{23}$ FLOPs.

\begin{figure}[t]
\begin{minipage}[c]{0.58\textwidth}
\centering
\includegraphics[width=\linewidth]{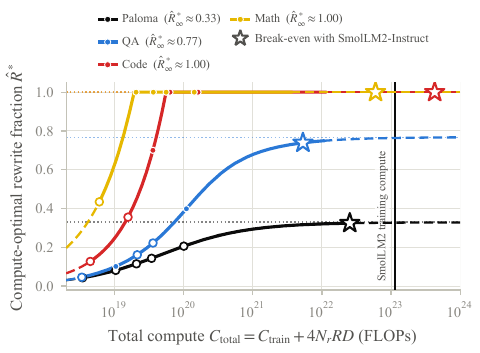}
\end{minipage}\hfill
\begin{minipage}[c]{0.39\textwidth}
\raggedright
\caption{\textbf{The estimated compute-optimal rewrite fraction increases with compute on every evaluation.} 
The curves show $\Rhat$ across evaluations and FLOP budgets.
Solid lines indicate that the compute budget is within the evaluated ladder, while dashed lines are extrapolations.
Points mark the fit optimum at each ladder budget; filled points land on a tested rewrite fraction and unfilled points fall between two. 
Stars mark predicted break-even with SmolLM2-Instruct. 
The vertical line marks the training FLOPs of SmolLM2-Instruct.
}
\label{fig:rstar_skills}
\end{minipage}
\end{figure} 

\begin{figure}[t]
\centering
\includegraphics[width=\textwidth]{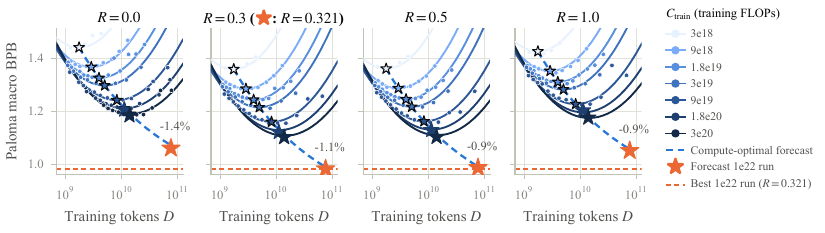}
\caption{\textbf{The surface fit to budgets through $3\times10^{20}$ FLOPs forecasts the $10^{22}$ runs to within $1.4\%$.} Each panel is one rewrite fraction $R$. Dots are runs and curves are the fit IsoFLOP slices, colored by training budget. The $R=0$ ladders reach $3\times10^{20}$ FLOPs and the $R>0$ ladders $9\times10^{19}$ FLOPs. Open stars mark the fit optimum at each budget, filled blue stars mark the compute-optimal runs at $1.8\times10^{20}$ and $3\times10^{20}$ FLOPs, and the dashed line extends the compute-optimal forecast to $10^{22}$ FLOPs. The orange star is the $10^{22}$ run, held out of the fit, labeled with its measured loss relative to the forecast. The $R=0.32$ run's forecast error is $-1.1\%$ whether the surface is evaluated at $R=0.3$ or at $R=0.32$. Figure~\ref{fig:isoflop_surface_appendix} shows all eleven rewrite fractions. 
}
\label{fig:isoflop_surface}
\end{figure}

\subsubsection{Compute-Limited Experimental Results}

\paragraph{The estimated optimal rewrite fraction increases in FLOP budget and varies between tasks (RQ1).}
Figure~\ref{fig:rstar_skills} shows the estimated optimal rewrite fraction $\Rhat$ for each different eval as a function of FLOP budget. 
For all evals, $\Rhat$ monotonically increases with FLOP budget.
This is consistent with the observation that as the total compute budget (and in particular the size of the target model) grows, the cost of generating synthetic rewrites shrinks relative to the cost of training.

We find that $\Rhat$ changes across target domains: on Code and Math, $\Rhat$ rises more quickly than on QA and Paloma.
On Code and Math, $\Rhat$ rises to $1.0$ at FLOP budgets within our scaling ladder; on QA and Paloma, our scaling law fit extrapolates asymptotes of $0.77$ and $0.32$ respectively. 
SmolLM2-1.7B was trained on FineMath, Stack-Edu, and StarCoderData \citep{benallal2025smollm2smolgoesbig,li2023starcoder}, which contain higher quality data for math and code capabilities than in the natural FineWeb-Edu corpus. 
The fact that math and code eventually require a rewrite fraction of $R=1$ suggests that it is easier to learn math and code capabilities by proxy through SmolLM2-1.7B rewrites than from the natural FineWeb-Edu data itself.

The differences in loss between the predicted optimum $\Rhat$ and suboptimal $R$ points are substantial: at $10^{22}$ FLOPs, training on rewrites alone ($R=1$) instead of natural data alone ($R=0$) reduces Math BPB from $1.33$ to $0.80$ and Code BPB from $1.04$ to $0.53$.

\paragraph{The estimated optimal rewrite fraction extrapolated from smaller budgets gives the lowest 
measured Paloma BPB at $10^{22}$ FLOPs, across all models we train (RQ1).}
Our scaling law extrapolation predicts that Paloma BPB at a total budget of $10^{22}$ FLOPs will be minimized at rewrite fraction $R=0.32$ (Figure~\ref{fig:rstar_skills}). 
We train four models at $10^{22}$ FLOPs that share $(N,D)$ and differ in $R \in \{0, 0.32, 0.5, 1.0\}$.
Of these, $R=0.32$ is indeed optimal (Figure~\ref{fig:isoflop_surface}).

\paragraph{The $10^{22}$-FLOP run at the predicted optimum reaches a lower Paloma BPB than its rewriter on $10$ of $16$ sources (RQ2).}
SmolLM2-1.7B-Instruct was trained on $11$T tokens at ${\approx}1.1\times10^{23}$ FLOPs.
The $10^{22}$-FLOP run at $R=0.32$ reaches a lower Paloma BPB than its rewriter on $10$ of the $16$ Paloma sources %
and comes within $0.012$ BPB of it on the macro average, $0.983$ against $0.972$ (Table~\ref{tab:paloma_sources}). 
The sources where the run does not break even with the rewriter are code (which the rewriter's training corpus covers more than our natural corpus FineWeb-Edu \citep[Figure 2]{benallal2025smollm2smolgoesbig}), social media, and two web mixtures. 
Table~\ref{tab:paloma_sources} also compares against a $10^{22}$ run at $R=0$ which underperforms $R=0.32$ on most benchmarks, showing that including some rewrites helps surpass the rewriter at this compute budget.
The rewriter's performance on a benchmark therefore does not represent a hard limit on the gains from its rewrites in the compute-limited setting.

\input{sections/tables/paloma_sources_main}

\subsection{Data-Limited Experiments}
\label{sec:data_eff_exp}

In the \textit{data-limited} setting, we assume a fixed FLOP budget, a limited natural seed set ($T \ll D$), and for each target model a corresponding \emph{bootstrapped rewriter} model trained for fewer FLOPs than that target model on the same fixed natural seed set.

\paragraph{Scaling ladder methodology.}
Throughout, we assume we have access to a small seed set of $T=5$B tokens sampled from FineWeb-Edu.
We train target models with $550$M, $837$M, $1$B, and $3$B parameters and measure each target model's total compute budget in \emph{epoch budgets} $W = C_{\mathrm{total}}/6NT$, the compute required to train an $N$-parameter model, epoching the entire seed set $W$ times.
We train models with epoch budgets $W \in \{3, 6, 9, \ldots, 24\}$.
At each budget, we sweep rewrite fractions $R \in \{0, 0.13, 0.29, 0.47, 0.71, 1\}$.\footnote{
The rewrite fractions are chosen so that the fraction of FLOPs spent generating and training on rewrites, $5R/(3+2R)$, runs from $0$ to $1$ in steps of $0.2$ (our rewriter has the same parameter count as our target model).}
We report Paloma bits per byte at the final checkpoint (best-checkpoint results: Appendix~\ref{data_eff_best_checkpoint}).

\paragraph{Rewriting methodology.}
For each target model, we pretrain a rewriter model with the same number of parameters on the seed set (i.e., $W=1$; Appendix~\ref{data_eff_base}).
Pretrained models typically do not follow instructions, so we then fine-tune the rewriter on GLM-5.2 \citep{glm52} rewrites of $1000$ seed-set documents ($7$M tokens).
We select the checkpoint with an LM judge and use it to rewrite the seed set under the same four FinePhrase prompts (Appendix~\ref{rewriter_sft}).\footnote{The small set of rewritten documents limits possible knowledge transfer from GLM-5.2 to the rewriter. Appendix~\ref{alignment} compares this recipe with the SmolLM2 SFT+DPO recipe.}
For synthetic rewrites within each mixture, we sample the four rewriting styles in proportion to their token counts.

\begin{figure}[t]
\centering
\includegraphics[width=\textwidth]{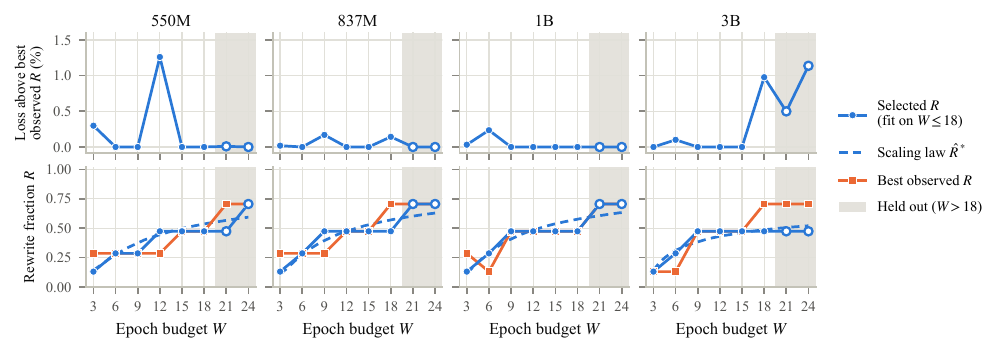}
\caption{
    \textbf{The $\Rhat$ selected by the scaling law costs at most $1.14\%$ in loss at held-out budgets.}
    The law (Eq.~\ref{eq:rewriting_scaling_law}) is fit per target size to budgets through $W=18$; $W=21$ and $W=24$ are held out (shaded, open markers). 
    \textbf{Top:} measured final-checkpoint Paloma macro BPB at $\Rhat$, as a percentage above the loss at the best observed $R$ on the same six-value grid. \textbf{Bottom:} $\Rhat$ (blue), the best observed $R$ (orange), and the law's optimum over continuous $R\in[0,1]$ (dashed). 
    }
\label{fig:de-ucurves}
\end{figure}

\begin{figure}[t]
\centering
\includegraphics[width=\textwidth]{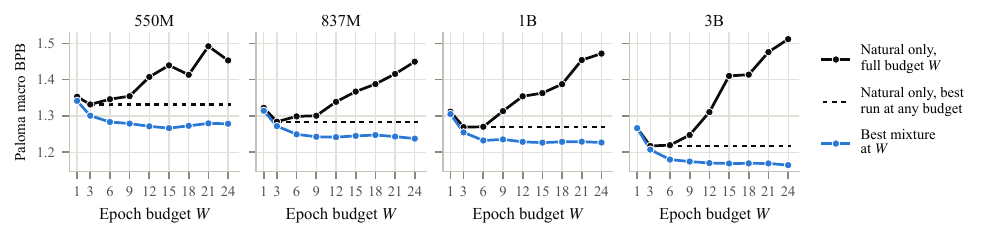}
\caption{Final-checkpoint Paloma macro BPB against the epoch budget $W$ for each target size. \textbf{Solid black:} natural-only epoching ($R=0$) for the full budget of $W$ epochs over the seed set. \textbf{Dotted black:} the best natural-only run at any budget, defined by the loss reached by stopping once further epochs raise it. \textbf{Blue:} best mixture at each $W$, the run with the lowest final loss at that budget over all rewrite fractions, so it coincides with the natural-only point wherever natural-only is best.}
\label{fig:de-ckpt}
\end{figure}

\subsubsection{Data-Limited Experimental Results}

\paragraph{The best observed rewrite fraction increases in epoch budget and is predictable at unseen budgets (RQ1).}
The best observed rewrite fraction---the tested $R$ resulting in the lowest final loss at each size and budget---increases with the epoch budget $W$ at every target size (Figure~\ref{fig:de-ucurves}).
For the $3$B target, the best observed $R$ increases from $0.13$ at $W=3$ to $0.71$ at $W=24$.

We predict $\Rhat$ in the data limited setting by fitting the rewriting scaling law (Eq.~\ref{eq:rewriting_scaling_law})
for each target size to budgets through $W=18$ and forecast $\Rhat$ for the held-out $W=21$ and $W=24$ runs.
Figure~\ref{fig:de-ucurves} presents the results of this experiment.
At $R^*$, the measured loss is within $0.74\%$ of the forecast at every target size for both held-out budgets.
For the $550$M, $837$M, and $1$B target models, $\Rhat$ empirically minimizes Paloma BPB among those tested $R$ (except for $550$M at $W=21$, where it is within $0.01\%$ of the Paloma BPB at the best choice of $R$).
For the $3$B target model at $W=21$, the scaling law selects $\Rhat=0.47$ and achieves a Paloma BPB of $1.175$, $0.5\%$ above the best observed $R=0.71$; at $W=24$ the scaling law selects $\Rhat=0.47$ and achieves a Paloma BPB of $1.178$, $1.14\%$ above the loss at the best observed $R=0.71$.

\begin{figure}[t]
\vspace{-1em}
\centering
\includegraphics[width=\textwidth]{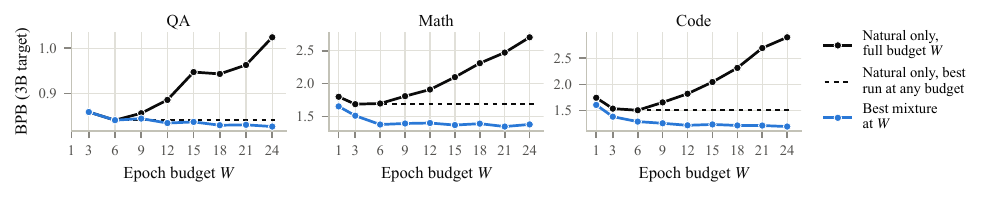}
\caption{\textbf{Rewrites beat natural-only epoching on Math and Code at every budget, but on QA only after $W=6$.} Final-checkpoint BPB for the $3$B target against the epoch budget $W$ on QA, Math, and Code, with the same three series as Figure~\ref{fig:de-ckpt}. Appendix~\ref{data_eff_domains} shows every target size.}
\label{fig:de-domains}
\end{figure}

\paragraph{Rewrites reach a lower loss than natural-only epoching at every model size (RQ2).}
Figure~\ref{fig:de-ckpt} compares natural-only ($R=0$) models to those trained at $\Rhat$ across a range of epoch budgets $W$;
in particular, these plots show the final checkpoint's Paloma BPB, with the result that for epoch budgets $W>3$ the natural-only runs overfit.
Training at $\Rhat$ results in lower Paloma BPB than training at $R=0$ at every size.

\paragraph{The best observed rewrite fraction depends on the task.}
Figure~\ref{fig:de-domains} repeats Figure~\ref{fig:de-ckpt} for the $3$B target on QA, Math, and Code. On Math and Code the best mixture with $R>0$ beats the best natural-only run at every budget. %
On QA the natural-only run is best through $W=6$, by $0.014$ BPB; beyond that, including synthetic rewrites outperforms.

%% file: sections/tables/paloma_sources_main.tex
\begin{table}[t]
\begin{minipage}[c]{0.50\textwidth}
\centering
\footnotesize\setlength{\tabcolsep}{3pt}\renewcommand{\arraystretch}{0.92}
\resizebox{1.0\linewidth}{!}{%
\begin{tabular}{r|ccc}
\toprule
Paloma source & $R=0$ & $R=0.32$ & Rewriter \\
\midrule
M2D2 Wikipedia & 0.726 & \textbf{0.716} & 0.758 \\
Penn Treebank & 0.851 & \textbf{0.843} & 0.882 \\
WikiText-103 & 0.682 & \textbf{0.673} & 0.711 \\
C4 English & \textbf{0.827} & 0.828 & 0.844 \\
Falcon RefinedWeb & 0.894 & \textbf{0.887} & 0.897 \\
Gab & 1.719 & \textbf{1.579} & 1.587 \\
C4 100 Domains & \textbf{0.790} & 0.793 & 0.797 \\
M2D2 S2ORC & 0.766 & \textbf{0.760} & 0.764 \\
mC4 & 0.841 & \textbf{0.825} & 0.828 \\
TwitterAAE & 2.591 & \textbf{2.453} & 2.455 \\
\midrule
Dolma v1.5 & 0.876 & 0.804 & \textbf{0.791} \\
Dolma 100 Subreddits & 1.089 & 1.022 & \textbf{1.001} \\
Manosphere & 1.112 & 1.057 & \textbf{1.030} \\
RedPajama & 0.946 & 0.775 & \textbf{0.737} \\
4chan & 1.162 & 1.023 & \textbf{0.952} \\
Dolma 100 PLs & 1.106 & 0.694 & \textbf{0.511} \\
\midrule
Macro average & 1.061 & 0.983 & \textbf{0.972} \\
\bottomrule
\end{tabular}
}%
\end{minipage}\hfill
\begin{minipage}[c]{0.47\textwidth}
\raggedright
\caption{\textbf{Our run at $10^{22}$ training FLOPs reaches a lower Paloma BPB than its rewriter on $10$ of $16$ sources.} Per-source Paloma BPB for the natural-only ($R=0$) and predicted-optimum ($R=0.32$) runs at $10^{22}$ training FLOPs and for SmolLM2-1.7B-Instruct, sorted by the predicted-optimum run's margin over the rewriter; the rule separates the sources where it is at or below the rewriter. Bold marks the lowest BPB per row. Appendix~\ref{hero_evals} reports all four runs at $10^{22}$ training FLOPs and the SmolLM2 base model.}
\label{tab:paloma_sources}
\end{minipage}
\end{table}

%% file: sections/related.tex
\section{Related Work}

\paragraph{Data-constrained pretraining.}
When the corpus is finite, additional compute means additional epochs, whose returns diminish \citep{muennighoff2023scalingdataconstrainedlanguage}. \citet{lovelace2026prescriptivescalinglawsdata} model this with a repetition penalty that activates beyond the first epoch, which we adopt as the $F(R, D, T)$ term of Eq.~(\ref{eq:rewriting_scaling_law}). \citet{kim2025pretraininginfinitecompute,kim2026dataefficientpretrainingscaling} study the same regime from the perspective of a fixed data budget and unlimited compute, which is our data-limited setting.

\paragraph{Synthetic rewriting.}
Rewriting prompts a language model to restate a pretraining document, typically in a target style, and trains on the result alongside or instead of the original \citep{maini2024rephrasingwebrecipecompute}. Rewrites are now a component of several high-quality pretraining corpora \citep{su2024nemotroncctransformingcommoncrawl,nguyen2025recyclingwebmethodenhance,niklaus2026synthesizehighqualitypretrainingdata,karimimahabadi2025nemotronccmath,fujii2025rewritingpretrainingdataboosts,benallal2025smollm2smolgoesbig}, and \citet{maini2025beyondweblessonsscalingsynthetic} report that their benefit persists with scale.

\paragraph{Rewriting as distillation.}
Whether rewrites help because the rewriter knows more than the corpus is a standing question. \citet{maini2024rephrasingwebrecipecompute} ablate the rewriter and find that a $1.8$B model retains most of the gain of a $7$B one. \citet{maini2025beyondweblessonsscalingsynthetic} find at trillion-token scale that returns to rewriter size saturate near $3$B parameters and that a rewriter's benchmark scores are a poor guide to the value of its rewrites. \citet{bootstrapped} remove the external rewriter altogether, training a synthesizer on the pretraining corpus itself, and show that a $3$B model still gains from the documents it synthesizes.
Section~\ref{sec:compute_eff_exp} and Section~\ref{sec:data_eff_exp} measure whether the rewriter's BPB caps the gains (RQ2) in each setting.

\paragraph{Scaling laws for synthetic pretraining data.}
Three recent studies fit scaling laws to synthetic pretraining data.
\citet{kang2025demystifyingsyntheticdatallm} train over a thousand models of $100$M--$3$B parameters on mixtures of CommonCrawl with rephrased and textbook-style synthetic data from a $7$B generator, fitting a separate Chinchilla law to each mixture: the best rephrased mixtures lie near $30\%$ synthetic, pure rephrased data underperforms natural text, and mixtures containing rephrased data are projected to reach a lower irreducible loss than natural text alone.
\citet{qin2026bridgingcomputedataoptimalpretraining} unify compute- and data-optimal scaling through a token-effectiveness function that discounts derived tokens, repetitions or SmolLM2-1.7B-instruct paraphrases of a fixed fresh corpus, relative to fresh ones, fit jointly on models of $14$M--$600$M parameters. They find that effectiveness falls with model size and tokens per parameter. %
\citet{qin2025scaling} scale synthetic math data alone, generated from FineWeb-Edu math documents by a $123$B model, in continued training of $1$B--$8$B base models, and find a rectified scaling law in synthetic tokens that plateaus near $300$B tokens, with larger models needing fewer.
None of the three account for compute spent on generation, so their optimal mixtures answer a different question: how much to rewrite when rewrites are free.
\citet{qin2026bridgingcomputedataoptimalpretraining} project that paraphrasing should stop helping beyond $7$B parameters. 
However, our $22$B-parameter run at the predicted optimum reaches $0.983$ Paloma BPB against $1.061$ for natural data alone at the same budget. 
All three also use an off-the-shelf generator stronger than the models they train. 
In the data-limited setting we pretrain the rewriter on the seed set and instruction-tune it on only $7$M tokens of external rewrites, so gains from its rewrites can draw on little knowledge the corpus lacks.
Appendix~\ref{infinite_compute} tests the token-effectiveness model of \citet{qin2026bridgingcomputedataoptimalpretraining} against ours.

%% file: sections/conclusion.tex
\section{Conclusion and Limitations}

We asked whether scaling laws predict the cost-adjusted optimal rewrite fraction and associated loss (RQ1) and whether rewrites help when targeting capabilities beyond the rewriter (RQ2). 
Our rewriting scaling law answers both. 
In the compute-limited setting, the rewrite fraction fit at small budgets minimizes loss among the fractions we train at $33\times$ the compute, and a pretrained model with 32\% rewritten training data outperforms the rewriter on 10/16 general web text domains. 
In the data-limited setting, we similarly find that the optimal rewrite fractions and losses of held-out budgets can be predicted via scaling laws, and that a positive proportion of rewrites is additive even when the rewriter model is less capable than the target model. 

\paragraph{Limitations.}
Several choices limit the extent of our results. 
One is the choice of the natural seed set: every optimal rewrite fraction we report is for FineWeb-Edu.
We also study this question on the DCLM seed set, where rewrites still help but by less (Appendix~\ref{dclm_scaling}), suggesting that a strong natural seed set benefits less from rewriting. 
We also test only the four FinePhrase rewrite styles and one external rewriter in each setting, so we cannot say how the optimal fraction varies across rewrite prompt distributions or across a family of rewriters. 
Finally, our compute accounting is approximate: we charge $4N_r$ FLOPs per rewritten token, which approximates decoding as costing the same FLOPs as prefill and assumes the rewrite is the same length as its source; we do not account for costs associated with pretraining the rewriter model.

%% file: sections/ai_use.tex
\subsection*{AI Use Statement}
This paper studies scaling laws for synthetic data generated by language models. 
Rewrites in the compute-limited setting are FinePhrase rewrites generated by SmolLM2-1.7B, and rewrites in the data-limited setting are generated by rewriters we pretrain ourselves and SFT on GLM-5.2 rewrites of $1000$ seed-set documents. 
Finally, an LM judge selects the rewriter checkpoint (Section~\ref{sec:recipe}, Appendices~\ref{rewriter_sft} and~\ref{generation}). 
We used AI assistants to discuss analysis choices, to implement and debug analysis code, for initial drafts of paper sections, and for feedback on prose.  
We have reviewed, verified, and updated code and text originating from AI.

%% file: sections/acks.tex
\subsubsection*{Acknowledgments}
We thank Lucas Torroba-Hennigen, Howard Yen, and Abbas Zeitoun for the many thoughtful discussions during the internship---we learned so much from you all. 
We also thank Rylan Schaeffer and Kaiyue Wen for their helpful feedback on drafts of this paper. 
Finally, we are grateful to the members of Danqi Chen’s, Sanjeev Arora’s, and Frederic Sala’s groups for their valuable feedback.

%% file: sections/appendix.tex
\section*{Appendix Roadmap}
\begin{figure}[t]
\begin{center}
\includegraphics[width=\textwidth]{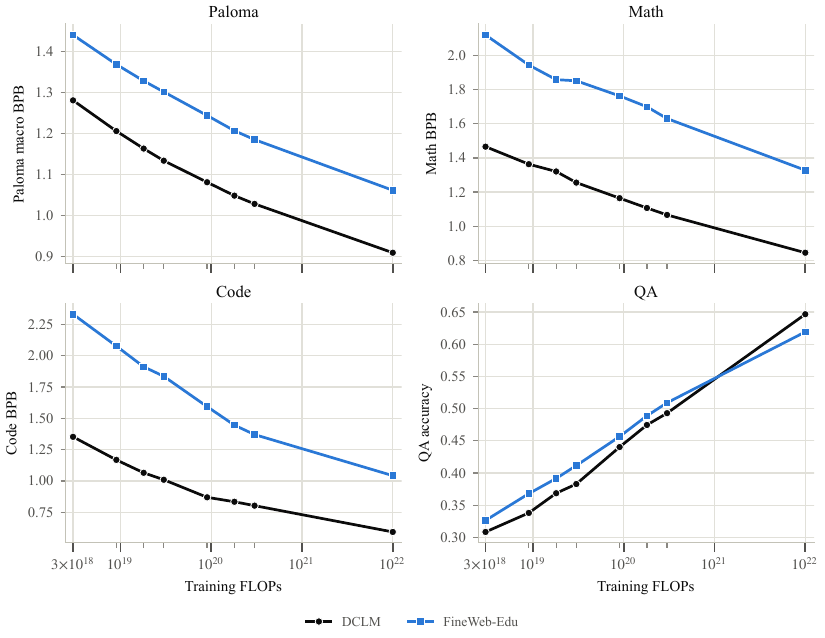}
\end{center}
\caption{Final evaluations of the observed compute-optimal natural-only DCLM and FineWeb-Edu runs. At each training budget and corpus, we select the $R=0$ run with the lowest final Paloma macro BPB and show that run's Paloma, Math, Code, and QA scores. Lower is better for the three BPB panels and higher for QA accuracy.}
\label{fig:dclm-scaling-comparison}
\end{figure}

\noindent
\begin{minipage}[t]{0.48\textwidth}
\vspace{0pt}
\textbf{Setup and evaluation}
\begin{itemize}[leftmargin=*,nosep]
\item \hyperref[fwedu_data]{Data sources} (page~\pageref*{fwedu_data})
\item \hyperref[model_arch]{Model architecture} (page~\pageref*{model_arch})
\item \hyperref[completedp]{Hyper-parameter recipe} (page~\pageref*{completedp})
\item \hyperref[evals]{Evaluation tasks and subtasks} (page~\pageref*{evals})
\item \hyperref[mixture_construction]{Mixture construction} (page~\pageref*{mixture_construction})
\end{itemize}

\textbf{Data-limited experiments}
\begin{itemize}[leftmargin=*,nosep]
\item \hyperref[data_eff_base]{Rewriter capacity} (page~\pageref*{data_eff_base})
\item \hyperref[lm_judge]{Checkpoint selection with an LM judge} (page~\pageref*{lm_judge})
\item \hyperref[alignment]{Alignment ablation} (page~\pageref*{alignment})
\item \hyperref[data_eff_appendix]{Complete budget grid} (page~\pageref*{data_eff_appendix})
\item \hyperref[fig:de-ucurves-all]{Full-budget U-curves} (Figure~\ref*{fig:de-ucurves-all}, page~\pageref*{fig:de-ucurves-all})
\item \hyperref[data_eff_domains]{Domain loss across target sizes} (Appendix~\ref*{data_eff_domains}, page~\pageref*{data_eff_domains})
\item \hyperref[data_eff_best_checkpoint]{Best-checkpoint evaluation} (page~\pageref*{data_eff_best_checkpoint})
\end{itemize}

\textbf{Rewrite analysis}
\begin{itemize}[leftmargin=*,nosep]
\item \hyperref[generation]{Rewrite generation and cost} (page~\pageref*{generation})
\item \hyperref[rewrite_faithful]{Style and line-content ablations} (page~\pageref*{rewrite_faithful})
\item \hyperref[qualitative_rewrites]{Qualitative rewrite examples} (page~\pageref*{qualitative_rewrites})
\item \hyperref[scaling_law_details]{Rewriting scaling law details} (page~\pageref*{scaling_law_details})
\item \hyperref[indep_R_N_D]{Allocation dependence on $R$} (page~\pageref*{indep_R_N_D})
\end{itemize}
\end{minipage}
\hfill
\begin{minipage}[t]{0.48\textwidth}
\vspace{0pt}
\textbf{Compute-limited experiments}
\begin{itemize}[leftmargin=*,nosep]
\item \hyperref[compute_eff_appendix]{Experimental grid} (page~\pageref*{compute_eff_appendix})
\item \hyperref[fig:isoflop_surface_appendix]{Full-$R$ IsoFLOP surface} (Figure~\ref*{fig:isoflop_surface_appendix}, page~\pageref*{fig:isoflop_surface_appendix})
\item \hyperref[fig:frontier_by_r]{Per-$R$ frontier and checkpoint check} (Figures~\ref*{fig:frontier_by_r}--\ref*{fig:checkpoint_by_r}, page~\pageref*{fig:frontier_by_r})
\item \hyperref[infinite_compute]{Infinite-compute extrapolation} (page~\pageref*{infinite_compute})
\item \hyperref[scaling_fit]{Fits and forward forecasts} (page~\pageref*{scaling_fit})
\item \hyperref[scaling_curv_ablation]{Curvature ablation} (page~\pageref*{scaling_curv_ablation})
\item \hyperref[knowledge_reasoning]{Domain-specific optima} (page~\pageref*{knowledge_reasoning})
\item \hyperref[hero_evals]{Evaluations of the $10^{22}$-FLOP runs} (page~\pageref*{hero_evals})
\item \hyperref[breakeven_parameters]{Break-even allocations} (page~\pageref*{breakeven_parameters})
\item \hyperref[loss_accuracy]{Loss- versus accuracy-based fits} (page~\pageref*{loss_accuracy})
\item \hyperref[scaling_setup]{Surface-specification ablations} (page~\pageref*{scaling_setup})
\end{itemize}

\textbf{DCLM diagnostics}
\begin{itemize}[leftmargin=*,nosep]
\item \hyperref[fig:dclm-ladder-grid]{DCLM IsoFLOP ladders} (Figure~\ref*{fig:dclm-ladder-grid}, page~\pageref*{fig:dclm-ladder-grid})
\item \hyperref[fig:dclm-scaling-comparison]{DCLM vs. FineWeb-Edu ($R=0$)} (Figure~\ref*{fig:dclm-scaling-comparison}, page~\pageref*{fig:dclm-scaling-comparison})
\item \hyperref[dclm_scaling]{Source quality: DCLM rewrite sweep and transfer} (page~\pageref*{dclm_scaling})
\end{itemize}
\end{minipage}

\section{Data Sources}
\label{fwedu_data}

\paragraph{Natural text.}
The primary experiments use FineWeb-Edu \citep{penedo2024finewebdatasetsdecantingweb}. We drop every document whose WARC file is dated 2023 or later, to limit LM-generated text in the natural corpus, and shuffle the remaining documents. The data-limited experiments fix a $5$B-token seed set from this corpus. The compute-limited experiments sample fresh documents at each training-token budget. The source-quality control replaces FineWeb-Edu with DCLM \citep{li2024datacomplmsearchgenerationtraining}.

\paragraph{Rewritten text.}
The compute-limited experiments draw rewrites from FinePhrase \citep{niklaus2026synthesizehighqualitypretrainingdata}, which contains approximately $338$M rewrites of FineWeb-Edu documents in each of the FAQ, math, table, and tutorial styles. The data-limited experiments generate rewrites of the seed set with size-matched bootstrapped rewriters (Appendix~\ref{generation}).

\section{Model Architecture}
\label{model_arch}
All runs use the dense Qwen3 architecture, the Llama~3 tokenizer \citep{grattafiori2024llama3herdmodels}, and a sequence length of $4096$. The hidden dimension $d$ determines each model size. The MLP intermediate dimension is $4d$. Attention heads have dimension $128$, giving $d/128$ query and key--value heads. The log-corrected depth-to-width ratio of \citet{levine2021depthtowidthinterplayselfattention} sets the number of layers:
\begin{align}
L(d) = \operatorname{round}\left( \frac{d}{55 + 4\log_2 d} \right).
\end{align}
This rule assigns $11$ layers at $d=1024$ and $21$ layers at $d=2048$. Appendix~\ref{completedp} specifies the candidate widths and selection rule.

\section{Hyper-parameter Tuning}
\label{completedp}
The Completed(d)P recipe of \citet{completed-hyperparameter}, implemented in Marin \citep{held2026_delphi}, scales one reference configuration to each run's batch size $B_{\mathrm{batch}}$ and training-token count $D$. We measure $B_{\mathrm{batch}}$ in sequences of $4096$ tokens.

\paragraph{Optimizer and schedule.}
We train with Adam and Hyperball~\citep{wen2026fantasticpretrainingoptimizersii}; \citet{wen2026fantasticpretrainingoptimizersii} show that Hyperball makes the optimal learning rate robust across model sizes. Following \citet{held2026_delphi}, we tune the learning rates on the $156$M-parameter reference model with Paloma as the objective, then scale the learning rates and the remaining hyper-parameters to each run's batch size and token count. Every run uses linear warm-up over the first $10\%$ of steps and linear decay to zero.

\paragraph{Reference point.}
The reference configuration is a $d=512$ ($156$M-parameter) model trained with $B_{\mathrm{batch},0}=64$ and $D_0=2.5\times10^9$ tokens on FineWeb-Edu. Its hyper-parameters are the best of a Gaussian-process search over $430$ configurations at this scale, selected on Paloma validation macro cross-entropy as in Delphi \citep{held2026_delphi}: Hyperball learning rate $\eta_0=6.30\times10^{-3}$, Adam learning rate $\eta^{\mathrm A}_0=6.56\times10^{-4}$, $\epsilon_0=1.85\times10^{-8}$, $\beta_1=0.9$, and $\beta_{2,0}=0.9999$.

\paragraph{Scaling rules.}
With $r/r_0 = (B_{\mathrm{batch}}\,D_0)/(B_{\mathrm{batch},0}\,D)$, a run with batch size $B_{\mathrm{batch}}$ and $D$ tokens uses
\begin{align}
\eta &= \min\left(0.01,\; \eta_0 \sqrt{B_{\mathrm{batch}}/B_{\mathrm{batch},0}}\,(D_0/D)^{0.3}\right), \\
\eta^{\mathrm{A}} &= \min\left(0.01,\; \eta^{\mathrm{A}}_0 \sqrt{r/r_0}\right), \\
\epsilon &= \epsilon_0 \sqrt{r_0/r}, \\
\beta_2 &= \operatorname{clip}\left(\beta_{2,0}^{\,B_{\mathrm{batch}}/B_{\mathrm{batch},0}},\; 0.9,\; 0.9999\right).
\end{align}
The $\beta_2$ rule keeps its half-life constant in tokens. Every run uses gradient clipping at $0.1$, $z$-loss weight $10^{-7}$, linear warm-up over the first $10\%$ of steps, and linear decay to zero.

\paragraph{Batch size.}
Each run targets $2^{16}$ steps. We round $D/(2^{16}\cdot4096)$ up to a power of two, then halve the batch size until $\eta<0.01$, $\eta^{\mathrm A}<0.01$, and $\beta_2>0.9$, subject to $8\leq B_{\mathrm{batch}}\leq8192$.

\paragraph{Candidate models per budget.}
For each training budget $C$, we enumerate hidden dimensions from $512$ in increments of $128$ when $C\leq2\times10^{19}$ FLOPs and $256$ otherwise. Each architecture follows Appendix~\ref{model_arch} and receives $D=C/(3F_{\mathrm{fwd}})$ tokens, where $F_{\mathrm{fwd}}$ denotes forward FLOPs per token. Budgets are labeled by this nominal $C$; the scaling-law fits and all compute accounting use $C=6ND$ as in Section~\ref{sec:scaling_model}. We exclude models above $12\mathrm{B}\cdot\max\!\left(1,\sqrt{C/(3\times10^{20})}\right)$ parameters or $250$ tokens per parameter.

\section{Evaluation Details}
\label{evals}

We evaluate language-model fit with Paloma \citep{magnusson2024paloma} and question answering, mathematics, and code with offline evaluation harnesses.\footnote{The SmolLM2 Paloma reference uses OLMES (\url{https://github.com/allenai/olmes}). The domain-likelihood suites use OLMo Eval (\url{https://github.com/allenai/olmo-eval}).} Unless stated otherwise, each metric is bits per byte (BPB): reference-continuation negative log-likelihood in bits divided by UTF-8 byte length. Lower is better.

\paragraph{Paloma.}
Paloma macro BPB averages 16 sources uniformly: 4chan, C4 100 Domains, C4 English, Dolma v1.5, Dolma 100 Programming Languages, Dolma 100 Subreddits, Falcon RefinedWeb, Gab, M2D2 S2ORC, M2D2 Wikipedia, Manosphere, mC4, Penn Treebank, RedPajama, TwitterAAE, and WikiText-103. Both settings use the training-loop evaluator at the models' trained $4096$-token context and pool bytes within each source. We evaluate SmolLM2 at a forced $4096$-token context and match the within-source pooling for the rewriter comparison.

\paragraph{Question answering.}
The QA suite contains ARC-Easy, ARC-Challenge, HellaSwag, and all 57 MMLU subjects. We first average the two ARC tasks and the 57 MMLU subjects within their families, then average ARC, HellaSwag, and MMLU uniformly. The MMLU subjects are:
\begin{itemize}
    \item \emph{STEM:} abstract algebra, anatomy, astronomy, college biology,
    college chemistry, college computer science, college mathematics, college
    physics, computer security, conceptual physics, electrical engineering,
    elementary mathematics, high-school biology, high-school chemistry,
    high-school computer science, high-school mathematics, high-school physics,
    high-school statistics, machine learning, and medical genetics;
    \item \emph{humanities:} formal logic, high-school European history,
    high-school US history, high-school world history, international law,
    jurisprudence, logical fallacies, moral disputes, moral scenarios,
    philosophy, prehistory, professional law, and world religions;
    \item \emph{social sciences:} econometrics, high-school geography,
    high-school government and politics, high-school macroeconomics,
    high-school microeconomics, high-school psychology, human sexuality,
    professional psychology, public relations, security studies, sociology, and
    US foreign policy; and
    \item \emph{other:} business ethics, clinical knowledge, global facts,
    human aging, management, marketing, college medicine, miscellaneous,
    nutrition, professional accounting, professional medicine, and virology.
\end{itemize}
Rank classification scores every answer choice. The scaling surfaces use gold-answer BPB (\texttt{qa\_rc:bpb:base\_easy}). The secondary metric is the fraction of examples on which the gold answer ranks first (\texttt{qa\_rc:base\_easy}).

\paragraph{Mathematics.}
The Math metric \texttt{math\_bpb:all} uniformly averages nine gold-answer BPB tasks: GSM8K-Platinum, TheoremQA, and the seven Minerva MATH subjects (algebra, counting and probability, geometry, intermediate algebra, number theory, prealgebra, and precalculus).

\paragraph{Code.}
The Code suite contains three-shot HumanEval, three-shot MBPP, and three-shot multilingual MBPP-v2fix in Bash, C, C++, C\#, Go, Haskell, Java, JavaScript, MATLAB, PHP, Python, R, Ruby, Rust, Scala, Swift, and TypeScript. We first average the 17 multilingual variants, then average HumanEval, MBPP, and multilingual MBPP uniformly (\texttt{code\_bpb:base\_easy}). The metric is reference-code BPB, not execution accuracy.

\section{Data Limited Base Rewriters}
\label{data_eff_base}

\begin{figure}[t]
\begin{center}
\includegraphics[width=\textwidth]{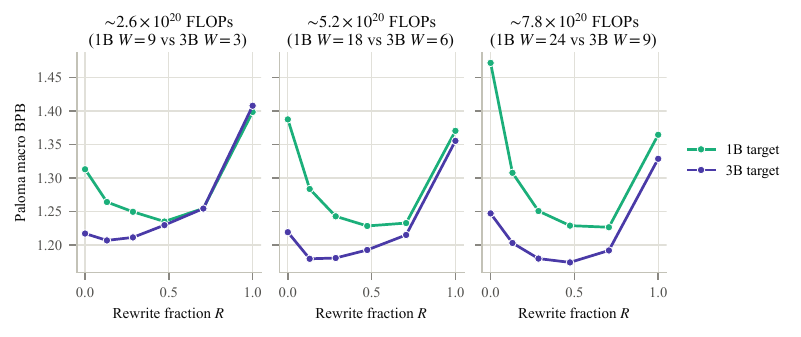}
\end{center}
\caption{$1$B and $3$B targets at matched total compute. Each panel pairs the nearest budgets on the grid, with FLOP ratios $0.97$, $0.97$, and $1.09$.}
\label{fig:de-size}
\end{figure}

\begin{figure}[t]
\begin{center}
\includegraphics[width=\textwidth]{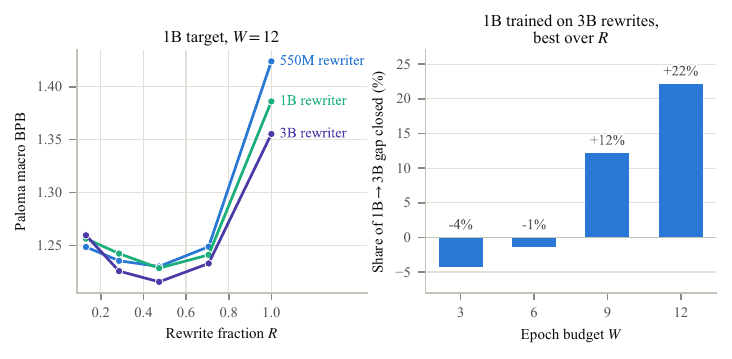}
\end{center}
\caption{Rewriter capacity. \textbf{Left:} $1$B target loss at $W=12$ with rewrites from the $550$M, $1$B, and $3$B rewriters. \textbf{Right:} share of the $1$B--$3$B target gap closed by training the $1$B target on $3$B rewrites, at each budget's best mixture.}
\label{fig:de-capacity}
\end{figure}

\paragraph{Effect of rewriter capacity.}
The capacity control trains the $1$B target on rewrites from the $550$M and $3$B rewriters at $W\le12$. At $W=12$, the $550$M rewriter sits $0.011$ BPB below the $3$B rewriter at $R=0.13$ and trails it by $0.069$ BPB at $R=1$ (Figure~\ref{fig:de-capacity}, left). Training the $1$B target on $3$B rewrites closes $52\%$ of the $1$B--$3$B target gap at $R=1$ and $W=12$, and between $-4\%$ and $22\%$ at each budget's best mixture (right). Figure~\ref{fig:de-size} pairs the $1$B and $3$B targets at the nearest total-compute budgets on the grid. The $3$B target has lower BPB in $17$ of the $18$ pairs, by $0.031$ BPB on average at the smallest budget pair and $0.088$ BPB at the largest.

\section{Rewriter Instruction Tuning}
\label{rewriter_sft}
Each data-limited rewriter receives supervised fine-tuning (SFT) after pretraining on the seed set. The SFT set contains $4000$ examples: GLM-5.2 \citep{glm52} rewrites of $1000$ seed-set documents under each of the four FinePhrase prompts, $7$M tokens in total, at sequence length $8192$. Following \citet{zhou2023limalessmorealignment}, we train with a batch size of $32$ examples and linear decay of the learning rate to zero, with no warm-up. We sweep the learning rate and the number of epochs per rewriter size and select the checkpoint with the LM judge of Appendix~\ref{lm_judge}.

\subsection{Checkpoint Selection with an LM Judge}
\label{lm_judge}
The in-house rewriter generates the synthetic data for every data-limited run (Section~\ref{sec:data_eff_exp}), so we choose its fine-tuning learning rate and number of epochs by the quality of the rewrites each checkpoint generates.
At $550$M and $837$M the checkpoints are the $12$ from learning rates $10^{-5}$ to $5\times10^{-4}$ at $10$ or $15$ epochs, at $3$B the six from $5\times10^{-5}$ to $2\times10^{-4}$, and at $1$B a shortlist of seven of $31$ fine-tuning runs that also covers $5$ epochs (Table~\ref{tab:lm-judge-scores}).
Every checkpoint rewrites the same $100$ randomly sampled FineWeb-Edu documents under the four FinePhrase prompts, $400$ rewrites per checkpoint.
The reference for each document and prompt is its FinePhrase rewrite, generated by SmolLM2-1.7B, the rewriter of the compute-limited setting.

GLM-5.2 \citep{glm52} and MiniMax-M2.7 \citep{minimax_m27} judge pairs of rewrites of the same document under the same prompt.
A pair is either a checkpoint's rewrite and the SmolLM2-1.7B rewrite or the rewrites of two checkpoints, and not every pair is judged.
A judge receives the source document, the rewrite prompt, and the two rewrites, and returns $+1$ if the first rewrite is better, $-1$ if the second is better, or $0$ for a tie (prompt below).
Each pair is judged in both presentation orders.
Elo ratings computed from each judge's preferences place every checkpoint and SmolLM2-1.7B on one scale, and the rating difference gives each checkpoint's expected win rate against SmolLM2-1.7B.
A checkpoint's score is this win rate averaged over the two judges.

\begin{tcolorbox}[title={Judge prompt},colback=white,colframe=black!35]
\small\raggedright
Compare how well the two rewrites rewrite the source document according to the style specified in the rewrite prompt.\newline
\newline
Source document: \texttt{\{source\}}\newline
\newline
Rewrite prompt: \texttt{\{prompt\}}\newline
\newline
Rewrite A: \texttt{\{rewrite\_a\}}\newline
\newline
Rewrite B: \texttt{\{rewrite\_b\}}\newline
\newline
Return +1 if A is better, -1 if B is better, or 0 for a tie.
\end{tcolorbox}

Table~\ref{tab:lm-judge-scores} ranks the checkpoints by score and marks the checkpoint that generates the seed-set rewrites at each size: a learning rate of $10^{-4}$ for $15$ epochs at $550$M and $1$B, $2\times10^{-4}$ for $10$ epochs at $837$M, and $5\times10^{-5}$ for $15$ epochs at $3$B.
This is the highest-scoring checkpoint at $550$M, $837$M, and $3$B.
At $1$B it scores $0.7$ points below the highest-scoring checkpoint, and the confidence intervals of the two overlap.
No checkpoint reaches a $50\%$ win rate against SmolLM2-1.7B.

\begin{table}[!htbp]
\centering
\small
\setlength{\tabcolsep}{9pt}
\begin{tabular}{r|ccc}
\toprule
Size & SFT learning rate & Epochs & Mean win rate (\%) \\
\midrule
550M & $10^{-4}$ & 15$^{\dagger}$ & 28.3 \\
 & $2\times 10^{-4}$ & 15 & 27.7 \\
 & $2\times 10^{-4}$ & 10 & 26.6 \\
 & $10^{-4}$ & 10 & 25.0 \\
 & $2\times 10^{-5}$ & 15 & 24.8 \\
 & $2\times 10^{-5}$ & 10 & 23.8 \\
 & $5\times 10^{-5}$ & 10 & 23.2 \\
 & $5\times 10^{-4}$ & 10 & 22.6 \\
 & $5\times 10^{-4}$ & 15 & 22.3 \\
 & $5\times 10^{-5}$ & 15 & 21.9 \\
 & $10^{-5}$ & 15 & 19.9 \\
 & $10^{-5}$ & 10 & 19.4 \\
\midrule
837M & $2\times 10^{-4}$ & 10$^{\dagger}$ & 32.5 \\
 & $2\times 10^{-4}$ & 15 & 31.5 \\
 & $10^{-4}$ & 15 & 30.9 \\
 & $10^{-4}$ & 10 & 29.7 \\
 & $5\times 10^{-5}$ & 15 & 28.0 \\
 & $2\times 10^{-5}$ & 10 & 26.4 \\
 & $5\times 10^{-5}$ & 10 & 26.3 \\
 & $2\times 10^{-5}$ & 15 & 25.3 \\
 & $5\times 10^{-4}$ & 15 & 25.0 \\
 & $5\times 10^{-4}$ & 10 & 24.4 \\
 & $10^{-5}$ & 15 & 23.2 \\
 & $10^{-5}$ & 10 & 20.9 \\
\midrule
1B & $2\times 10^{-4}$ & 10 & 33.3 \\
 & $10^{-4}$ & 15$^{\dagger}$ & 32.6 \\
 & $10^{-4}$ & 10 & 31.7 \\
 & $2\times 10^{-4}$ & 5 & 31.7 \\
 & $10^{-4}$ & 5 & 30.7 \\
 & $2\times 10^{-4}$ & 15 & 30.6 \\
 & $5\times 10^{-5}$ & 15 & 30.3 \\
\midrule
3B & $5\times 10^{-5}$ & 15$^{\dagger}$ & 38.8 \\
 & $10^{-4}$ & 10 & 38.5 \\
 & $5\times 10^{-5}$ & 10 & 38.4 \\
 & $10^{-4}$ & 15 & 37.4 \\
 & $2\times 10^{-4}$ & 10 & 34.5 \\
 & $2\times 10^{-4}$ & 15 & 33.9 \\
\bottomrule
\end{tabular}
\caption{\textbf{Win rates against the SmolLM2-1.7B FinePhrase rewrite, averaged over both judges.} Each row is a rewriter checkpoint. Scores are the Elo-derived win rate against the SmolLM2-1.7B rewrite of the same document and prompt, averaged over GLM-5.2 and MiniMax-M2.7. $\dagger$ marks the checkpoint that generates the seed-set rewrites. Rows are ordered by score within each size. The $1$B results cover seven shortlisted configurations.}
\label{tab:lm-judge-scores}
\end{table}

\section{Alignment}
\label{alignment}

\begin{figure}[t]
\begin{center}
\includegraphics[width=\textwidth]{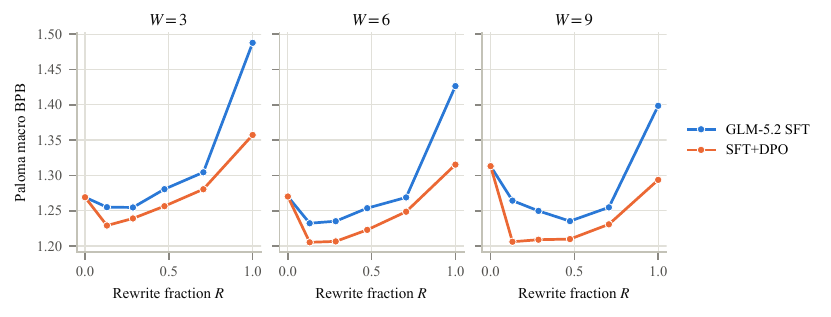}
\end{center}
\caption{Final-checkpoint Paloma macro BPB for the $1$B target trained on rewrites from the SFT and SFT+DPO rewriters at $W\in\{3,6,9\}$. Both conditions share the natural-only $R=0$ runs.}
\label{fig:de-alignment}
\end{figure}

\begin{figure}[t]
\begin{center}
\includegraphics[width=0.8\textwidth]{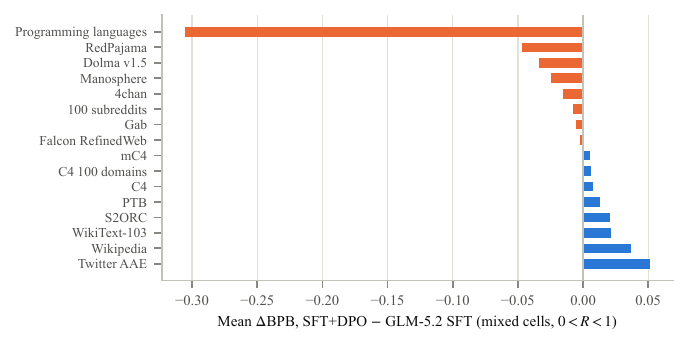}
\end{center}
\caption{Mean Paloma BPB change per source from replacing the SFT rewriter with the SFT+DPO rewriter across the $16$ mixed runs ($0<R<1$, $W\le12$). Orange bars denote lower BPB under SFT+DPO and blue bars lower BPB under the SFT rewriter, matching the colors of Figure~\ref{fig:de-alignment}.}
\label{fig:de-align-domains}
\end{figure}

The data-limited rewriters receive supervised fine-tuning (SFT) on GLM-5.2 rewrites and LM-judge checkpoint selection (Appendix~\ref{rewriter_sft}). The alignment control instead follows the SmolLM2 post-training procedure \citep{benallal2025smollm2smolgoesbig}: SFT followed by direct preference optimization (DPO) on the $1$B rewriter, with no LM-judge selection. Across the $15$ runs with $R>0$ at $W\in\{3,6,9\}$, SFT+DPO lowers Paloma macro BPB in all $15$ and by $0.046$ BPB on average (Figure~\ref{fig:de-alignment}). The programming-languages and RedPajama sources account for $80\%$ of the reduction, and BPB increases on $8$ of the $16$ Paloma sources (Figure~\ref{fig:de-align-domains}).

\section{Data-Limited Experimental Results}
\label{data_eff_appendix}

\begin{figure}[ht!]
\begin{center}
\includegraphics[width=\textwidth]{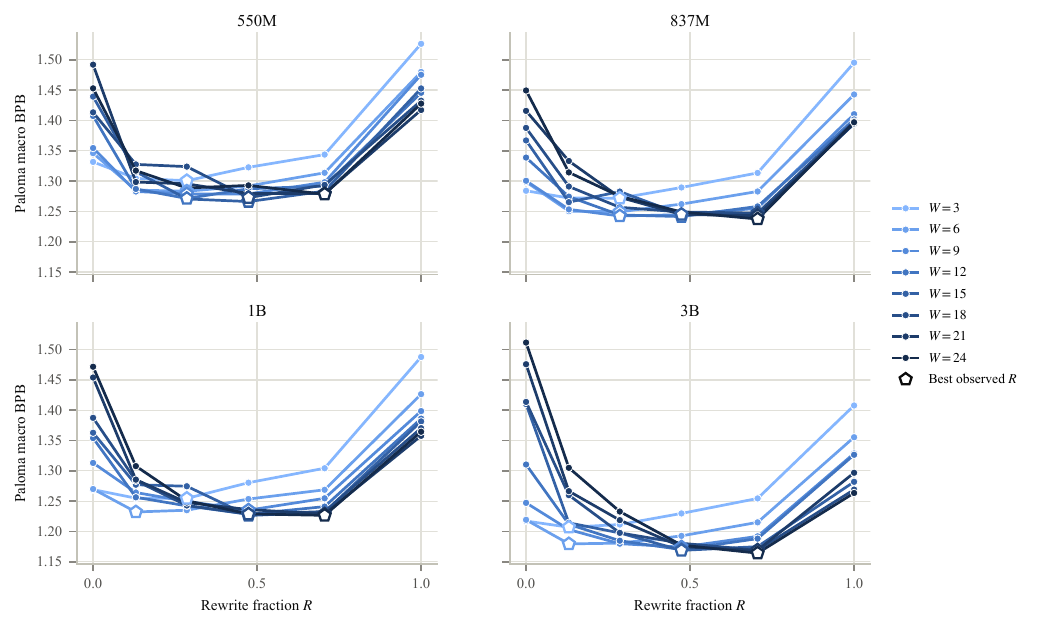}
\end{center}
\caption{Final-checkpoint Paloma macro BPB for all $192$ data-limited runs. Each panel shows one target size at all eight epoch budgets $W$ and six rewrite fractions $R$. Open pentagons mark the best observed mixture at each size and budget.}
\label{fig:de-ucurves-all}
\end{figure}

Figure~\ref{fig:de-ucurves-all} reports the full data-limited grid: four target sizes ($550$M, $837$M, $1$B, and $3$B parameters), eight epoch budgets $W\in\{3,6,9,12,15,18,21,24\}$, and six rewrite fractions $R\in\{0,0.13,0.29,0.47,0.71,1\}$. Table~\ref{tab:de-budgets} gives the total compute, accounting for rewrites, for every size and budget. The rewriter generates every rewrite token a run trains on, and no rewrite token is trained on twice, so the repetition penalty in Eq.~(\ref{eq:de_law}) applies to the natural seed set only. Figure~\ref{fig:de-ucurves} in the main text shows $W\in\{3,6,12,24\}$.

\begin{table}[t]
\caption{Total compute, accounting for rewrites, for the complete data-limited grid, in units of $10^{20}$ FLOPs. Each compute value has six rewrite-fraction runs.}
\label{tab:de-budgets}
\begin{center}
\small
\begin{tabular}{r|cccccccc}
\toprule
\multirow{2}{*}{Target size} & \multicolumn{8}{c}{Epoch budget $W$} \\
\cmidrule(lr){2-9}
 & $3$ & $6$ & $9$ & $12$ & $15$ & $18$ & $21$ & $24$ \\
\midrule
$550$M & $0.495$ & $0.991$ & $1.486$ & $1.981$ & $2.476$ & $2.972$ & $3.467$ & $3.962$ \\
$837$M & $0.753$ & $1.507$ & $2.260$ & $3.013$ & $3.766$ & $4.520$ & $5.273$ & $6.026$ \\
$1$B & $0.898$ & $1.796$ & $2.695$ & $3.593$ & $4.491$ & $5.389$ & $6.287$ & $7.185$ \\
$3$B & $2.606$ & $5.211$ & $7.817$ & $10.422$ & $13.028$ & $15.633$ & $18.239$ & $20.844$ \\
\bottomrule
\end{tabular}
\end{center}
\end{table}

\subsection{Domain Loss Across Target Sizes}
\label{data_eff_domains}
\begin{figure}[t]
\begin{center}
\includegraphics[width=\textwidth]{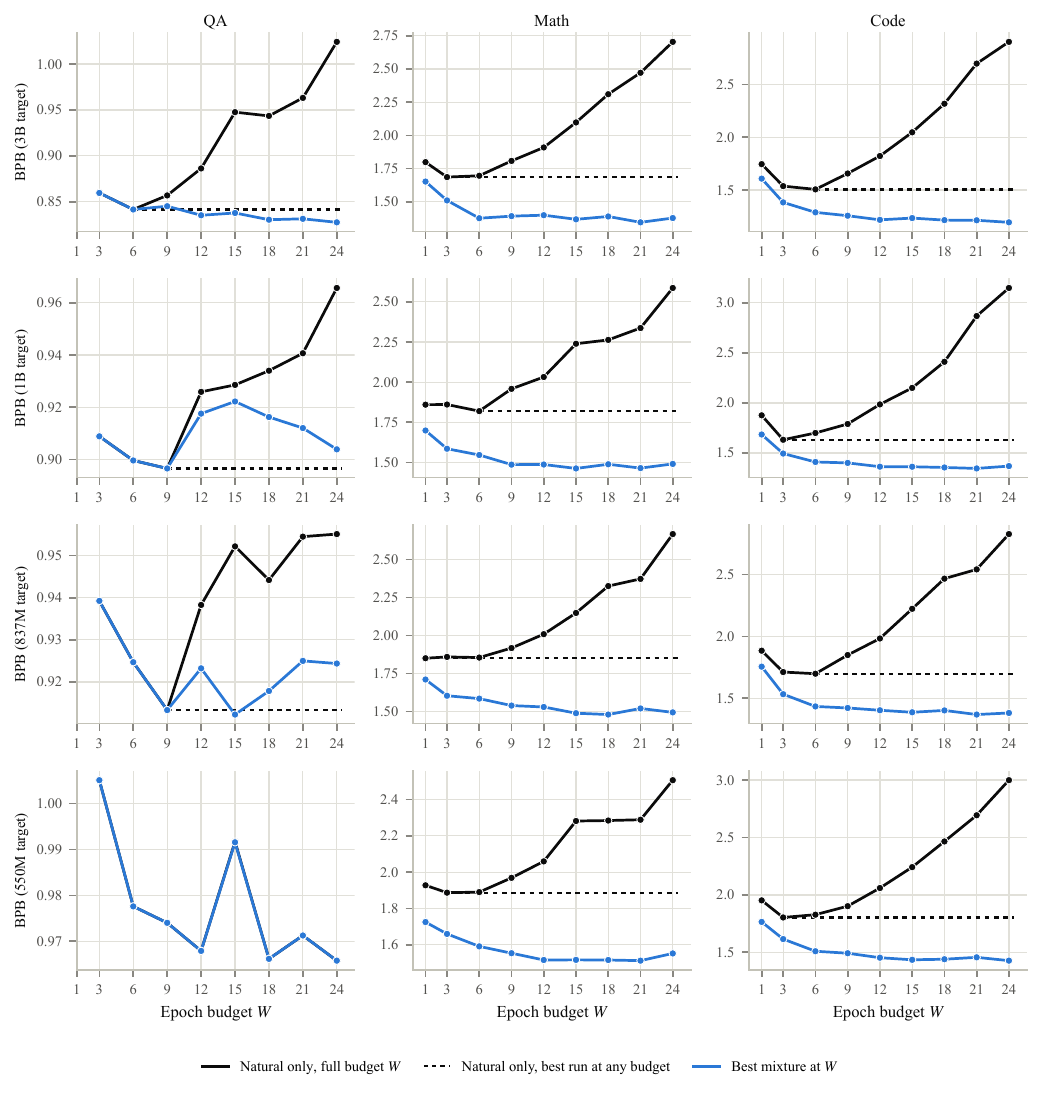}
\end{center}
\caption{Final-checkpoint BPB against the epoch budget $W$ for every target size (rows) on the QA, Math, and Code suites (columns). Series as in Figure~\ref{fig:de-ckpt}.}
\label{fig:de-domains-all}
\end{figure}

Figure~\ref{fig:de-domains-all} repeats Figure~\ref{fig:de-domains} for every target size. The best observed rewrite fraction on QA is $0$ at $W\le6$ for the $3$B target, at $W\le9$ for the $1$B and $837$M targets, and at every budget but one for the $550$M target. On Math and Code it is interior at every size from $W=3$. The QA, Math, and Code values come from our evaluation harness (Appendix~\ref{evals}).

\subsection{Best-Checkpoint Evaluation}
\label{data_eff_best_checkpoint}

Section~\ref{sec:data_eff_exp} reports final checkpoints. The best-checkpoint analysis instead selects, for each run, the saved checkpoint with the lowest Paloma validation loss whenever it improves on the final checkpoint by at least $0.01$ nats, and evaluates Paloma, QA, Math, and Code at that checkpoint. Paloma alone determines the selection. Selection changes the best observed rewrite fraction on Paloma in $4$ of the $32$ size--budget settings.

\begin{figure}[ht!]
\begin{center}
\includegraphics[width=\textwidth]{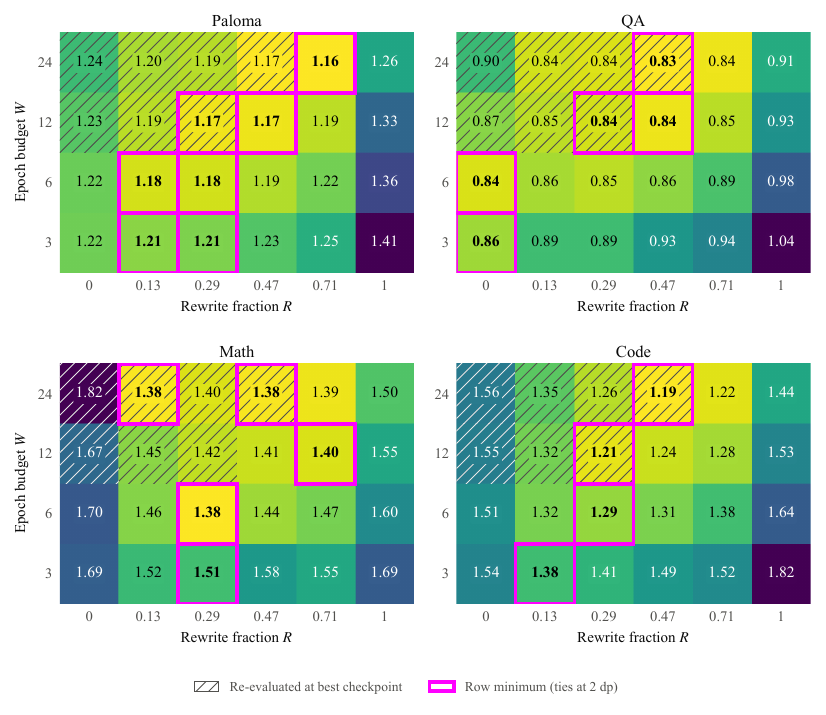}
\end{center}
\caption{Best-checkpoint BPB for the $3$B target by epoch budget $W$ and rewrite fraction $R$. Hatching marks runs re-evaluated at an earlier checkpoint whose Paloma validation loss is at least $0.01$ nats below the final checkpoint. Magenta outlines mark row minima, including ties after rounding to two decimals. Paloma alone selects each checkpoint.}
\label{fig:de-best-checkpoint-evals}
\end{figure}

Figure~\ref{fig:de-best-checkpoint-evals} shows the $3$B target at $W\in\{3,6,12,24\}$. At $W=24$, selection lowers natural-only Paloma BPB from $1.512$ to $1.241$, below pure rewriting at $1.264$. The lowest Paloma BPB in the figure is at $W=24$ and $R=0.71$. QA and Code reach their minima at $W=24$ and $R=0.47$, and Math ties at $R=0.13$ and $R=0.47$ after rounding to two decimals.

\section{Mixture Construction}
\label{mixture_construction}
A mixture at rewrite fraction $R$ draws its natural tokens from the filtered FineWeb-Edu documents (Appendix~\ref{fwedu_data}) and its rewritten tokens from the FinePhrase rewrites of those documents, pooled across the math, FAQ, table, and tutorial styles and sampled by tokens.
We train with Levanter \citep{levanter} and build every mixture with its mixture dataset. The natural corpus and the rewrite pool enter as two components with weights $1-R$ and $R$. Levanter partitions the training stream into fixed-size blocks and assigns the examples of each block to the two components in proportion to their weights, so every block-aligned prefix of the stream has rewrite fraction $R$ up to rounding. The assignment is a deterministic function of the seed, so runs at the same $R$ see the same interleaving.

\section{Generating Rewrites}
\label{generation}

\paragraph{Compute-limited generation.}
SmolLM2-1.7B generates the FinePhrase rewrites. We account for each rewritten training token as one input pass over its source document and one output pass over the rewrite, $4N_r$ FLOPs with $N_r=1.71$B (Eq.~\ref{eq:charged_compute}).

\paragraph{Data-limited generation.}

Each target size uses a rewriter of the same size, pretrained on the seed set and instruction-tuned at sequence length $8192$ on GLM-5.2 rewrites of $1000$ seed-set documents under the four FinePhrase prompts ($4000$ examples). The rewriter then rewrites the seed set under the same prompts at temperature $1.0$. Appendix~\ref{sec:data_eff_method} converts the generated volume to the rewrite fraction $R$.

We generate the synthetic pool before target-model training by repeatedly sampling fresh rewrites of the same seed documents under the same four FinePhrase prompts at temperature $1.0$. The pool remains fixed during training: we do not regenerate rewrites between training epochs. Each synthetic token used by a run is seen once, while natural tokens may be repeated.

\section{Rewrite-Style and Line-Content Ablations}
\label{rewrite_faithful}

We sample $6{,}000$ documents from each FinePhrase variant and classify every line as prose, list, table, header, or code (Table~\ref{tab:rewrite-line-content}). Table rewrites are $59\%$ table lines. Tutorial rewrites have the largest code share, $0.80\%$ of lines. This classification measures surface form and not faithfulness to the source document.

\begin{table}[t]
\caption{Line composition of $6{,}000$ documents per FinePhrase variant, in percent of classified lines.}
\label{tab:rewrite-line-content}
\begin{center}
\small
\begin{tabular}{r|ccccc}
\toprule
Variant & Prose & List & Table & Header & Code \\
\midrule
Natural  & $84.41$ & $14.32$ & $1.26$  & $0.01$ & $0.00$ \\
FAQ      & $64.69$ & $33.01$ & $0.50$  & $1.74$ & $0.06$ \\
Math     & $69.02$ & $30.06$ & $0.19$  & $0.28$ & $0.45$ \\
Table    & $29.21$ & $10.07$ & $58.99$ & $1.68$ & $0.05$ \\
Tutorial & $49.16$ & $45.34$ & $0.38$  & $4.31$ & $0.80$ \\
\bottomrule
\end{tabular}
\end{center}
\end{table}

\paragraph{Style contrasts.}
The content ablation trains one run per arm and rewrite fraction, for $20$ arms and $10$ fractions. Four style-only arms keep one rewrite style, and four style-out arms remove one style. FAQ style-only has higher Paloma BPB than FAQ style-out at all $10$ fractions, by $0.0226$ BPB on average. Math style-only is higher at $7$ of $10$ fractions ($+0.0032$ BPB on average). Table style-only is lower at all $10$ fractions ($-0.0039$ BPB), and tutorial style-only is lower at $8$ of $10$ ($-0.0038$ BPB).

\paragraph{Line-type contrasts.}
Five arms remove one line type from every rewrite: code, headers, lists, prose, or tables. We compare each against the full mixture at the nine fractions below $R=0.9$ and exclude the full-mixture run at $R=0.9$. Every removal raises mean BPB, by $0.0120$ to $0.0255$ BPB, with the largest increase from removing prose.

\section{Rewriting Scaling Law Details}
\label{scaling_law_details}

This section collects the derivations and the per-setting instantiations of the rewriting scaling law (Section~\ref{sec:scaling_model}), Eq.~(\ref{eq:rewriting_scaling_law}).

\paragraph{Mixture-dependent loss floor.}
We fit $L$ in bits per byte; every statement below holds in any logarithmic base, with $\lambda_n$ and $\lambda_s$ in Eq.~(\ref{eq:floor}) rescaled accordingly.
$E(R)$ is the irreducible loss of the mixture: the cross-entropy that a model with unlimited capacity and data would reach on the evaluation distribution when trained on a mixture of $(1-R)$ natural and $R$ rewritten data.
We write $p_e$ for the evaluation distribution, $p_n$ and $p_s$ for the natural and rewrite distributions, and $p_R = (1-R)\,p_n + R\,p_s$ for the training mixture's distribution. A model with unlimited capacity and data converges to $p_R$, so
\begin{equation}
E(R) = H(p_e) + \mathrm{KL}\!\left(p_e \,\|\, p_R\right).
\label{eq:floor_identity}
\end{equation}
Note that only the KL divergence depends on $R$. Because $\mathrm{KL}(p\,\|\,q)$ is convex in $q$ and $p_R$ is linear in $R$, $E(R)$ is convex in $R$ for every evaluation distribution. Evaluated on natural text ($p_e = p_n$) it is minimized at $R = 0$ and increases with $R$; evaluated on the rewrites themselves ($p_e = p_s$) the divergence falls to zero as $R \to 1$; evaluated on a distribution that neither source matches exactly, such as Paloma, the minimizer can be interior.
A two-component approximation to the divergence (Appendix~\ref{scaling_er_derivation}) gives the parametric form
\begin{equation}
E(R) = E_0 - \lambda_n \log\!\left(1-(1-\rho_n)R\right)
-\lambda_s \log\!\left(\rho_s+(1-\rho_s)R\right). 
\label{eq:floor}
\end{equation}
Appendix~\ref{scaling_er_derivation} derives Eq.~(\ref{eq:floor}) and interprets $\lambda_n,\rho_n,\lambda_s,\rho_s$ as the share of the evaluation distribution each source covers better and the other source's relative density there.
The first term is the cost of \emph{removing} natural text and the second the benefit of \emph{adding} rewrites. The minimizer $R^*_\infty = \argmin_R E(R)$, the floor-optimal rewrite fraction, is unique because Eq.~(\ref{eq:floor}) is strictly convex for positive $\lambda_n,\lambda_s$, and is interior when $E'(0)<0<E'(1)$ (Appendix~\ref{scaling_er_derivation}).

\paragraph{Repetition penalty.}
$F(R,D,T) = P\,(1 - R)^{s}\,\bigl((1-R)D/T - 1\bigr)_+$ is the repetition penalty of \citet{lovelace2026prescriptivescalinglawsdata}, diluted by rewrites at rate $s$. It is zero when $(1-R)D \le T$, as in every compute-limited run. 

\paragraph{Rewrites shift the floor only.}
$R$ enters Eq.~(\ref{eq:rewriting_scaling_law}) through $E(R)$ and, when $(1-R)D > T$, through the dilution factor $(1-R)^s$ and the natural epochs $(1-R)D/T$ of the repetition penalty; it does not enter the capacity and data terms. This is the central modeling assumption of the paper: rewrites change \emph{what} the model converges to, not \emph{how fast} it gets there in $N$ and $D$. It implies that the best loss at any training budget decomposes as
\begin{equation}
L^*(C, R) = L^*_{\mathrm{base}}(C) + \tilde{g}(R), \qquad \tilde{g}(R) = E(R) - E(0).
\label{eq:floor_shift}
\end{equation}
Here, $L^*(C, R) = \min_{6ND = C} L(N, D, R, T\to\infty)$ is the best loss over $(N, D)$ at training budget $C$ and $L^*_{\mathrm{base}}(C) \equiv L^*(C, 0)$. Figure~\ref{fig:frontier_by_r} (Appendix~\ref{scaling}) tests how closely $N^*(C)$ and $D^*(C)$ remain stable across $R$.

\paragraph{Accounting for rewrites.}
Section~\ref{sec:scaling_model} gives the generation cost of $4N_r$ FLOPs per rewritten training token for a rewriter with $N_r$ parameters, so
\begin{equation}
C_{\mathrm{total}}
=
C + 4N_r R D,
\qquad
C=6ND.
\label{eq:charged_compute}
\end{equation}
We do not account for the cost of training the rewriter as it is amortized across all experiments.

\subsection{Compute-Limited Setting}
\label{sec:compute_eff_method}

\paragraph{Setup.}
The compute-limited setting reflects the non-frontier case in which natural data is plentiful and our goal is to characterize the strongest possible model subject to a fixed compute budget.
We use an external rewriter: one that we do not train ourselves, is stronger than any model that we train, and is trained with more compute than our largest training run, and we do not account for its training cost.
A mixture with rewrite fraction $R$ draws a fraction $R$ of its training tokens from rewrites and the remaining $1-R$ from natural text.
We fit the law to IsoFLOP sweeps: at each training budget $C$ we train one grid of model sizes $N$ with $D = C/(6N)$ tokens, and every rewrite fraction $R$ uses the same grid.

\paragraph{Fitting the surface.}
We fit Eq.~(\ref{eq:rewriting_scaling_law}) with $F=0$ to every run, since $(1-R)D \le T$ for every compute-limited run, to obtain the floor $E(R)$, subtract the fitted shift $\tilde g(R)$ from every run, and fit the $(N,D)$ backbone to the pooled runs with Chinchilla Approach 2 \citep{hoffmann2022trainingcomputeoptimallargelanguage}. A quadratic in $\log D$ at each budget gives $D^*(C)$ and $L^*_{\mathrm{base}}(C)$, and power laws across budgets give $\alpha$ and $\beta$. We add an off-frontier curvature term that vanishes with zero slope on the frontier and leaves $N^*(C)$, $D^*(C)$, and $L^*_{\mathrm{base}}(C)$ unchanged. Appendix~\ref{scaling_fit} gives the exponent conversion, the fitted parameters, the leave-one-budget-out forecasts, and the curvature ablation.
\paragraph{Compute-optimal rewrite fraction.}
Under Eq.~(\ref{eq:floor_shift}) the floor-optimal fraction $R^*_\infty$ minimizes loss at any \emph{training} budget. Accounting for generation changes the answer: at a fixed total budget, every rewritten token is paid for twice, once to generate and once to train on, so raising $R$ shrinks the training budget available. Because natural data is plentiful in this setting, we assume an unlimited supply of natural tokens and define $L(N, D, R) = \lim_{T\to\infty} L(N, D, R, T)$. As $T\to\infty$ the repetition penalty $F(R, D, T)$ vanishes. The compute-optimal allocation minimizes this surface jointly over model size, token count, and rewrite fraction at a fixed total budget:
\begin{equation}
\bigl(N^*, D^*, \Rhat\bigr)
=
\argmin_{N,\, D,\, R \in [0, 1]}
L(N, D, R)
\quad
\text{s.t.}
\quad
6ND + 4N_r R D = C_{\mathrm{total}},
\label{eq:rstar}
\end{equation}
where we evaluate $L(N,D,R)$ with the off-frontier curvature term of Eq.~(\ref{eq:curv}) added, as in the fitted surface. Because each rewritten token also costs generation compute, the optimum with $R>0$ trains a slightly larger model on fewer tokens than the natural-only allocation at the same training compute. The generation cost is linear in $D$ while the training loss improves as a power law in compute, so the overhead of rewriting becomes relatively cheaper as budgets grow, and $\Rhat(C_{\mathrm{total}})$ increases from near $0$ toward $\Rhat_\infty$ (Figure~\ref{fig:rstar_skills}).

\subsection{Data-Limited Setting}
\label{sec:data_eff_method}

\paragraph{Setup.}

The data-limited setting reflects the frontier case in which no stronger rewriter exists, we target a fixed model size, and the amount of natural data is fixed, but compute is the plentiful resource. A fixed seed set of $T$ natural tokens and a target model with $N$ parameters are given, and an internal rewriter of the same size, pretrained on the seed set, generates the rewrites. We do not account for the rewriter's training cost, which is shared across all data-limited experiments at that size. As above, budgets are marginal compute given the rewriter. As in the main paper, we measure the total compute budget by $W=C_{\mathrm{total}}/(6NT)$. A run with $D$ training tokens and rewrite fraction $R$ trains on $(1-R)D$ natural tokens and $RD$ synthetic tokens, each synthetic token seen once. Training and generation together cost $6ND+4NRD=6NTW$, giving $D=3TW/(3+2R)$. The run therefore passes over the natural seed set $3W(1-R)/(3+2R)$ times.

\paragraph{Data-limited scaling model.}

With the target size fixed at $N_0$, the capacity term $A/N_0^{\alpha}$ in Eq.~(\ref{eq:rewriting_scaling_law}) is a constant and is absorbed into the floor. At epoch budget $W$ and rewrite fraction $R$, the run sees $D = 3TW/(3+2R)$ tokens, counting natural and synthetic tokens alike: $3W(1-R)/(3+2R)$ epochs over the seed set and the rest fresh rewrites, none repeated. Substituting into Eq.~(\ref{eq:rewriting_scaling_law}) gives
\begin{equation}
L_{\mathrm{DE}}(R,W)=E(R)+\frac{B}{\left(3TW/(3+2R)\right)^{\beta}}+P\,(1-R)^{s}\left(\frac{3W(1-R)}{3+2R}-1\right)_+ .
\label{eq:de_law}
\end{equation}
At $R=0$ the law reduces to the epoching law of \citet{lovelace2026prescriptivescalinglawsdata}. At $R=1$ the repetition penalty vanishes. In between, rewrites add fresh tokens and dilute repetition of the seed set. We fit the law separately for each target--rewriter pair, pooling all epoch budgets and rewrite fractions, because the size-matched rewriter changes the quality of the synthetic data with the target size (Figure~\ref{fig:de-capacity}). The minimizer of the fitted law over $R$ is $\Rhat(W)$.

\section{Dependence of Efficient-Frontier Allocation on \texorpdfstring{$R$}{R}}
\label{indep_R_N_D}

The shared surface fits one compute-optimal token count $D^*(C)$ across rewrite fractions. We test this restriction on the $583$ runs of the five budgets with all eleven rewrite fractions. Both fits give every budget--fraction cell its own loss level. The restricted fit shares $D^*(C)$ across fractions. The alternative adds ten fraction-specific offsets to $\log_{10}D^*$, with $R=0$ as the reference.

The offsets lower RMSE from $0.00831$ to $0.00762$ BPB, and the largest shift is $D^*(R)/D^*(0)=1.196$ at $R=0.8$. A wild bootstrap that resamples residual signs jointly within each architecture ($400$ draws) rejects equal $D^*$ across fractions ($p=1/401$). Figure~\ref{fig:frontier_by_r} shows that a shift of this size does not change the selected model size, and Appendix~\ref{scaling_setup} compares richer surface specifications on the same runs.

\section{Compute-Limited Experimental Results}
\label{compute_eff_appendix}

\subsection{Scaling Model Details}
\label{scaling}

The FineWeb-Edu ladders contain five training budgets from $3\times10^{18}$ to $9\times10^{19}$ FLOPs at eleven rewrite fractions $R\in\{0,0.1,\ldots,1\}$, and natural-only ($R=0$) ladders at $1.8\times10^{20}$ and $3\times10^{20}$ FLOPs. The surface fit uses these $604$ Paloma BPB observations and the $10^{22}$-FLOP hero run ($R=0.321$) as a frontier anchor. Nonzero-$R$ runs at $1.8\times10^{20}$ and $3\times10^{20}$ FLOPs, the $10^{21}$ runs, and the remaining $10^{22}$ runs are validation runs outside the fit. The forward forecasts in Appendix~\ref{scaling_fit} refit the surface without each held-out budget.

\begin{figure}[p]
\begin{center}
\includegraphics[width=\textwidth]{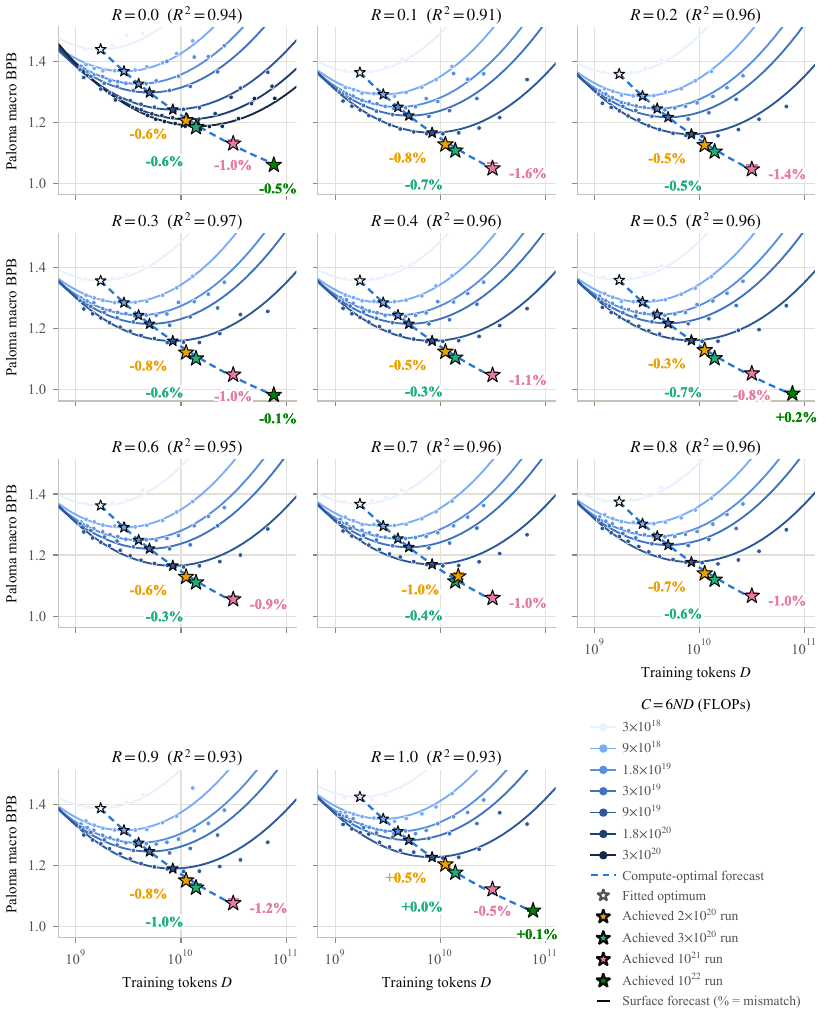}
\end{center}
\caption{The shared $L(N,D,R)$ surface as per-$R$ IsoFLOP slices at all eleven rewrite fractions, in Paloma macro BPB. Dots mark runs, open stars mark fitted frontier optima, and filled stars mark the $10^{21}$ and $10^{22}$ runs. Labels report measured minus fitted loss as a percentage of fitted loss.}
\label{fig:isoflop_surface_appendix}
\end{figure}

\begin{figure}[t]
\begin{center}
\includegraphics[width=\textwidth]{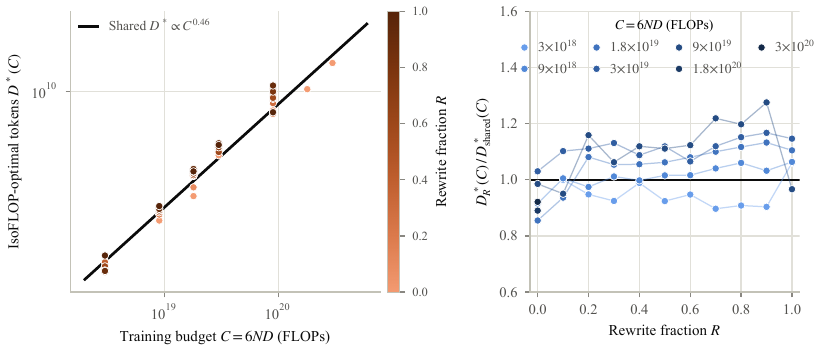}
\end{center}
\caption{Per-$R$ IsoFLOP optima against the shared frontier. \textbf{Left:} compute-optimal token count $D^*_R(C)$ from a separate Approach-2 parabola for each rewrite fraction and budget, colored by $R$, with the shared frontier $D^*(C)\propto C^{0.46}$. \textbf{Right:} ratio of each per-$R$ optimum to the shared frontier, by budget.}
\label{fig:frontier_by_r}
\end{figure}

\paragraph{The frontier location does not depend on $R$.}
The surface fit pools all rewrite fractions before locating the IsoFLOP optima (Appendix~\ref{sec:compute_eff_method}), so the fitted $N^*(C)$ and $D^*(C)$ cannot vary with $R$ by construction. Figure~\ref{fig:frontier_by_r} tests the assumption by fitting a separate Approach-2 parabola in $\log D$ to each $(R,C)$ sweep and taking its vertex. The $56$ per-$R$ optima over seven budgets lie within $0.85$--$1.28\times$ of the shared $D^*(C)$ (RMS $10\%$), and per-$R$ power laws give token exponents between $0.45$ and $0.57$ against the shared $0.46$. The spread across $R$ at a fixed budget matches the spread of the $R=0$ ladder across budgets ($0.85$--$1.03\times$).

\begin{figure}[t]
\begin{center}
\includegraphics[width=\textwidth]{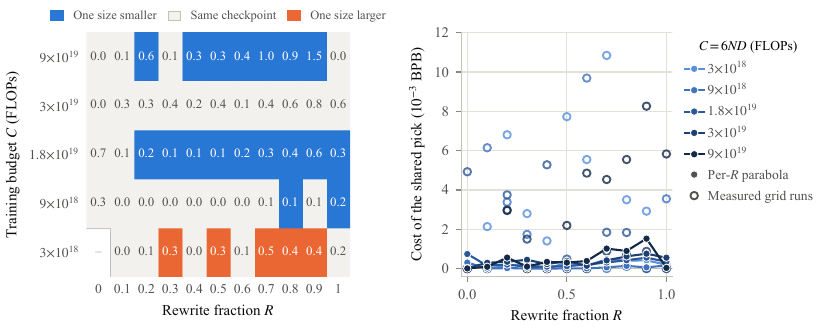}
\end{center}
\caption{Whether fitting the frontier per $R$ changes the selected model size. \textbf{Left:} for each sweep with a full rewrite ladder, the grid size nearest the per-$R$ vertex relative to the one nearest the shared $N^*(C)$. Cells report the BPB cost of the shared choice on that sweep's own IsoFLOP parabola. \textbf{Right:} the same cost by rewrite fraction, on the parabola (filled) and between the two measured grid runs (open).}
\label{fig:checkpoint_by_r}
\end{figure}

Figure~\ref{fig:checkpoint_by_r} asks whether this spread changes the selected model size. The grid steps by $1.2$--$1.4\times$ in $N$. In $23$ of $54$ sweeps the per-$R$ vertex selects a neighboring grid size, never one two steps away. Most of these cases are at $1.8\times10^{19}$ and $9\times10^{19}$ FLOPs, where the shared $N^*$ falls midway between two grid sizes. On each sweep's own parabola, the shared choice costs at most $0.0015$ BPB ($0.0002$ BPB at the median), against the $0.082$-BPB floor gain at $\Rhat_\infty$. The measured gap between the two grid runs (median $0.0015$, maximum $0.0108$ BPB) matches the parabola residuals and does not grow with $R$.

\subsubsection{Do Rewrites Still Help in the Infinite-Compute Limit?}
\label{infinite_compute}

\begin{figure}[t]
\begin{minipage}[c]{0.58\textwidth}
\centering
\includegraphics[width=\linewidth]{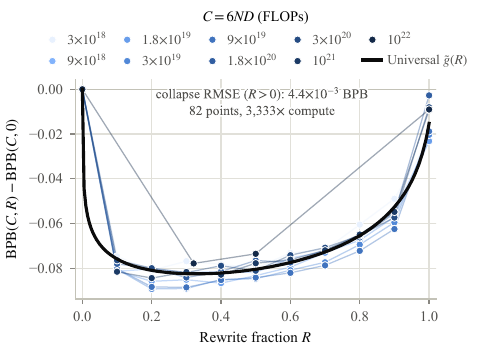}
\end{minipage}\hfill
\begin{minipage}[c]{0.39\textwidth}
\raggedright
\caption{\textbf{Rewrite gains collapse onto one curve across $3{,}333\times$ compute.} IsoFLOP-optimal Paloma BPB after subtracting each budget's $R=0$ value. The fitted $\tilde g(R)$ has RMSE $0.0044$ BPB over $82$ observations from nine budgets.}
\label{fig:collapse}
\end{minipage}
\end{figure}

Figure~\ref{fig:collapse} shows the measured evidence that rewrites shift the floor independently of scale. After subtracting each budget's $R=0$ value, $82$ IsoFLOP-optimal observations from nine budgets collapse onto one curve $\tilde g(R)$ with RMSE $0.0044$ BPB.

\begin{figure}[t]
\begin{center}
\includegraphics[width=0.4\textwidth]{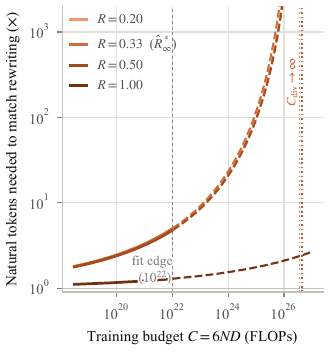}
\end{center}
\caption{Natural tokens required to match the estimated floor-optimal mixture, as a function of training compute. The final fit gives $5.01\times$ at $10^{22}$ FLOPs and diverges at $C_{\mathrm{div}}=3.61\times10^{26}$ FLOPs. Dashed curves exceed the fit range. The $671$-run fit gives $12.17\times$ and $1.75\times10^{24}$ FLOPs.}
\label{fig:data_multiplier}
\end{figure}

The final fit gives $\Rhat_{\infty}=0.328$ and a floor shift of $-0.0825$ BPB. Extrapolating both frontiers, a natural-only model at $10^{22}$ training FLOPs needs $5.01\times$ as many tokens to match the estimated floor-optimal mixture, and the natural-only frontier stops reaching the mixture frontier at $3.61\times10^{26}$ FLOPs (Figure~\ref{fig:data_multiplier}). These magnitudes depend on the fit specification. An independent fit to all $671$ runs, including the validation runs but holding three $10^{22}$-FLOP runs out, gives a $12.17\times$ multiplier at $10^{22}$ FLOPs with $95\%$ bootstrap interval $[6.39,45.01]$ and divergence at $1.75\times10^{24}$ FLOPs. No run exceeds $10^{22}$ FLOPs, so both multipliers are extrapolations.

\subsubsection{\texorpdfstring{$E(R)$}{E(R)} Derivation}
\label{scaling_er_derivation}

Let $p_e$ denote the evaluation distribution, $p_n$ the natural training distribution, and $p_s$ the rewrite distribution. The mixture distribution is
\begin{align}
p_R(x)=(1-R)p_n(x)+Rp_s(x).
\label{eq:mixture_distribution}
\end{align}
Partition the evaluation support into a natural-favored region $\mathcal A$ and a rewrite-favored region $\mathcal B$. On $\mathcal A$, approximate $p_s(x)=\rho_n p_n(x)$ with $\rho_n<1$. On $\mathcal B$, approximate $p_n(x)=\rho_s p_s(x)$ with $\rho_s<1$. Let $\lambda_n$ and $\lambda_s$ denote the evaluation mass of the two regions in the chosen logarithmic unit.

Equation~\ref{eq:mixture_distribution} then becomes
\begin{align}
p_R(x)&=p_n(x)\left[1-(1-\rho_n)R\right], &&x\in\mathcal A,\\
p_R(x)&=p_s(x)\left[\rho_s+(1-\rho_s)R\right], &&x\in\mathcal B.
\end{align}
Substituting into the evaluation cross-entropy $-\mathbb E_{x\sim p_e}\log p_R(x)$ and absorbing terms independent of $R$ into $E_0$ gives
\begin{align}
E(R)=E_0
-\lambda_n\log\left[1-(1-\rho_n)R\right]
-\lambda_s\log\left[\rho_s+(1-\rho_s)R\right].
\label{eq:mixture-kl-floor}
\end{align}
The derivatives are
\begin{align}
E'(R)
&=\frac{\lambda_n(1-\rho_n)}{1-(1-\rho_n)R}
-\frac{\lambda_s(1-\rho_s)}{\rho_s+(1-\rho_s)R},\\
E''(R)
&=\frac{\lambda_n(1-\rho_n)^2}{[1-(1-\rho_n)R]^2}
+\frac{\lambda_s(1-\rho_s)^2}{[\rho_s+(1-\rho_s)R]^2}>0.
\end{align}
The floor is convex, with unconstrained minimizer
\begin{align}
R^*_{\infty}
=\frac{\lambda_s(1-\rho_s)-\lambda_n(1-\rho_n)\rho_s}
{(1-\rho_n)(1-\rho_s)(\lambda_n+\lambda_s)}.
\label{eq:floor_optimum_closed_form}
\end{align}
The constrained solution clips Eq.~\ref{eq:floor_optimum_closed_form} to $[0,1]$.

\subsubsection{Chinchilla Approach 2 \texorpdfstring{$\to$}{to} Approach 3 Conversion}
\label{scaling_conversion}

Approach 2 fits the compute-optimal token and loss frontiers
\begin{align}
D^*(C)&=a_D C^p,\\
L^*(C)&=E+a_L C^{-\gamma}.
\end{align}
With $C=6ND$, the corresponding model frontier is
\begin{align}
N^*(C)=\frac{1}{6a_D}C^{1-p}.
\end{align}
For an Approach-3 surface $L(N,D)=E+A N^{-\alpha}+B D^{-\beta}$, both reducible-loss terms share the frontier exponent. Matching exponents gives
\begin{align}
\alpha=\frac{\gamma}{1-p},\qquad
\beta=\frac{\gamma}{p}.
\end{align}
The first-order condition gives the capacity and data terms shares $(1-p)$ and $p$ of the frontier reducible loss, so
\begin{align}
A&=(1-p)a_L\left(\frac{1}{6a_D}\right)^{\alpha},\\
B&=p\,a_L a_D^{\beta}.
\end{align}
For the final FineWeb-Edu fit, $a_D=4.5033$, $p=0.46447$, $a_L=67.2942$, and $\gamma=0.10963$, giving $\alpha=0.20472$ and $\beta=0.23604$.

\subsubsection{Fitting \texorpdfstring{$L(N, D, R, T)$}{L(N, D, R, T)}}
\label{scaling_fit}
\paragraph{Data-limited case.}
Each target--rewriter pair contributes $48$ runs: eight epoch budgets by six rewrite fractions. We fit Eq.~(\ref{eq:de_law}) to final-checkpoint Paloma BPB separately for each pair, because the size-matched rewriter changes the quality of the synthetic data with the target size (Figure~\ref{fig:de-capacity}). Each fit pools all epoch budgets and rewrite fractions and uses nine parameters: the five floor parameters, $B$, $\beta$, $P$, and $s$ (Table~\ref{tab:de-fit}). We minimize the unweighted sum of squared residuals with bounded least squares ($\rho_n,\rho_s\in[10^{-4},1]$, $\beta\in[0.05,3]$, $s\in[0,8]$) from six fixed starts and $60$ seeded random starts, keeping the lowest-cost solution. Table~\ref{tab:de-fit} reports point estimates only; the fit script and its cached outputs are released with the code.

\begin{table}[t]
\caption{Data-limited fits of Eq.~(\ref{eq:de_law}) to final-checkpoint Paloma BPB, one fit per target--rewriter pair, $48$ runs per row. RMSE, $E_0$, $\lambda_n$, $\lambda_s$, and $P$ are in BPB; $B$ is in BPB$\cdot$tokens$^{\beta}$, with $D$ in tokens as in Eq.~(\ref{eq:de_law}); $\rho_n$, $\rho_s$, $\beta$, and $s$ are dimensionless. $\rho_s$ sits at its $10^{-4}$ lower bound except for the $1$B pair.}
\label{tab:de-fit}
\begin{center}
\small\setlength{\tabcolsep}{4pt}
\begin{tabular}{r|c|ccccccccc}
\toprule
Target & RMSE & $E_0$ & $\lambda_n$ & $\lambda_s$ & $\rho_n$ & $\rho_s$ & $B$ & $\beta$ & $P$ & $s$ \\
\midrule
$550$M & $0.0147$ & $1.259$ & $0.0220$ & $0.0045$ & $0.0002$ & $0.0001$ & $1.55\times10^{18}$ & $1.939$ & $0.0079$ & $6.990$ \\
$837$M & $0.0085$ & $1.170$ & $0.0458$ & $0.0045$ & $0.0105$ & $0.0001$ & $4.17\times10^{8}$ & $0.963$ & $0.0096$ & $1.539$ \\
$1$B & $0.0115$ & $1.209$ & $0.0194$ & $0.0000$ & $0.0002$ & $0.0249$ & $1.86\times10^{15}$ & $1.636$ & $0.0113$ & $5.075$ \\
$3$B & $0.0150$ & $1.040$ & $0.0823$ & $0.0041$ & $0.0872$ & $0.0001$ & $4.44\times10^{5}$ & $0.647$ & $0.0174$ & $2.082$ \\
\bottomrule
\end{tabular}
\end{center}
\end{table}

Paloma BPB is U-shaped in rewrite fraction. The fitted optimum increases with epoch budget within every target--rewriter pair. The repetition coefficient $P$ increases across pairs. The best-mixture asymptote is no higher than the natural-only asymptote at every pair and lies strictly below it at three; the $1$B fit places the two at parity and attributes its rising $\Rhat(W)$ to the repetition penalty. The fit does not isolate a target-size effect, because the size-matched rewriter changes the quality of the synthetic data with the target size.

\begin{figure}[t]
\begin{center}
\includegraphics[width=\textwidth]{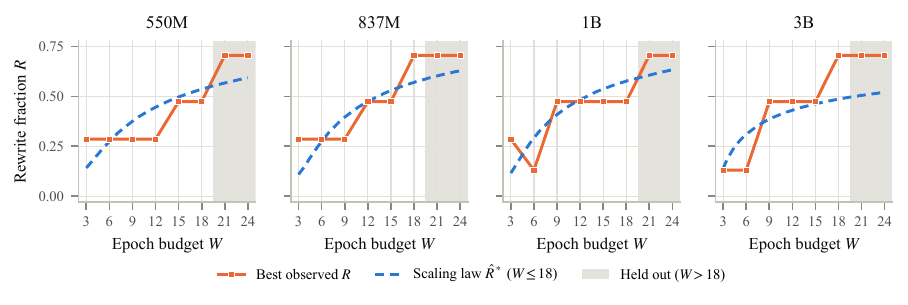}
\end{center}
\caption{Estimated optimal rewrite fraction $\Rhat$ and best observed $R$ as a function of epoch budget $W$ for each target--rewriter pair. Orange squares mark the best observed $R$ on the tested grid. Dashed blue curves show $\Rhat$ from each pair's scaling-law fit to $W\le18$, and the shaded budgets $W=21$ and $W=24$ are held out. $\Rhat$ increases with budget within every pair. Across pairs, the rewriter changes with the target size.}
\label{fig:de-rstar}
\end{figure}

Out of sample, we refit Eq.~(\ref{eq:de_law}) for each target--rewriter pair to the $36$ runs at $W\le18$ and forecast the six rewrite fractions at each of the held-out budgets $W=21$ and $W=24$. Figure~\ref{fig:de-ucurves} reports the $R$ this fit selects at every budget and its measured cost relative to the best observed $R$. The mean error is within $\pm1.4\%$ at every pair and budget, and the measured loss at the best observed $R$, which is $R=0.71$ at every pair and both budgets, is within $0.74\%$ of its forecast (Table~\ref{tab:de-lobo}). The largest errors are at $R=0.13$ for the $550$M, $837$M, and $1$B pairs and at $R=0.29$ for the $3$B pair. The error is $(L_{\mathrm{measured}}-L_{\mathrm{predicted}})/L_{\mathrm{predicted}}$, as in Table~\ref{tab:ce-lobo}.

\begin{table}[t]
\caption{Forward data-limited forecasts. Each row fits the $36$ runs at $W\le18$ and forecasts the six rewrite fractions at the held-out budget $W$. RMSE is in BPB.}
\label{tab:de-lobo}
\begin{center}
\small
\begin{tabular}{r|cccccc}
\toprule
\multirow{2}{*}{Target} & Held-out & Fit & Held-out & Mean & Per-$R$ & Error at \\
 & $W$ & RMSE & RMSE & error & range & best obs.\ $R$ \\
\midrule
$550$M & $21$ & $0.0114$ & $0.0267$ & $-0.73\%$ & $[-3.98,+1.98]\%$ & $-0.28\%$ \\
$550$M & $24$ & $0.0114$ & $0.0294$ & $-1.39\%$ & $[-3.90,+0.85]\%$ & $-0.26\%$ \\
$837$M & $21$ & $0.0083$ & $0.0060$ & $+0.13\%$ & $[-0.29,+0.86]\%$ & $-0.27\%$ \\
$837$M & $24$ & $0.0083$ & $0.0140$ & $-0.42\%$ & $[-1.98,+0.79]\%$ & $-0.68\%$ \\
$1$B   & $21$ & $0.0102$ & $0.0225$ & $-0.50\%$ & $[-3.01,+2.05]\%$ & $+0.05\%$ \\
$1$B   & $24$ & $0.0102$ & $0.0212$ & $-0.81\%$ & $[-2.97,+0.91]\%$ & $+0.03\%$ \\
$3$B   & $21$ & $0.0143$ & $0.0163$ & $+0.27\%$ & $[-1.94,+2.14]\%$ & $-0.50\%$ \\
$3$B   & $24$ & $0.0143$ & $0.0201$ & $-0.04\%$ & $[-2.22,+2.65]\%$ & $-0.74\%$ \\
\bottomrule
\end{tabular}
\end{center}
\end{table}

\paragraph{Compute-limited case.}
The surface fits the $604$ observations in three stages: the mixture-KL floor, the pooled Approach-2 frontiers promoted through Appendix~\ref{scaling_conversion}, and the curvature term of Appendix~\ref{scaling_curv_ablation}. Table~\ref{tab:ce-fit} lists the parameters. The RMSE over all runs is $0.01449$ BPB. The promoted frontiers are
\begin{align}
D^*(C)&=4.5033\,C^{0.46447},\\
L^*(C,0)&=0.80569+67.2942\,C^{-0.10963},\\
N^*(C)&=\frac{C}{6D^*(C)},
\end{align}
which give $(N^*,D^*)=(0.554\text{B},3.009\text{B})$ at $10^{19}$ FLOPs and $(22.39\text{B},74.44\text{B})$ at $10^{22}$ FLOPs.

\begin{table}[t]
\caption{Final FineWeb-Edu surface parameters. $N$ and $D$ are absolute parameter and token counts, and $C=6ND$ is in FLOPs. The fitted $\rho_s$ reaches its lower bound. The fitted floor at $R=0$ is $E(0)=E_0-\lambda_s\log\rho_s=0.805691$ (Eq.~\ref{eq:floor}).}
\label{tab:ce-fit}
\begin{center}
\small
\begin{tabular}{cl@{\qquad}||cl}
\toprule
Parameter & Value & Parameter & Value \\
\midrule
$E_0$       & $0.701293$ & $A$      & $18.3516$ \\
$\lambda_n$ & $0.024066$ & $\alpha$ & $0.204717$ \\
$\rho_n$    & $0.024598$ & $B$      & $44.5853$ \\
$\lambda_s$ & $0.011335$ & $\beta$  & $0.236040$ \\
$\rho_s$    & $10^{-4}$  & $H$      & $3.00277$ \\
             &            & $\eta$   & $0.072262$ \\
\bottomrule
\end{tabular}
\end{center}
\end{table}

\paragraph{Surface fit.}
We fit the joint surface $L(N,D,R)$ to the $604$ runs of the IsoFLOP ladders, which include the $10^{22}$-FLOP hero run as the largest-budget point. 
In sample, the fit has RMSE $0.0145$ BPB ($R^2=0.95$) over all $604$ runs, a root-mean-square relative error of $1.1\%$. 
The residual concentrates in the over-trained tail: on the $296$ runs within $2\times$ of the fitted token optimum $D^*(C)$, the RMSE is $0.0096$ BPB ($R^2=0.98$) and the relative error $0.7\%$. 
The quantities we use downstream are the compute-optimal losses, the lowest measured loss at each of the $58$ pairs of budget and rewrite fraction. 
The fitted frontier reproduces these to within $0.75\%$ (median $0.19\%$). 
Stated in compute rather than loss, the largest of these errors is the amount by which the fitted compute-optimal loss changes when the training FLOPs change by $19\%$, and the median error is the amount from a $4\%$ change in training FLOPs. 
Figure~\ref{fig:isoflop_surface_appendix} shows the fitted ladders at all eleven rewrite fractions. 
Out of sample, we perform a forward leave-one-budget-out analysis: we fit the surface to the budgets below each held-out budget and forecast every $R$. 
These forecasts have mean errors within $\pm2\%$ at $2$--$3\times$ extrapolation, and forecasting the $10^{22}$ runs from a fit range ending at $3\times10^{20}$ FLOPs gives a mean error of $-1.0\%$, with every measured loss $0.9$--$1.4\%$ below its forecast (Table~\ref{tab:ce-lobo}).

Table~\ref{tab:ce-lobo} reports forward leave-one-budget-out forecasts. Each row fits only the budgets below the held-out budget and forecasts every rewrite fraction at that budget. The error is $(L_{\mathrm{measured}}-L_{\mathrm{predicted}})/L_{\mathrm{predicted}}$, so negative values are conservative forecasts.

\begin{table}[t]
\caption{Forward compute-limited forecasts by training budget. The $10^{22}$ row fits through $3\times10^{20}$ FLOPs and predicts the four rewrite fractions trained at the hero-run allocation ($R\in\{0,0.321,0.5,1\}$).}
\label{tab:ce-lobo}
\begin{center}
\small
\begin{tabular}{ccccc}
\toprule
Test training FLOPs & Largest fit training FLOPs & Extrap. & Mean error & Per-$R$ range \\
\midrule
$3\times10^{19}$   & $1.8\times10^{19}$ & $1.7\times$  & $-0.50\%$ & $[-0.81,-0.16]\%$ \\
$9\times10^{19}$   & $3\times10^{19}$   & $3.0\times$  & $-0.36\%$ & $[-0.77,+0.02]\%$ \\
$1.8\times10^{20}$ & $9\times10^{19}$   & $2.0\times$  & $-0.21\%$ & $[-0.56,+0.80]\%$ \\
$3\times10^{20}$   & $1.8\times10^{20}$ & $1.7\times$  & $-0.70\%$ & $[-1.18,-0.13]\%$ \\
$10^{21}$           & $3\times10^{20}$   & $3.3\times$  & $-1.32\%$ & $[-1.87,-0.73]\%$ \\
$10^{22}$           & $3\times10^{20}$   & $33.3\times$ & $-1.04\%$ & $[-1.40,-0.85]\%$ \\
\bottomrule
\end{tabular}
\end{center}
\end{table}

\subsubsection{Repetition-Penalty Form Ablation}
\label{de_penalty_ablation}

\begin{figure}[t]
\begin{center}
\includegraphics[width=\textwidth]{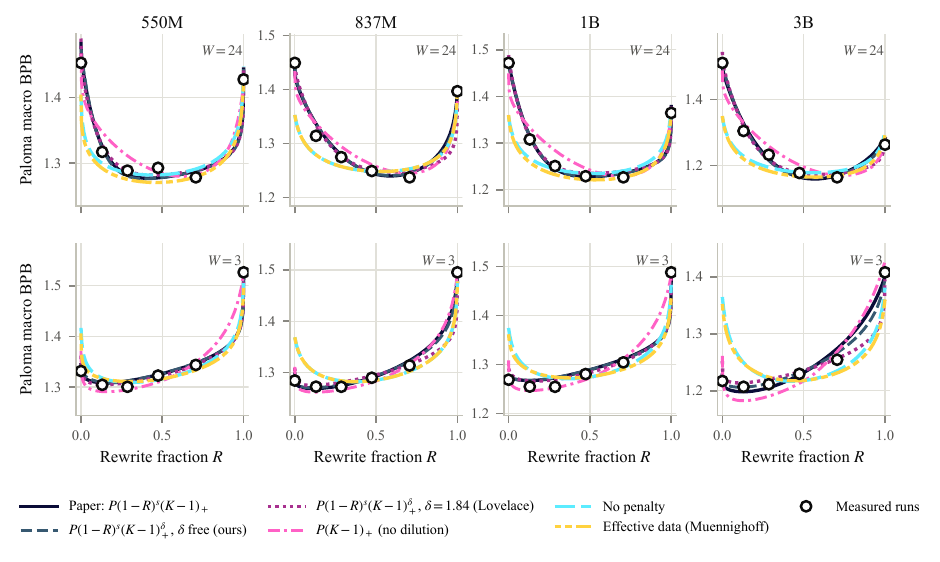}
\end{center}
\caption{Fitted Paloma BPB against rewrite fraction under each repetition penalty (full $48$-run fits), at $W=24$ (top) and $W=3$ (bottom), with the measured runs. The four penalties with an additive term are nearly coincident at $W=24$ between the tested fractions; the two forms without one fall $0.05$--$0.16$ BPB below the $R=0$ run at $W=24$ and $0.07$--$0.15$ BPB above it at $W=3$. The no-dilution form places its $W=3$ minimum below the measured points at small $R$.}
\label{fig:de-penalty-ucurves}
\end{figure}

\begin{figure}[t]
\begin{center}
\includegraphics[width=\textwidth]{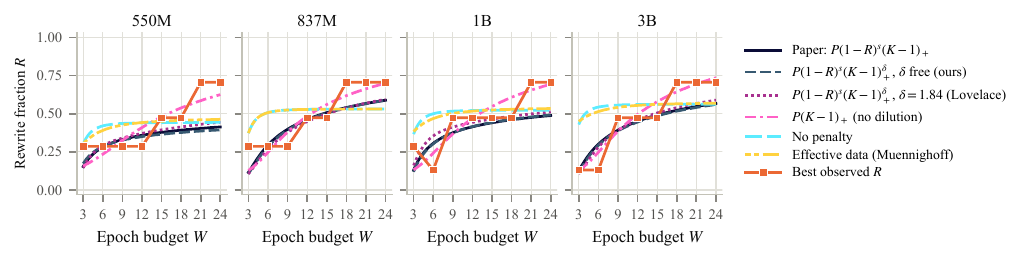}
\end{center}
\caption{Estimated optimal rewrite fraction $\Rhat(W)$ under each repetition penalty (full $48$-run fits), with the best observed $R$ at each budget. All forms with the dilution factor $(1-R)^{s}$ give $\Rhat(24)$ between $0.39$ and $0.59$, below the best observed $R=0.71$ at $W\ge18$; the no-dilution form rises faster ($0.62$--$0.74$) but at the cost of fit quality and a degenerate floor (Table~\ref{tab:de-penalty-ablation}). The forms without an additive penalty are flat in $W$.}
\label{fig:de-penalty-rstar}
\end{figure}

\begin{figure}[t]
\begin{center}
\includegraphics[width=\textwidth]{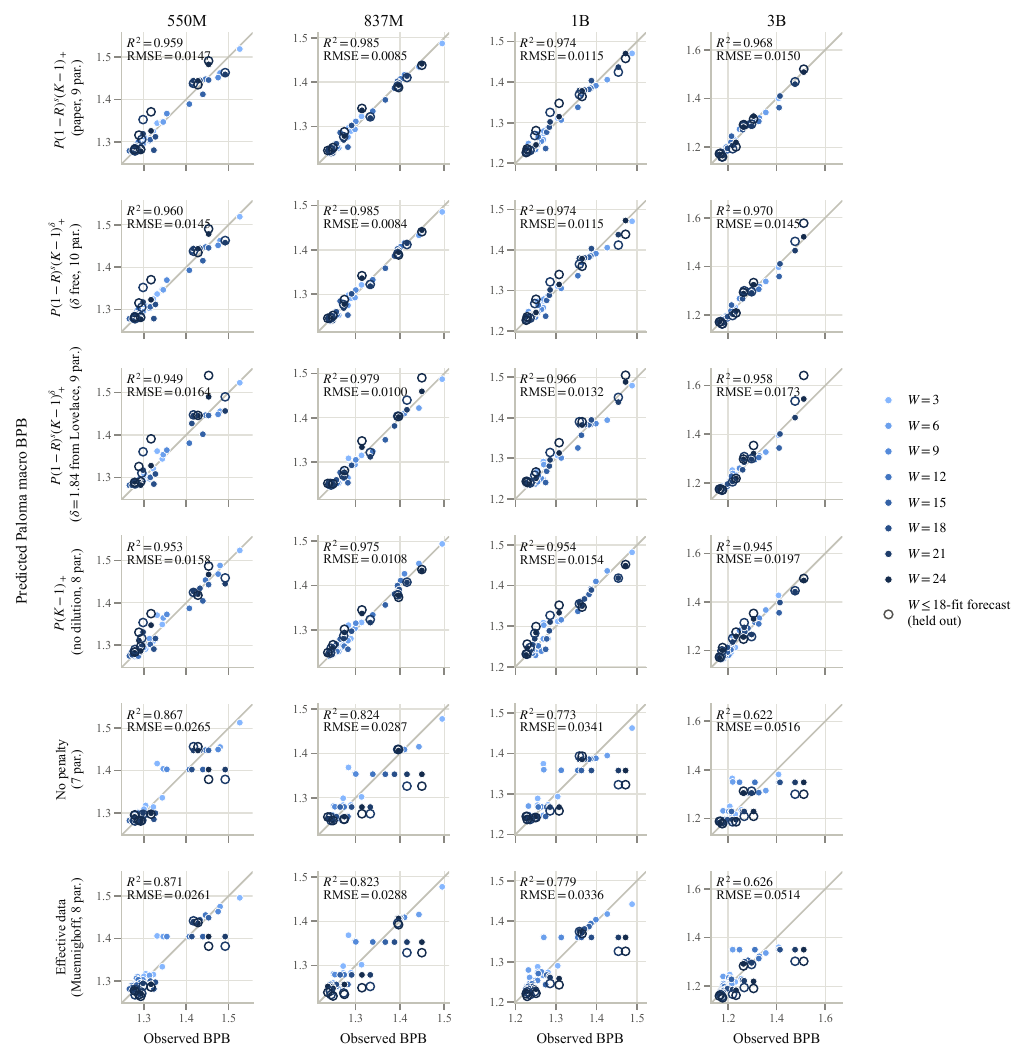}
\end{center}
\caption{Observed against predicted Paloma macro BPB for every repetition penalty (rows, in the order of Table~\ref{tab:de-penalty-ablation}) and target--rewriter pair (columns), all $48$ runs of the full fit, coloured by epoch budget $W$, with the identity line. Hollow markers are the $W=21$ and $W=24$ forecasts of the $W\le18$ refit. Each panel reports $R^2$ and RMSE of the full fit; axis limits are shared within a column. The three dilution forms and the no-dilution form lie along the identity line; the two forms without an additive penalty flatten to horizontal bands at large $W$, where the $R=0$ runs cannot be separated from their neighbours by a term in $D$ alone.}
\label{fig:de-penalty-fitgrid}
\end{figure}

Eq.~(\ref{eq:de_law}) charges repeated natural tokens with the additive penalty $F=P\,(1-R)^{s}((1-R)D/T-1)_+$, with $(1-R)D/T=3W(1-R)/(3+2R)$ natural epochs. The factor $((1-R)D/T-1)_+$ is the one-parameter penalty of \citet{lovelace2026prescriptivescalinglawsdata} (their Eq.~6), linear in the epochs beyond the first, with the capacity ratio $N/T$ absorbed into $P$ because both are fixed within a data-limited fit; the dilution factor $(1-R)^{s}$ is ours. We refit the $48$ runs of every target--rewriter pair, with the floor $E(R)$, $B$, and $\beta$ free and the protocol of Table~\ref{tab:de-fit}, under six penalties: (i) the paper's $P(1-R)^{s}((1-R)D/T-1)_+$; (ii) $P(1-R)^{s}((1-R)D/T-1)_+^{\delta}$ with $\delta\in[0.5,3]$ free, the superlinear exponent on the repetition count of their Eq.~8; (iii) the same with $\delta$ fixed at $1.84$, the value that \citet{lovelace2026prescriptivescalinglawsdata} fit for their Eq.~8 on their own runs; we do not fit it, and this variant tests whether their exponent transfers to our setting (labeled ``$\delta=1.84$ (Lovelace)'' in Figures~\ref{fig:de-penalty-ucurves}--\ref{fig:de-penalty-fitgrid}); (iv) $P((1-R)D/T-1)_+$, their Eq.~6 without dilution ($s=0$); (v) no penalty, $P=0$, so that $D$ enters the Chinchilla term only; and (vi) no additive penalty but the effective-data term of \citet{muennighoff2023scalingdataconstrainedlanguage} in place of $D$, with the saturating count applied to the natural epochs only, since rewrites are never repeated. Their Eq.~7 exponent on $N/U_D$ is not identifiable here, because $N$ and $T$ do not vary within a pair. Per pair, $8$ runs have $(1-R)D/T=0$ ($R=1$), one has $0<(1-R)D/T<1$ ($W=3$, $R=0.71$), and $39$ have $(1-R)D/T>1$ (up to $(1-R)D/T=24$ at $R=0$), so every penalty is identified on the same $39$ cells and the data cannot separate $((1-R)D/T-1)_+$ from $(1-R)D/T$ (Section~\ref{scaling_fit}).

\begin{table}[t]
\caption{Repetition-penalty form ablation for Eq.~(\ref{eq:de_law}), one fit per target--rewriter pair, $48$ runs per fit. RMSE is in BPB and $R^2=1-\mathrm{SS}_{\mathrm{res}}/\mathrm{SS}_{\mathrm{tot}}$ over the $48$ runs. The held-out columns refit the $36$ runs at $W\le18$ and forecast the six rewrite fractions at $W=21$ and $W=24$: mean error is the mean of $(L_{\mathrm{measured}}-L_{\mathrm{predicted}})/L_{\mathrm{predicted}}$ over the six, error at best obs.\ $R$ is the same quantity at the best observed $R$ ($R=0.71$ at every pair and both budgets), and $R_{\mathrm{sel}}$ is the tested fraction with the lowest forecast loss, the selection rule of Figure~\ref{fig:de-ucurves}. $\Rhat_{24}$ is the minimizer of the full $48$-run fit over $R\in[0,1]$ at $W=24$. The parameter count includes the five floor parameters, $B$, and $\beta$. Fitted parameters are in \texttt{repetition\_penalty/de\_penalty\_ablation.csv}.}
\label{tab:de-penalty-ablation}
\begin{center}
\scriptsize
\setlength{\tabcolsep}{2.2pt}
\resizebox{\textwidth}{!}{%
\begin{tabular}{r|c|c|cc|cc|cc|cc|c}
\toprule
\multirow{2}{*}{Penalty} & \multirow{2}{*}{Params} & \multirow{2}{*}{Target} & \multirow{2}{*}{RMSE} & \multirow{2}{*}{$R^2$} & \multicolumn{2}{c}{Mean error} & \multicolumn{2}{c}{Error at best obs.\ $R$} & \multicolumn{2}{c}{$R_{\mathrm{sel}}$} & \multirow{2}{*}{$\hat{R}^*_{24}$} \\
\cmidrule(lr){6-7}\cmidrule(lr){8-9}\cmidrule(lr){10-11}
 & & & & & $W=21$ & $W=24$ & $W=21$ & $W=24$ & $W=21$ & $W=24$ & \\
\midrule
$P(1-R)^{s}((1-R)D/T-1)_+$ & $9$ & $550$M & $0.0147$ & $0.959$ & $-0.73\%$ & $-1.39\%$ & $-0.28\%$ & $-0.26\%$ & $0.47$ & $0.71$ & $0.41$ \\
(paper) &  & $837$M & $0.0085$ & $0.985$ & $+0.13\%$ & $-0.42\%$ & $-0.27\%$ & $-0.68\%$ & $0.71$ & $0.71$ & $0.59$ \\
 &  & $1$B & $0.0115$ & $0.974$ & $-0.50\%$ & $-0.81\%$ & $+0.05\%$ & $+0.03\%$ & $0.71$ & $0.71$ & $0.49$ \\
 &  & $3$B & $0.0150$ & $0.968$ & $+0.27\%$ & $-0.04\%$ & $-0.50\%$ & $-0.74\%$ & $0.47$ & $0.47$ & $0.57$ \\
\midrule
$P(1-R)^{s}((1-R)D/T-1)_+^{\delta}$ & $10$ & $550$M & $0.0145$ & $0.960$ & $-0.73\%$ & $-1.39\%$ & $-0.28\%$ & $-0.26\%$ & $0.47$ & $0.71$ & $0.39$ \\
$\delta$ free &  & $837$M & $0.0084$ & $0.985$ & $+0.09\%$ & $-0.48\%$ & $-0.27\%$ & $-0.68\%$ & $0.71$ & $0.71$ & $0.59$ \\
 &  & $1$B & $0.0115$ & $0.974$ & $-0.24\%$ & $-0.40\%$ & $+0.01\%$ & $-0.01\%$ & $0.71$ & $0.71$ & $0.49$ \\
 &  & $3$B & $0.0145$ & $0.970$ & $-0.30\%$ & $-0.98\%$ & $-0.34\%$ & $-0.56\%$ & $0.47$ & $0.47$ & $0.56$ \\
\midrule
$P(1-R)^{s}((1-R)D/T-1)_+^{1.84}$ & $9$ & $550$M & $0.0164$ & $0.949$ & $-1.45\%$ & $-2.60\%$ & $-0.67\%$ & $-0.75\%$ & $0.47$ & $0.71$ & $0.44$ \\
 &  & $837$M & $0.0100$ & $0.979$ & $-0.37\%$ & $-1.24\%$ & $-0.72\%$ & $-1.16\%$ & $0.47$ & $0.47$ & $0.59$ \\
 &  & $1$B & $0.0132$ & $0.966$ & $-1.09\%$ & $-1.64\%$ & $-1.14\%$ & $-1.32\%$ & $0.47$ & $0.47$ & $0.51$ \\
 &  & $3$B & $0.0173$ & $0.958$ & $-1.01\%$ & $-2.19\%$ & $-0.52\%$ & $-0.80\%$ & $0.47$ & $0.47$ & $0.59$ \\
\midrule
$P((1-R)D/T-1)_+$ & $8$ & $550$M & $0.0158$ & $0.953$ & $-0.84\%$ & $-1.61\%$ & $-0.34\%$ & $-0.44\%$ & $0.71$ & $0.71$ & $0.62$ \\
(no dilution) &  & $837$M & $0.0108$ & $0.975$ & $-0.01\%$ & $-0.67\%$ & $-0.37\%$ & $-0.92\%$ & $0.71$ & $0.71$ & $0.70$ \\
 &  & $1$B & $0.0154$ & $0.954$ & $-0.69\%$ & $-1.13\%$ & $-0.16\%$ & $-0.37\%$ & $0.71$ & $0.71$ & $0.69$ \\
 &  & $3$B & $0.0197$ & $0.945$ & $-0.54\%$ & $-1.38\%$ & $-0.21\%$ & $-0.75\%$ & $0.71$ & $0.71$ & $0.74$ \\
\midrule
No penalty & $7$ & $550$M & $0.0265$ & $0.867$ & $+0.86\%$ & $+0.83\%$ & $-1.24\%$ & $-1.31\%$ & $0.47$ & $0.47$ & $0.44$ \\
 &  & $837$M & $0.0287$ & $0.824$ & $+1.93\%$ & $+2.13\%$ & $-1.09\%$ & $-1.54\%$ & $0.47$ & $0.47$ & $0.53$ \\
 &  & $1$B & $0.0341$ & $0.773$ & $+1.45\%$ & $+1.96\%$ & $-1.26\%$ & $-1.45\%$ & $0.47$ & $0.47$ & $0.52$ \\
 &  & $3$B & $0.0516$ & $0.622$ & $+3.01\%$ & $+3.75\%$ & $-1.60\%$ & $-1.99\%$ & $0.47$ & $0.47$ & $0.56$ \\
\midrule
Effective data & $8$ & $550$M & $0.0261$ & $0.871$ & $+1.66\%$ & $+1.85\%$ & $-0.07\%$ & $+0.14\%$ & $0.47$ & $0.47$ & $0.46$ \\
(Muennighoff) &  & $837$M & $0.0288$ & $0.823$ & $+2.76\%$ & $+3.17\%$ & $+0.04\%$ & $-0.17\%$ & $0.47$ & $0.47$ & $0.53$ \\
 &  & $1$B & $0.0336$ & $0.779$ & $+2.50\%$ & $+3.31\%$ & $+0.28\%$ & $+0.45\%$ & $0.47$ & $0.47$ & $0.53$ \\
 &  & $3$B & $0.0514$ & $0.626$ & $+4.36\%$ & $+5.44\%$ & $+0.32\%$ & $+0.36\%$ & $0.47$ & $0.47$ & $0.57$ \\
\bottomrule
\end{tabular}%
}
\end{center}
\end{table}

Three conclusions follow (Table~\ref{tab:de-penalty-ablation}, Figures~\ref{fig:de-penalty-ucurves}--\ref{fig:de-penalty-fitgrid}). Figure~\ref{fig:de-penalty-fitgrid} shows the fits run by run: the paper form and the two $\delta$ variants have indistinguishable $R^2$ ($0.959$--$0.985$ across pairs for the paper form, within $0.011$ of it for $\delta$ free or fixed), the no-dilution form loses $0.006$--$0.023$, and the two forms without an additive penalty fall to $R^2=0.62$--$0.87$, lowest at $3$B, where the penalty is largest. First, an additive penalty is needed: dropping it raises RMSE by $1.8$--$3.4\times$, pushes $\beta$ to its upper bound of $3$, and under-predicts the held-out $W=24$ runs by $0.8$--$3.8\%$ on average because the $R=0$ run at $W=24$ cannot be reached by a term in $D$ alone; the effective-data variant fits no better, since its saturation parameter goes to $0.85$--$0.92$ with $\beta$ at its lower bound at three pairs and to its upper bound of $200$, which is the no-penalty form, at the fourth. Second, the exponent on the repetition count is not identified by this grid. Freeing $\delta$ lowers RMSE by at most $0.0005$ BPB and returns $\delta=0.77,1.10,1.05,1.19$ across pairs, indistinguishable from the linear form on the tested fractions, while $s$ moves to compensate ($s=8$, its bound, at $550$M). Fixing $\delta=1.84$, the value \citet{lovelace2026prescriptivescalinglawsdata} obtain from residuals against a single-epoch Chinchilla fit with $N/U_D$ varied, is worse at every pair ($+0.0015$ to $+0.0023$ BPB), drives $\beta$ to its bound at $1$B, and doubles the held-out mean error; their exponent does not transfer once the floor and $s$ are free. Third, the dilution factor $(1-R)^{s}$ is what the linear form needs to coexist with the mixture floor: without it ($s=0$) RMSE rises by $0.0011$--$0.0047$ BPB and the fit is degenerate, with $E_0\to0$ and $\beta$ at its lower bound at two pairs, so the Chinchilla term becomes a near-constant and the floor absorbs the penalty's $R$-dependence. On the selection metric of Figure~\ref{fig:de-ucurves} the picture is less flattering to the paper's form: with the $W\le18$ fit it selects $R=0.47$ instead of the best observed $R=0.71$ at $3$B (measured cost $0.50\%$ at $W=21$ and $1.14\%$ at $W=24$) and at $550$M, $W=21$ ($0.01\%$), as do the $\delta$ variants, whereas the no-dilution form selects $0.71$ at all eight held-out cells despite its worse RMSE, because its $\Rhat(W)$ rises faster. All forms with dilution place the continuous $\Rhat(24)$ at $0.39$--$0.59$, below the best observed $R$, and the two forms without a penalty are flat in $W$. We keep the linear form of Eq.~(\ref{eq:de_law}): it is the most parsimonious form that reproduces the U-curves, and the extra exponent buys nothing the data can resolve. The exponent $s$, not $\delta$, governs how fast $\Rhat$ grows with budget, and it is weakly identified ($s=1.5$--$7.0$ across pairs), so extrapolations of $\Rhat(W)$ beyond $W=24$ should be read with that in mind.

\subsubsection{Ablating the Curvature Term}
\label{scaling_curv_ablation}

The compute-limited surface adds to Eq.~(\ref{eq:rewriting_scaling_law}) with $F=0$ a term fit to the excess curvature of the measured IsoFLOP parabolas,
\begin{equation}
\mathrm{Curv}(N, D) = \frac{H}{C^{\eta}}\left[\log_{10}\frac{D}{D^*(C)}\right]^2, \qquad C = 6ND,
\label{eq:curv}
\end{equation}
so that
\begin{align}
L_{\mathrm{CE}}(N,D,R) = E(R) + \frac{A}{N^{\alpha}} + \frac{B}{D^{\beta}} + \mathrm{Curv}(N, D).
\label{eq:cc_surface}
\end{align}
The term vanishes with zero slope on the frontier, so it changes off-frontier predictions without moving $N^*(C)$, $D^*(C)$, or $L^*_{\mathrm{base}}(C)$.

We ablate the term on the $671$-run fit of Appendix~\ref{infinite_compute}, holding the floor and Chinchilla terms fixed. Removing it raises the residual sum of squares from $0.0608$ to $0.2726$, with likelihood-ratio statistic $1006.6$. The architecture-wise wild bootstrap of Appendix~\ref{indep_R_N_D}, over $79$ architectures and $400$ draws, rejects the restriction ($p=1/401$).

\subsection{Domain-Specific Compute Optima}
\label{knowledge_reasoning}

\paragraph{Break-even with SmolLM2-Instruct by evaluation.}
On QA, the lowest fitted QA BPB over model size, token count, and rewrite fraction matches SmolLM2-1.7B-Instruct at $5.3\times10^{21}$ total FLOPs with $\Rhat=0.74$, and the $10^{22}$-FLOP run trained at the QA optimum measures $0.696$ QA BPB against $0.709$ for SmolLM2-Instruct. On Math and Code the fitted minimum matches SmolLM2-Instruct only at ${\approx}6\times10^{22}$ and ${\approx}4\times10^{23}$ total FLOPs, $6\times$ and $43\times$ the largest budget we trained, and the Code crossing lies above SmolLM2's ${\approx}1.1\times10^{23}$ training FLOPs. SmolLM2's training corpus includes FineMath and Stack-Edu \citep{benallal2025smollm2smolgoesbig} and StarCoderData \citep{li2023starcoder}, math and code data that FineWeb-Edu largely lacks. For each domain, the $10^{22}$-FLOP run trained at that domain's fitted optimum measures a higher loss than its surface forecasts, by $4\%$ on QA, $14\%$ on Math, and $32\%$ on Code, against $1\%$ for the Paloma-optimal run. We hypothesize two causes. First, we tuned the training recipe on Paloma, whose macro average weights $16$ sources equally: ten web or encyclopedic text sources, five social-media sources, one code source, and no math source (Appendix~\ref{evals}). Web and encyclopedic text carries the knowledge that QA evaluations test, so the forecast error is smallest on Paloma and then QA. Second, the rewrite styles include math and not code. The rewrites contain at most $0.8\%$ code lines and add lists, tables, and headers rather than code (Appendix~\ref{rewrite_faithful}), so code remains rare in the training data even at $R=1$, and the Code surface is fit to the narrowest range of loss.

\begin{figure}[t]
\begin{center}
\includegraphics[width=\textwidth]{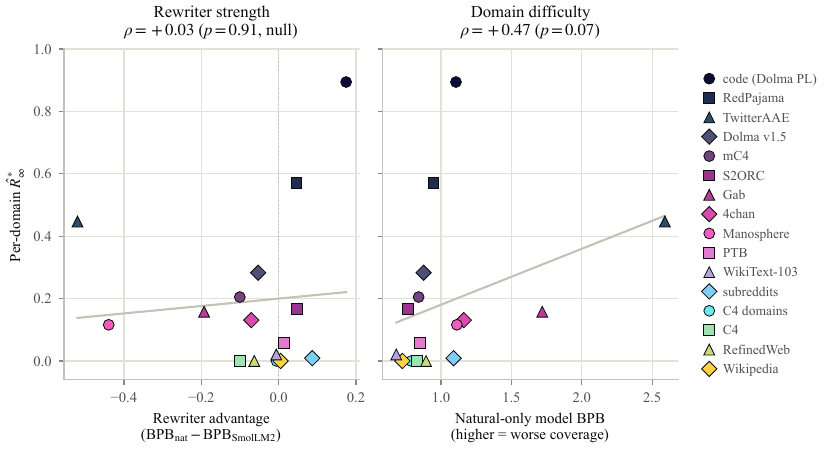}
\end{center}
\caption{The estimated floor-optimal rewrite fraction $\Rhat_\infty$ against two properties of each Paloma source. \textbf{Left:} $\Rhat_\infty$ has no association with the rewriter's relative strength on the source (Spearman $\rho=0.03$, $p=0.91$). \textbf{Right:} $\Rhat_\infty$ increases with the natural-data model's BPB on the source (Spearman $\rho=0.47$, $p=0.07$).}
\label{fig:rstar_domains}
\end{figure}

\begin{figure}[t]
\begin{center}
\includegraphics[width=\textwidth]{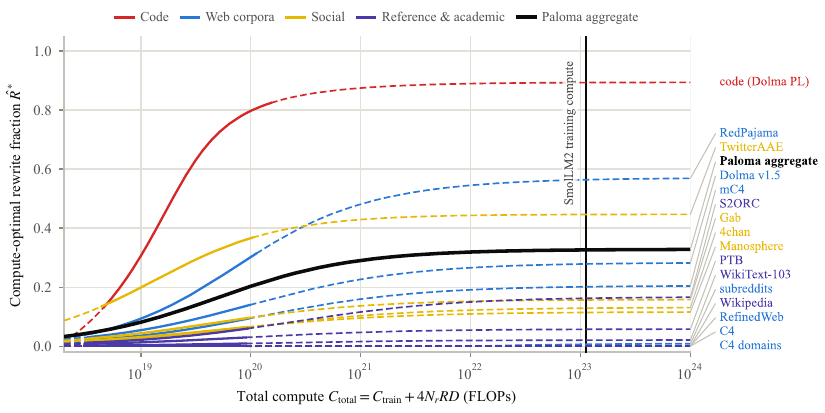}
\end{center}
\caption{Estimated compute-optimal rewrite fraction $\Rhat(C_{\mathrm{total}})$ for each Paloma source. Solid curves cover the measured ladders through ${\approx}2\times10^{20}$ total FLOPs, and dashed curves extrapolate. The thick curve is Paloma macro BPB. The vertical line marks SmolLM2's training compute.}
\label{fig:rstar_domains_traj}
\end{figure}

\begin{figure}[t]
\begin{center}
\includegraphics[width=\textwidth]{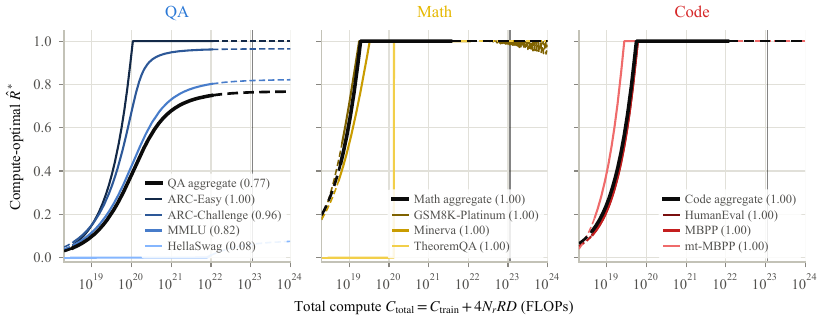}
\end{center}
\caption{Estimated compute-optimal rewrite fraction $\Rhat(C_{\mathrm{total}})$ for each task. Legends report $\Rhat_\infty$, and thick curves show the aggregate trajectories of Figure~\ref{fig:rstar_skills}. MMLU, Minerva, and multilingual MBPP average $57$, $7$, and $17$ subtasks. TheoremQA switches from $R=0$ to $R=1$ because its fitted floor is nearly linear in $R$; its fit excludes the two budgets with the fewest model widths.}
\label{fig:rstar_evals_traj}
\end{figure}

Figures~\ref{fig:rstar_domains_traj} and~\ref{fig:rstar_evals_traj} decompose the aggregate trajectories of Figure~\ref{fig:rstar_skills} into Paloma sources and domain-suite tasks. Across Paloma sources, $\Rhat_\infty$ reaches ${\approx}0.89$ for programming languages and $0.57$ for RedPajama, while C4, RefinedWeb, and Wikipedia remain at $\Rhat=0$ throughout the evaluated compute range (Figure~\ref{fig:rstar_domains_traj}). Every Math and Code task reaches $\Rhat_\infty=1$. Within QA, ARC-Easy, ARC-Challenge, and MMLU reach $\Rhat_\infty\geq0.82$, HellaSwag remains below $0.1$, and the QA aggregate reaches $0.77$ (Figure~\ref{fig:rstar_evals_traj}).

Each trajectory comes from a separate surface fit to its aggregate. The Paloma, QA, Math, and Code fits contain $604$, $597$, $608$, and $610$ observations and have RMSE $0.0148$, $0.0195$, $0.0301$, and $0.0681$ BPB. Their estimated floor optima are $0.329$, $0.768$, $1.0$, and $1.0$. The Math and Code estimates lie at the boundary, and their break-even points lie beyond the measured ladders.

\begin{figure}[t]
\centering
\includegraphics[width=\textwidth]{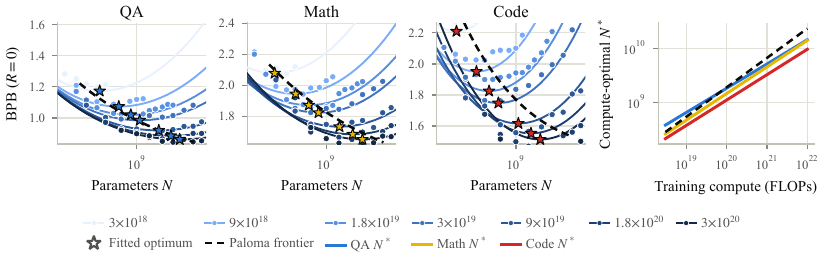}
\caption{\textbf{Knowledge QA is capacity-hungry and Code is data-hungry.} \textbf{Left three panels:} natural-only ($R=0$) IsoFLOP slices from each evaluation's own surface fit, loss against parameter count, colored by training budget, with dots for runs and stars for the fitted optimum at each budget. The dashed line is the Paloma frontier $N^*(C)$ drawn on that evaluation's surface. \textbf{Right:} compute-optimal parameter count of each evaluation's fit against training compute. The QA optimum has more parameters than the Code optimum at every budget, $1.5\times$ at $10^{22}$ FLOPs.}
\label{fig:alloc_by_eval}
\end{figure}

\paragraph{Compute-optimal allocation by evaluation.}
Each per-evaluation surface fit has its own compute-optimal frontier $N^*(C)$ and $D^*(C)$, shared across rewrite fractions within the fit.
Figure~\ref{fig:alloc_by_eval} shows each evaluation's natural-only IsoFLOP slices with its fitted optima and the Paloma frontier overlaid, and Table~\ref{tab:alloc_by_eval} reports the frontier at two training budgets.
The QA frontier allocates the fewest tokens per parameter and the Code frontier the most, with Math between them: at $10^{22}$ FLOPs the QA optimum has $1.5\times$ the parameters of the Code optimum at the same training compute.
This reproduces the finding of \citet{roberts2025computeoptimalscalingskills} that knowledge QA is capacity-hungry while code is data-hungry, and that the compute-optimal parameter count depends on the skill composition of the validation set by up to ${\approx}50\%$.
Our fits reach the same conclusion on ladders that include rewrites at every fraction, so the allocation difference is a property of the evaluation rather than of the rewrite fraction.
Relative to the Paloma frontier, which increases more steeply in $N$, the QA optimum lies at more parameters below ${\approx}10^{20}$ FLOPs and fewer above, while the Code optimum lies at fewer parameters at every budget.

\begin{table}[t]
\caption{Compute-optimal allocation from each evaluation's surface fit at two training budgets. $D^*/N^*$ is tokens per parameter.}
\label{tab:alloc_by_eval}
\begin{center}
\small
\begin{tabular}{r|ccc|ccc}
\toprule
\multirow{2}{*}{Evaluation} & \multicolumn{3}{c}{$C=10^{19}$} & \multicolumn{3}{c}{$C=10^{22}$} \\
\cmidrule(lr){2-4}\cmidrule(lr){5-7}
 & $N^*$ & $D^*$ & $D^*/N^*$ & $N^*$ & $D^*$ & $D^*/N^*$ \\
\midrule
Paloma & $0.55$B & $3.0$B & $5.5$ & $23.5$B & $70.9$B & $3.0$ \\
QA & $0.66$B & $2.5$B & $3.8$ & $14.6$B & $114.4$B & $7.8$ \\
Math & $0.48$B & $3.5$B & $7.3$ & $14.2$B & $117.2$B & $8.2$ \\
Code & $0.38$B & $4.4$B & $11.5$ & $9.7$B & $171.7$B & $17.7$ \\
\bottomrule
\end{tabular}
\end{center}
\end{table}

\subsection{Evaluations of the $10^{22}$-FLOP Runs}
\label{hero_evals}
Table~\ref{tab:hero_aggregate} compares the $10^{22}$-FLOP runs with SmolLM2 using the matched Paloma reference and the offline QA, Math, and Code references. Table~\ref{tab:hero_individual} reports the $10^{22}$ runs by Paloma source. Tables~\ref{tab:hero_tasks} and~\ref{tab:hero_subtasks} break the QA, Math, and Code aggregates into their tasks, subjects, and languages.

\input{sections/tables/hero_aggregate}

\input{sections/tables/hero_paloma_domains}

\input{sections/tables/hero_skill_tasks}

\input{sections/tables/hero_subtasks}

\subsection{Breakeven Parameters}
\label{breakeven_parameters}
\subsubsection{Definition and Relation to Training Compute}

For each comparison metric $m$, break-even is the smallest total compute, accounting for rewrites, at which the fitted compute-optimal frontier reaches SmolLM2-Instruct's score:
\begin{align}
C_{\mathrm{BE},m}=\min\left\{C_{\mathrm{total}}:
L_m^*(C_{\mathrm{total}})\leq L_{m,\mathrm{SmolLM2\text{-}inst}}\right\}.
\end{align}
Here $L_m^*(C_{\mathrm{total}})$ is the loss at the joint optimum of Eq.~(\ref{eq:rstar}) on metric $m$'s fitted surface.
SmolLM2-Instruct enters only as the reference score. Its own training compute is $1.1286\times10^{23}$ FLOPs. The crossing accounts for rewrite generation but not SmolLM2's training. Paloma uses the $0.9716$ BPB reference matched to our byte-pooled aggregation and $4096$-token context. QA, Math, and Code use the same offline evaluations for both model families.

\subsubsection{Breakeven \texorpdfstring{$N$, $D$, and $R$}{N, D, and R} per Domain}

Table~\ref{tab:breakeven-allocation} gives the fitted allocation at each matched crossing. The training-compute column is $6N^*D^*$, and its difference from total compute is the generation cost. Paloma, QA, and Math cross below SmolLM2's training compute, and Code crosses above it.

\begin{table*}[t]
\caption{Fitted break-even allocations against SmolLM2-Instruct. Paloma crosses at $2.53\times10^{22}$ FLOPs with $\Rhat=0.325$, using matched within-source aggregation and context length. QA crosses at $5.32\times10^{21}$ FLOPs with $\Rhat=0.739$. The Math and Code crossings lie at the $R=1$ boundary and beyond the measured ladders.}
\label{tab:breakeven-allocation}
\begin{center}
\small
\begin{tabular}{r|ccccccc}
\toprule
\multirow{2}{*}{Metric} & \multirow{2}{*}{Reference} & $C_{\mathrm{total}}$ & $C$ & $N^*$ & $D^*$ & \multirow{2}{*}{$\hat{R}^*$} & \multirow{2}{*}{Gen. share} \\
 & & (FLOPs) & (FLOPs) & (B) & (B) & & \\
\midrule
Paloma & $0.9716$ & $2.528\times10^{22}$ & $2.505\times10^{22}$ & $38.960$ & $107.145$ & $0.325$ & $0.94\%$ \\
QA BPB & $0.7090$ & $5.316\times10^{21}$ & $4.946\times10^{21}$ & $11.254$ & $73.245$ & $0.739$ & $6.96\%$ \\
Math BPB & $0.7074$ & $5.962\times10^{22}$ & $5.769\times10^{22}$ & $34.113$ & $281.855$ & $1.000$ & $3.23\%$ \\
Code BPB & $0.4256$ & $4.274\times10^{23}$ & $4.189\times10^{23}$ & $55.993$ & $1246.794$ & $1.000$ & $2.00\%$ \\
\bottomrule
\end{tabular}
\end{center}
\end{table*}

\subsection{Loss- vs Accuracy-Based Fits}
\label{loss_accuracy}
\subsubsection{Parametric Accuracy Model}

As a check on the BPB criterion, we fit QA accuracy directly on all $607$ accuracy observations by mapping the surface through a logistic link:
\begin{align}
\widehat{a}(N,D,R)&=\frac{\theta}{1+\exp z(N,D,R)},\\
z(N,D,R)&=E(R)+\frac{A}{N^\alpha}+\frac{B}{D^\beta}
+H C^{-\eta}\left[\log_{10}\frac{D}{D^*(C)}\right]^2,
\end{align}
where $C=6ND$ and $E(R)$ is the mixture-dependent floor of Eq.~(\ref{eq:floor}). The fit minimizes a soft-$L_1$ objective with scale $0.02$, initialized from the QA-BPB surface, and refits every parameter.

\subsubsection{Accuracy Fits}

The observed QA accuracies span $0.3041$--$0.5129$. The fit has $R^2=0.98795$, RMSE $0.00453$, and mean absolute error $0.00345$. Its ceiling $\theta$ reaches the upper bound $1$ and its intercept $E_0$ the lower bound $-10$, so the asymptotic parameters are weakly identified although the in-grid fit is accurate.

With the $1.71$B rewriter cost, the accuracy fit gives $\Rhat(10^{22})=0.638$. It reaches the SmolLM2-instruct and base QA accuracies, $0.606$ and $0.617$, at $3.336\times10^{21}$ and $4.484\times10^{21}$ FLOPs. Both references exceed every accuracy in the fitting grid, so both crossings are extrapolations. The QA hero run obtains $0.62017$ accuracy against a forecast of $0.64734$. The QA-BPB surface gives $\Rhat(10^{22})=0.749$ and crosses the instruct BPB reference at $5.316\times10^{21}$ FLOPs. Both criteria favor a large rewrite fraction, and the exact estimated optimum and crossing depend on the criterion.

\subsection{Scaling Setup Ablations}
\label{scaling_setup}
\subsubsection{Alternative Surface Specifications}

We compare three nested surface models on the $583$ runs of Appendix~\ref{indep_R_N_D}. M1 shares the compute-optimal center across rewrite fractions and gives each fraction a floor shift. M2 adds an $R$-specific center and curvature. M3 fits an independent log-quadratic to every $(C,R)$ cell. Table~\ref{tab:surface-model-ladder} reports the residual error and Bayesian information criterion (BIC) of each.

\begin{table}[t]
\caption{Surface-model ladder on the same $583$ runs. BIC favors the shared-center model M1 although the richer models reduce residual error.}
\label{tab:surface-model-ladder}
\begin{center}
\small
\begin{tabular}{r|cccc}
\toprule
Model & Parameters & $R^2$ & RMSE & BIC \\
\midrule
M1: shared center & $25$ & $0.98304$ & $0.00858$ & $-5383$ \\
M2: center by $R$ & $45$ & $0.98579$ & $0.00785$ & $-5359$ \\
M3: cell quadratics & $165$ & $0.99022$ & $0.00651$ & $-4813$ \\
\bottomrule
\end{tabular}
\end{center}
\end{table}

Nested $F$ tests reject M1 against M2 ($p=4.73\times10^{-12}$) and against M3 ($p=7.79\times10^{-10}$). The $R$-dependent floor shifts nevertheless agree within ${\approx}0.005$ BPB across the three models, and M1 attains $99.3\%$ of M3's $R^2$ with $15\%$ of its parameters. We use M1's shared center.

\subsection{Optimal Rewrite Fraction across Source Quality}
\label{dclm_scaling}

\begin{figure}[t]
\begin{center}
\includegraphics[width=0.49\textwidth]{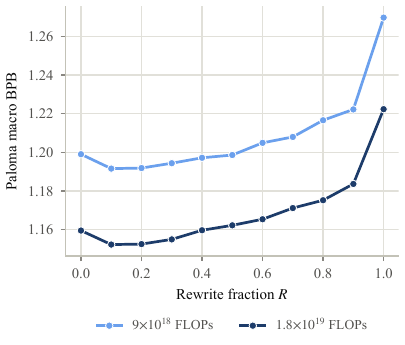}\hfill
\includegraphics[width=0.49\textwidth]{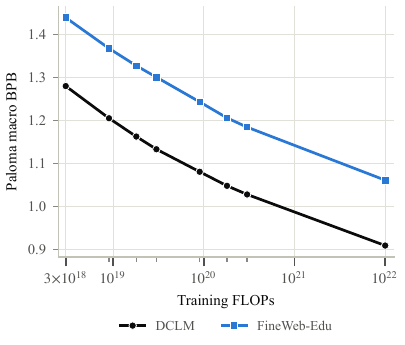}
\end{center}
\caption{\textbf{Left:} Paloma macro BPB against $R$ on DCLM at two training budgets: $550$M parameters and $2.9$B tokens at $9\times10^{18}$ FLOPs, and $669$M parameters and $4.7$B tokens at $1.8\times10^{19}$ FLOPs. \textbf{Right:} final Paloma macro BPB of the observed compute-optimal natural-only ($R=0$) DCLM and FineWeb-Edu run at each training budget shown. Each point selects the lowest-Paloma run within its corpus and budget.}
\label{fig:dclm}
\end{figure}

\paragraph{The estimated floor-optimal rewrite fraction is smaller, but non-zero, on a stronger source corpus.}
We replace FineWeb-Edu with DCLM \citep{li2024datacomplmsearchgenerationtraining} and sweep $R$ at fixed $N$ and $D$ at training budgets of $C = 9\times10^{18}$ and $1.8\times10^{19}$ FLOPs. These sweeps do not account for generation, so they estimate the floor $E(R)$ and its optimum $R^*_\infty$ rather than the compute-optimal $R^*$ of Section~\ref{sec:scaling_model}. DCLM is the stronger corpus: at every budget with final evaluations for both corpora, the natural-only DCLM run with the lowest Paloma macro BPB lies below its FineWeb-Edu counterpart (Figure~\ref{fig:dclm}, right), and DCLM has lower Paloma BPB in all $77$ matched width--depth cells over seven budgets (Appendix~\ref{dclm_scaling}). On DCLM, $R=0.1$ minimizes Paloma BPB at both budgets, and its gain over $R=0$ is smaller than the gain of the best rewrite fraction over $R=0$ on the FineWeb-Edu sweeps at the same budgets (Figure~\ref{fig:dclm}, left). The rest of this appendix fits a surface to DCLM and reports its estimated floor-optimal rewrite fraction and the matched comparison with FineWeb-Edu.

The DCLM sweep gives a fitted $\Rhat_\infty\approx0.08$ (Figure~\ref{fig:dclm}). This appendix gives the natural-only ladder, the floor fit, and the matched comparison with FineWeb-Edu.

\paragraph{Natural-only ladder.}
The DCLM ladder contains $81$ completed $R=0$ runs at eight budgets, $3\times10^{18}$, $9\times10^{18}$, $1.8\times10^{19}$, $3\times10^{19}$, $9\times10^{19}$, $1.8\times10^{20}$, $3\times10^{20}$, and $10^{22}$ FLOPs, with $5$, $10$, $15$, $10$, $14$, $13$, $13$, and $1$ runs (Figure~\ref{fig:dclm-ladder-grid}). The DCLM runs follow the recipe of Appendix~\ref{completedp} with a reference point tuned on DCLM: a Gaussian-process search over the Hyperball learning rate, Adam learning rate, and $\epsilon$ at $d=512$ and $2.5\times10^{9}$ tokens, with $\beta_1=0.9$ and $\beta_{2,0}=0.9999$ fixed, gives $\eta_0=1.45\times10^{-2}$, $\eta^{\mathrm A}_0=3.28\times10^{-3}$, and $\epsilon_0=8.20\times10^{-12}$. An Approach-2 fit to the budget frontiers gives
\begin{align}
L(N,D)=0.67800+\frac{46.677}{N^{0.26207}}+\frac{29.947}{D^{0.21364}},
\end{align}
which implies $D^*(C)\propto C^{0.55091}$ and fits the frontier with RMSE $0.000989$ BPB, against $\alpha=0.20$ and $D^*\propto C^{0.46}$ on FineWeb-Edu. The single $10^{22}$ run has no within-budget width control and enters the fit as an anchor only.

\begin{figure}[t]
\begin{center}
\includegraphics[width=\textwidth]{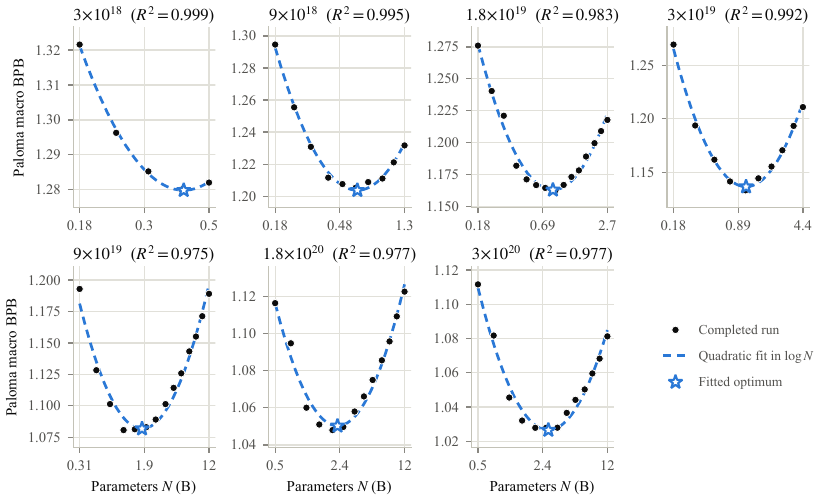}
\end{center}
\caption{The seven width-controlled DCLM IsoFLOP ladders. Points are completed runs, curves are per-budget quadratics in log model size, and stars mark their minima. Each panel reports the $R^2$ of its quadratic. Every budget has an interior minimum over model size. The single-run $10^{22}$ anchor is omitted.}
\label{fig:dclm-ladder-grid}
\end{figure}

\paragraph{DCLM rewrite sweep.}
The sweep contains $40$ runs at the two configurations of Section~\ref{sec:compute_eff_exp}, one per $(C,R)$ cell. Refitting the mixture-KL floor gives $\Rhat_{\infty}=0.078$, a $0.00577$-BPB gain over $R=0$, and RMSE $0.00162$ BPB.

\paragraph{Cross-corpus comparison.}
Matching DCLM runs to FineWeb-Edu $R=0$ runs by budget, width, and depth gives $77$ pairs over seven budgets. DCLM has lower Paloma BPB in all $77$ pairs. The pairs are not exact: the DCLM models have $1.02$--$1.16\times$ the parameters and $0.91$--$0.99\times$ the tokens, and the two corpora use different tuned reference points.

\clearpage
\section{Qualitative Rewrite Examples}
\label{qualitative_rewrites}

We selected the ten examples below to illustrate the four styles and the range of rewriter sizes, not to assess fidelity. Each pairs a source document with its FinePhrase rewrite and one bootstrapped-rewriter rewrite, spanning the FAQ, table, tutorial, and math styles and rewriters from $550$M to $3$B parameters. Several rewrites contain errors that the rewriter introduced: arithmetic mistakes in the math style, invented details in the FAQ and tutorial styles, and a contradiction of the source in one biology example. We do not measure aggregate fidelity, and the loss results in Sections~\ref{sec:compute_eff_exp} and~\ref{sec:data_eff_exp} do not depend on it.

\input{sections/qualitative_rewrite_examples}

%% file: sections/tables/hero_aggregate.tex
\begin{table}[t]
\caption{\textbf{The $10^{22}$-FLOP hero run nearly matches SmolLM2-Instruct on Paloma.} The Paloma column uses byte-pooled BPB at a $4096$-token context for both model families. The QA, Math, and Code columns retain their matched offline evaluations. The first four rows share the Paloma-optimal $(N,D)$, hence training compute $C \approx 10^{22}$ FLOPs, and differ only in $R$; their total compute $C_{\mathrm{total}} = 6ND + 4N_r RD$ varies with $R$. SmolLM2 is trained on $11$T tokens at ${\approx}1.1\times10^{23}$ FLOPs, $11\times$ the total budget of the hero run.}
\label{tab:hero_aggregate}
\begin{center}
\small\setlength{\tabcolsep}{4.5pt}
\begin{tabular}{r|c|ccc|cccc|c}
\toprule
\multirow{2}{*}{Optimized for} & \multirow{2}{*}{Model} & \multirow{2}{*}{$R$} & \multirow{2}{*}{$N$} & \multirow{2}{*}{$D$} & \multicolumn{4}{c}{BPB $\downarrow$} & Acc.\ $\uparrow$ \\
\cmidrule(lr){6-9}\cmidrule(lr){10-10}
 & & & & & Paloma & QA & Math & Code & QA \\
\midrule
-- & Natural-only & $0$ & 21.9B & 76B & 1.061 & 0.731 & 1.328 & 1.042 & 0.619 \\
Paloma & Hero & $0.32$ & 21.9B & 76B & 0.983 & 0.700 & 0.842 & 0.587 & \textbf{0.622} \\
-- & Grid & $0.5$ & 21.9B & 76B & 0.988 & 0.698 & 0.820 & 0.568 & 0.620 \\
-- & Grid & $1.0$ & 21.9B & 76B & 1.052 & 0.719 & 0.800 & 0.529 & 0.594 \\
\midrule
QA & Grid & $0.74$ & 13.8B & 120B & 0.988 & 0.696 & 0.801 & 0.520 & 0.620 \\
Code & Grid & $1.0$ & 9.7B & 172B & 1.107 & 0.710 & 0.788 & 0.507 & 0.597 \\
\midrule
-- & SmolLM2-1.7B & -- & 1.7B & 11T & \textbf{0.943} & \textbf{0.667} & 0.707 & \textbf{0.399} & 0.617 \\
-- & SmolLM2-1.7B-Instruct & -- & 1.7B & 11T & 0.972 & 0.709 & \textbf{0.697} & 0.426 & 0.606 \\
\bottomrule
\end{tabular}
\end{center}
\end{table}

%% file: sections/tables/hero_paloma_domains.tex
\begin{table}[t]
\caption{\textbf{The $R=0.321$ hero run minimizes Paloma macro BPB among the runs at $10^{22}$ training FLOPs.} The four $22$B runs share $(N,D)=(21.9\mathrm{B},76\mathrm{B})$; the best value among them per row is bold. The SmolLM2 columns are evaluated on the same Paloma validation split at a $4096$-token context. The $R=0.321$ run reaches a lower BPB than SmolLM2-1.7B-Instruct on $10$ of the $16$ sources.}
\label{tab:hero_individual}
\begin{center}
\small\setlength{\tabcolsep}{4pt}
\begin{tabular}{r|cccc|cc}
\toprule
\multirow{2}{*}{Domain} & \multicolumn{4}{c}{$10^{22}$ training FLOPs, $22$B} & \multicolumn{2}{c}{SmolLM2-1.7B} \\
\cmidrule(lr){2-5}\cmidrule(lr){6-7}
 & $R=0$ & $R=0.321$ & $R=0.5$ & $R=1.0$ & Base & Instruct \\
\midrule
4chan & 1.162 & \textbf{1.023} & 1.034 & 1.143 & 0.926 & 0.952 \\
C4 100 Domains & \textbf{0.790} & 0.793 & 0.797 & 0.844 & 0.772 & 0.797 \\
C4 English & \textbf{0.827} & 0.828 & 0.833 & 0.874 & 0.817 & 0.844 \\
Dolma v1.5 & 0.876 & \textbf{0.804} & 0.808 & 0.863 & 0.763 & 0.791 \\
Dolma 100 PLs & 1.106 & 0.694 & \textbf{0.680} & 0.698 & 0.486 & 0.511 \\
Dolma 100 Subreddits & 1.089 & \textbf{1.022} & 1.026 & 1.066 & 0.973 & 1.001 \\
Falcon RefinedWeb & 0.894 & \textbf{0.887} & 0.891 & 0.943 & 0.869 & 0.897 \\
Gab & 1.719 & \textbf{1.579} & 1.590 & 1.658 & 1.550 & 1.587 \\
M2D2 S2ORC & 0.766 & \textbf{0.760} & 0.763 & 0.823 & 0.737 & 0.764 \\
M2D2 Wikipedia & 0.726 & \textbf{0.716} & 0.723 & 0.789 & 0.728 & 0.758 \\
Manosphere & 1.112 & \textbf{1.057} & 1.063 & 1.123 & 1.000 & 1.030 \\
mC4 & 0.841 & \textbf{0.825} & 0.829 & 0.890 & 0.801 & 0.828 \\
Penn Treebank & 0.851 & \textbf{0.843} & 0.848 & 0.955 & 0.851 & 0.882 \\
RedPajama & 0.946 & \textbf{0.775} & 0.777 & 0.826 & 0.710 & 0.737 \\
TwitterAAE & 2.591 & \textbf{2.453} & 2.459 & 2.580 & 2.426 & 2.455 \\
WikiText-103 & 0.682 & \textbf{0.673} & 0.679 & 0.759 & 0.681 & 0.711 \\
\midrule
Macro average & 1.061 & \textbf{0.983} & 0.988 & 1.052 & 0.943 & 0.972 \\
\bottomrule
\end{tabular}
\end{center}
\end{table}

%% file: sections/tables/hero_skill_tasks.tex
\begin{table}[t]
\caption{\textbf{Task-level scores behind the QA, Math, and Code aggregates of Table~\ref{tab:hero_aggregate}.} Columns are the runs at $10^{22}$ training FLOPs with downstream evaluations; best value per row in bold. MMLU, Minerva MATH, and multilingual MBPP are macro averages over their subtasks (Table~\ref{tab:hero_subtasks}). SmolLM2 task-level scores are not archived.}
\label{tab:hero_tasks}
\begin{center}
\small\setlength{\tabcolsep}{3pt}
\begin{tabular}{l|cccc}
\toprule
\multirow{2}{*}{Task} & Natural-only & Grid & QA-optimal & Code-optimal \\
 & $R=0$, 21.9B & $R=0.5$, 21.9B & $R=0.74$, 13.8B & $R=1.0$, 9.7B \\
\midrule
\multicolumn{5}{l}{\emph{QA (BPB $\downarrow$)}} \\
\quad ARC-Easy & 0.540 & \textbf{0.490} & \textbf{0.490} & 0.502 \\
\quad ARC-Challenge & 0.736 & \textbf{0.672} & 0.676 & 0.686 \\
\quad HellaSwag & 0.704 & 0.704 & \textbf{0.703} & 0.718 \\
\quad MMLU (57 subjects) & 0.850 & 0.809 & \textbf{0.801} & 0.818 \\
\midrule
\multicolumn{5}{l}{\emph{QA (accuracy $\uparrow$)}} \\
\quad ARC-Easy & 0.806 & \textbf{0.818} & 0.809 & 0.798 \\
\quad ARC-Challenge & 0.533 & \textbf{0.566} & 0.561 & 0.516 \\
\quad HellaSwag & \textbf{0.738} & 0.725 & 0.733 & 0.702 \\
\quad MMLU (57 subjects) & \textbf{0.449} & 0.444 & 0.442 & 0.433 \\
\midrule
\multicolumn{5}{l}{\emph{Math (BPB $\downarrow$)}} \\
\quad GSM8K-Platinum & 0.825 & 0.450 & 0.446 & \textbf{0.432} \\
\quad Minerva MATH (7 subjects) & 1.191 & 0.541 & 0.508 & \textbf{0.499} \\
\quad TheoremQA & \textbf{2.794} & 3.140 & 3.201 & 3.162 \\
\midrule
\multicolumn{5}{l}{\emph{Code (BPB $\downarrow$)}} \\
\quad HumanEval (3-shot) & 0.822 & 0.503 & \textbf{0.446} & 0.446 \\
\quad MBPP (3-shot) & 1.130 & 0.679 & 0.635 & \textbf{0.619} \\
\quad Multilingual MBPP (17 languages) & 1.175 & 0.521 & 0.480 & \textbf{0.456} \\
\bottomrule
\end{tabular}
\end{center}
\end{table}

%% file: sections/tables/hero_subtasks.tex
\begin{table}[t]
\caption{\textbf{Minerva MATH subjects and multilingual MBPP languages} (BPB, lower is better) for the runs of Table~\ref{tab:hero_tasks}; best value per row in bold.}
\label{tab:hero_subtasks}
\begin{center}
\small\setlength{\tabcolsep}{5pt}
\begin{tabular}{l|cccc}
\toprule
\multirow{2}{*}{Subtask} & Natural-only & Grid & QA-optimal & Code-optimal \\
 & $R=0$, 21.9B & $R=0.5$, 21.9B & $R=0.74$, 13.8B & $R=1.0$, 9.7B \\
\midrule
\multicolumn{5}{l}{\emph{Minerva MATH}} \\
\quad Algebra & 1.215 & 0.557 & 0.524 & \textbf{0.515} \\
\quad Counting and probability & 1.078 & 0.520 & 0.496 & \textbf{0.492} \\
\quad Geometry & 1.132 & 0.551 & 0.517 & \textbf{0.509} \\
\quad Intermediate algebra & 1.324 & 0.589 & 0.544 & \textbf{0.530} \\
\quad Number theory & 1.101 & 0.604 & 0.574 & \textbf{0.566} \\
\quad Prealgebra & 1.039 & 0.494 & 0.472 & \textbf{0.469} \\
\quad Precalculus & 1.446 & 0.473 & 0.431 & \textbf{0.415} \\
\midrule
\multicolumn{5}{l}{\emph{Multilingual MBPP}} \\
\quad Bash & 1.422 & 0.675 & 0.589 & \textbf{0.564} \\
\quad C & 1.037 & 0.392 & 0.367 & \textbf{0.346} \\
\quad C++ & 1.011 & 0.385 & 0.353 & \textbf{0.336} \\
\quad C\# & 1.002 & 0.324 & 0.295 & \textbf{0.282} \\
\quad Go & 1.366 & 0.495 & 0.435 & \textbf{0.410} \\
\quad Haskell & 0.965 & 0.654 & 0.606 & \textbf{0.586} \\
\quad Java & 0.985 & 0.283 & 0.262 & \textbf{0.251} \\
\quad JavaScript & 1.170 & 0.469 & 0.440 & \textbf{0.415} \\
\quad MATLAB & 1.067 & 0.545 & 0.507 & \textbf{0.490} \\
\quad PHP & 1.143 & 0.439 & 0.419 & \textbf{0.391} \\
\quad Python & 1.130 & 0.679 & 0.635 & \textbf{0.619} \\
\quad R & 1.279 & 0.589 & 0.547 & \textbf{0.542} \\
\quad Ruby & 1.278 & 0.682 & 0.638 & \textbf{0.602} \\
\quad Rust & 1.458 & 0.650 & 0.580 & \textbf{0.521} \\
\quad Scala & 1.222 & 0.650 & 0.603 & \textbf{0.565} \\
\quad Swift & 1.256 & 0.479 & 0.446 & \textbf{0.420} \\
\quad TypeScript & 1.180 & 0.470 & 0.430 & \textbf{0.411} \\
\bottomrule
\end{tabular}
\end{center}
\end{table}

%% file: sections/qualitative_rewrite_examples.tex
\lstdefinestyle{qualitativerewrite}{
  basicstyle=\ttfamily\footnotesize,
  breaklines=true,
  breakatwhitespace=true,
  columns=fullflexible,
  keepspaces=true,
  frame=single,
  framerule=0.25pt,
  rulecolor=\color{black!25},
  backgroundcolor=\color{black!2},
  framexleftmargin=4pt,
  framexrightmargin=4pt,
  framesep=2pt,
  aboveskip=4pt,
  belowskip=8pt,
  showstringspaces=false
}

\begingroup
\let\qualitativesubsubsection\subsubsection
\let\qualitativeparagraph\paragraph
\renewcommand{\subsubsection}[1]{%
  \Needspace{8\baselineskip}%
  \qualitativesubsubsection{#1}%
}
\renewcommand{\paragraph}[1]{%
  \Needspace{6\baselineskip}%
  \qualitativeparagraph{#1}\mbox{}\par\nobreak%
}

\subsubsection{550M: FAQ}
\noindent\textbf{Checkpoint:} \nolinkurl{delphi-glm-550m-sft-llama3-lr1e-04-ep15}\\
\textbf{Document:} \nolinkurl{<urn:uuid:f719a404-47c8-4aab-bc52-a5ce965f66dc>}

\paragraph{Original.}
\begin{lstlisting}[style=qualitativerewrite]
What does NEPAD stand for?
The New Partnership for Africa's Development.
What are the origins of NEPAD?
The NEPAD strategic framework document was prepared by the leaders of the five initiating states (Algeria, Egypt, Nigeria, Senegal, South Africa), in response to a mandate given to them by the Summit of the Organisation of African Unity (OAU). The 37th Summit of the OAU in July 2001 formally adopted the strategic framework document.
What is NEPAD?
NEPAD is a VISION and STRATEGIC FRAMEWORK FOR AFRICA'S RENEWAL.
NEPAD is a programme of the African Union (AU), adopted in 2001 by African leaders, with the primary objectives of poverty eradication, promotion of sustainable growth and development, and the empowerment of women through building genuine partnerships at country, regional and global levels.
NEPAD is a blueprint for Africa's development in the 21st century and provides unique opportunities for Africa to address the critical challenges facing the continent, including the attainment of the Millennium Development Goals (MDGs) and other continental and international agreed upon goals.
Why is there a need for NEPAD?
NEPAD is designed to address the current challenges facing the African continent. Issues such as the escalating poverty levels and underdevelopment of African countries and the continued marginalisation of Africa from the global economy needed a radical new intervention, spearheaded by African leaders - that would bring forth a new Vision for Africa's Renewal.
What is the NEPAD programme of action?
It is a detailed action plan derived from the NEPAD Strategic Framework document and the NEPAD Initial Action Plan, adopted by the African Union Summit in Durban in June 2002. The NEPAD Programme of Action is a holistic, comprehensive and integrated sustainable development initiative for the revival of Africa.
What are the NEPAD primary objectives?
What are the principles of NEPAD?
What are the elements of NEPAD's strategic focus?
In sum, NEPAD calls for adapted policy reforms and increased investments in the areas of agriculture and food security, science and technology, environment, trade and market access, governance, infrastructure (energy, transport and water sanitation, and information and communication technologies), gender and capacity development.
How is the NEPAD programme governed?
The work of the NEPAD Planning and Coordinating Agency - which is tasked with the implementation of the NEPAD Programme - is overseen by the NEPAD Heads of State and Government Orientation Committee (HSGOC) and a Steering Committee.
The Chairperson of the African Union Commission exercises supervisory authority over the Agency.
Basically, the main features of the AU/NEPAD governance structure include the African Union Assembly of Heads of State and Government (AU Assembly); NEPAD Heads of State and Government Orientation Committee (HSGOC) and a Steering Committee.
What is the NEPAD Planning and Coordinating Agency (NEPAD Agency)?
The NEPAD Agency was established by the 14th African Union Summit decision as the institutional vehicle for implementing the African Union Development agenda. Designated as the technical body of the African Union, the core mandate of the NEPAD Agency is to facilitate and coordinate the implementation of regional and continental priority programmes and projects and to push for partnerships, resource mobilisation and research and knowledge management.
\end{lstlisting}

\paragraph{FinePhrase.}
\begin{lstlisting}[style=qualitativerewrite]
NEPAD stands for New Partnership for Africa's Development, a program of the African Union created by the leaders of five initiating states (Algeria, Egypt, Nigeria, Senegal, South Africa) in 1999 and officially adopted by the 37th Summit of the OAU in 2001. The NEPAD strategic framework document addresses critical issues including poverty, sustainable growth, and enhancing women's empowerment through building genuine partnerships at all levels. It is aimed at achieving Millennium Development Goals and global, continental, and international agreed upon goals.
\end{lstlisting}

\paragraph{550M rewrite.}
\begin{lstlisting}[style=qualitativerewrite]
# Frequently Asked Questions: NEPAD, VISION, and STRATEGY for African Development

## Foundational Concepts

**What is NEPAD?**
The New Partnership for Africa's Development (NEPAD) is a global partnership designed for African development. The name stands for New Partnership for Africa's Development, and it defines the rationale for NEPAD's interventions and initiatives.

**What are the origins of NEPAD?**
The strategic framework document was prepared by five initiating states (Algeria, Egypt, Nigeria, Senegal, South Africa), who responded to a mandate given to them by the Summit of the Organisation of African Unity (OAU). It was signed by leaders of the five founding states and was officially adopted by the Summit in July 2001.
\end{lstlisting}

\subsubsection{837M: Table}
\noindent\textbf{Checkpoint:} \nolinkurl{delphi-glm-837m-sft-llama3-lr2e-04-ep10}\\
\textbf{Document:} \nolinkurl{<urn:uuid:ccdd3db1-ea0b-4b2b-885f-f0b5243ec8b3>}

\paragraph{Original.}
\begin{lstlisting}[style=qualitativerewrite]
"Predator" bacteria (green) surround "prey" bacteria (red) in this petri dish version of the Serengeti. Rather than eating their prey, however, predator cells release a chemical that activates a suicide gene in the prey. Prey cells also release a chemical, but one that promotes survival of the predators. Researchers genetically programmed the cells to "communicate" with each other in this way and function as a synthetic ecosystem. The artificial system acts as an experimental model and can help us understand behaviors in more complex, natural ecosystems. July 9, 2008
Courtesy of Hao Song, Duke University.
Full Story: (http://publications.nigms.nih.gov/computinglife/predator.htm)
\end{lstlisting}

\paragraph{FinePhrase.}
\begin{lstlisting}[style=qualitativerewrite]
| Cell Type | Predator | Prey |
|-----------|----------|-------|
| Green     | Red          | No            |
| 1 Death   | 1 Survival | 0             |
| 1 Death   | 0 Deactivation| 1             |
| Deactivates| Deactivates| Deactivates |

Question: What are the potential advantages and disadvantages of creating an engineered synthetic ecosystem?

Answer: The potential advantages include gaining insights into complex behaviors in ecosystems, testing hypotheses, and developing more efficient methods for bioremediation and biotechnology. However, the disadvantages may comprise risks in unintended consequences of introducing alien life forms, unforeseen side effects on native ecosystems, and challenges in long-term sustainability and control of the engineered ecosystem.
\end{lstlisting}

\paragraph{837M rewrite.}
\begin{lstlisting}[style=qualitativerewrite]
| Aspect | Details |
|--------|---------|
| **Topic** | Predator bacteria (green) surrounded "prey" bacteria (red) in a petri dish |
| **Purpose** | Study behavior of predators in artificial ecosystems |
| **Technique** | Involves releasing a chemical that activates a suicide gene in prey |
| **Cell Design** | Cells communicate via suicidal gene expression |
| **Experimental Setup** | Petri dish with prey and predator cells |
| **Key Finding** | Predator cells release chemical that activates suicide gene |
| **Research Significance** | Helps understand natural ecosystem behaviors |
| **Application** | Understanding natural ecosystem behavior in complex environments |

**Question:** How does the "Predator" bacteria's suicide gene function contribute to its ecosystem-based behavior?

**Answer:** The suicide gene in question functions as a chemical that activates a suicide gene in the prey cells, but the chemical itself promotes survival of the predators rather than suicide. This dual role allows the bacteria to both eat prey cells and release a chemical that promotes their survival, promoting ecosystem sustainability.
\end{lstlisting}

\subsubsection{1B: Table}
\noindent\textbf{Checkpoint:} \nolinkurl{delphi-glm-sft-llama3-lr1e-04-ep15}\\
\textbf{Document:} \nolinkurl{<urn:uuid:48cb395a-8ea3-4ee8-a473-4e98f269603f>}

\paragraph{Original.}
\begin{lstlisting}[style=qualitativerewrite]
Definition from Wiktionary, the free dictionary
Alternative forms
- acousticks (obsolete)
Audio (US) (file)
See -ics regarding the treatment of such nouns as singular.
- The physical quality of a space for performing music.
- (physics) The science of sounds, teaching their nature, phenomena and laws.
- Acoustics, then, or the science of sound, is a very considerable branch of physics. - Sir John Herschel.
Usage notes
- The science was previously divided by some writers into diacoustics, which explains the properties of sounds coming directly from (sic! Webster) the ear; and catacoustica, which treats of reflected sounds or echoes. This division is now obsolete.
Related terms
quality of a space for doing music
physics: a science of sounds
- The translations below need to be checked and inserted above into the appropriate translation tables, removing any numbers. Numbers do not necessarily match those in definitions. See instructions at Help:How to check translations.
\end{lstlisting}

\paragraph{FinePhrase.}
\begin{lstlisting}[style=qualitativerewrite]
**Table:**

| **Subject** | **Description**                     |
| ----------- | ----------------------------------- |
| Audio         | Defined as the collection of sounds |
| that a man can hear | in the universe. In this sense    |
|  ----------- | ----------------------------------- |
| Physics      | The scientific study of motion    |
|  ----------- | ----------------------------------- |
| Sound        | The vibration which occurs when    |
| a sound wave happens | objects of sufficient mass    |
|  ----------- | ----------------------------------- |
| Reflected    | The echo produced when a sound    |
| wave bounces back     | from a reflective surface   |
|  ----------- | ----------------------------------- |
| Catacoustics | A branch of physics dealing     |
|  with reflected sounds or echoes    |

**Question and Answer Pair:**

## Question:
What branch of physics deals with the study of reflections or echoes in space, leading to the concept of acoustics?

## Answer:
Acoustics.
\end{lstlisting}

\paragraph{1B rewrite.}
\begin{lstlisting}[style=qualitativerewrite]
| Aspect | Details |
|:-------|:--------|
| Definition | The physical quality of a space for performing music |
| Alternative forms | Acoustics (obsolete) |
| Audio (US) | File |
| Definition | The science of sounds, teaching their nature, phenomena, and laws |
| Usage | They were previously divided by some writers into diacoustics; explains properties of sounds coming from the ear |
| Related terms | Quality of a space for doing music |
| Definition | The science of sounds, teaching their nature, phenomena, and laws |
| Usage | Were previously divided by some writers into diacoustics; explains division now obsolete |
| Related terms | Quality of a space for doing music |
| Definition | The science of sounds, teaching their nature, phenomena, and laws |
| Usage | Were formerly divided by some writers into diacoustics; explains division now obsolete |
| Related terms | Quality of a space for doing music |

**Question:** How does the usage note for "sound" differ from how it was historically defined?

**Answer:** The usage note states that "sound" was previously divided into diacoustics, explains that this division is now obsolete, and references "quality of a space for doing music" as a related term.
\end{lstlisting}

\subsubsection{1B: FAQ}
\noindent\textbf{Checkpoint:} \nolinkurl{delphi-glm-sft-llama3-lr1e-04-ep15}\\
\textbf{Document:} \nolinkurl{<urn:uuid:4fd3037a-b829-4827-8fa6-c3aeb3e57890>}

\paragraph{Original.}
\begin{lstlisting}[style=qualitativerewrite]
Pain Management and Bioelectric Therapy
- Introduction to bioelectric therapy
- What conditions are treated with bioelectric therapy?
- How effective is bioelectric therapy for pain relief?
- What happens during bioelectric therapy?
- What are the side effects of bioelectric therapy?
- How often should people get bioelectric therapy?
- How do people prepare for bioelectric therapy?
Introduction to Bioelectric Therapy
Bioelectric therapy is a safe, drug-free treatment option for people in pain. It is used to treat some chronic pain and acute pain conditions. It relieves pain by blocking pain messages to the brain. When you are injured, pain receptors send a message to the central nervous system (the brain and spinal cord). The message is registered as pain by certain cells in the body. Using bioelectric currents, bioelectric therapy relieves pain by interrupting pain signals before they reach the brain. Bioelectric therapy also prompts the body to produce endorphins which help to relieve pain.
What Conditions Are Treated With Bioelectric Therapy?
Bioelectric therapy can be used to treat chronic and acute pain conditions including:
- Complex regional pain syndrome, also known as reflex sympathetic dystrophy or RSD.
- Back pain.
- Muscle pain.
- Headaches and migraines.
- Disorders of blood flow in the upper and lower limbs.
- Temporomandibular joint (TMJ) syndrome (which affects the jaw).
- Disorders of the nervous system, such as diabetic neuropathy.
- Pain and ulcers of the skin resulting from poor circulation or scleroderma (a chronic condition that can cause thickening or hardening of the skin).
Bioelectric therapy isn't right for everyone. It is not recommended for people who:
- Have a pacemaker.
- Are pregnant.
- Have thrombosis (blood clots in the arms or legs).
- Have a bacterial infection.
Chronic Pain/Back Pain
Find tips and advances in treatment.
\end{lstlisting}

\paragraph{FinePhrase.}
\begin{lstlisting}[style=qualitativerewrite]
Bioelectric therapy is a non-invasive, drug-free treatment that can help manage chronic pain and back pain, reduce inflammation, and speed up the healing process.
\end{lstlisting}

\paragraph{1B rewrite.}
\begin{lstlisting}[style=qualitativerewrite]
# Frequently Asked Questions: Pain Management and Bioelectric Therapy

## Understanding Bioelectric Therapy

**What is bioelectric therapy?**
Bioelectric therapy is a safe, drug-free treatment option for people in pain. It is used to treat some chronic pain and acute pain conditions. It relieves pain by blocking pain messages to the brain.

**What conditions does bioelectric therapy treat?**
Bioelectric therapy treats some chronic pain conditions and acute pain conditions. It relieves pain by blocking pain messages to the brain.

**How effective is bioelectric therapy for pain relief?**
Bioelectric therapy is highly effective for pain relief. It works by interrupting pain signals before they reach the brain, rather than stimulating the brain to produce morphine or endorphins themselves.

**What happens during bioelectric therapy?**
During bioelectric therapy, electrodes are placed on the skin to deliver electrical current. This current then passes through the body, interrupting pain signals that would otherwise reach the brain. The affected area is then exposed to the therapeutic bioelectric currents.

## Side Effects and Precautions

**What are the side effects of bioelectric therapy?**
While most bioelectric therapies are safe and effective, some side effects can occur. These can include skin reactions, infections, and potential injury to the brain or nervous system.

**How often should people get bioelectric therapy?**
People who receive bioelectric therapy should take it as directed to ensure proper pain relief.

## Treatment Duration

**When should people get bioelectric therapy?**
People who receive bioelectric therapy should take it as directed because it interrupts pain signals before they reach the brain. The treatment also prompts the body to produce endorphins, which help to relieve pain.

## Special Considerations

**Are bioelectric therapy and pain management exactly the same thing?**
No. While they share many treatments, bioelectric therapy and pain management are distinct. Bioelectric therapy uses electrical currents to block pain messages, while pain management uses medication to manage the pain.

**Who should not take bioelectric therapy?**
People who have a pacemaker should not take bioelectric therapy. People who are pregnant should also avoid bioelectric therapy. These peoples' bodies may not react to the electrical current as it passes through them.

## Additional Resources

**Where can I find more information about bioelectric therapy?**
Additional information can be found in the resources for pain management and bioelectric therapy. Topics include history, types of pain conditions treated, effective treatment methods, side effects, and safety concerns.
\end{lstlisting}

\subsubsection{3B: Table}
\noindent\textbf{Checkpoint:} \nolinkurl{delphi-glm-3b-sft-llama3-lr5e-05-ep15}\\
\textbf{Document:} \nolinkurl{<urn:uuid:3e09eef3-39c3-41eb-a404-bb93536daa7f>}

\paragraph{Original.}
\begin{lstlisting}[style=qualitativerewrite]
When Your Child Has Pharyngitis or Tonsillitis
|Pain or redness can occur in these areas.|
What Causes Pharyngitis or Tonsillitis?
Most commonly, pharyngitis and tonsillitis are caused by a viral or bacterial infection.
What Are the Symptoms of Pharyngitis or Tonsillitis?
The main symptom of both conditions is a sore throat. Your child may also have a fever, redness or swelling of the throat, and trouble swallowing.
How Is Pharyngitis or Tonsillitis Diagnosed?
The healthcare provider will examine your child's throat. The healthcare provider might swab (wipe) your child's throat. This swab will be tested for the bacteria that causes an infection called strep throat. If needed, a blood test can be done to check for a viral infection, such as mononucleosis.
How Is Pharyngitis or Tonsillitis Treated?
If your child's sore throat is caused by a bacterial infection, the healthcare provider may prescribe antibiotics. Otherwise, you can treat your child's sore throat at home. To do this:
Give your child acetaminophen or ibuprofen to ease the pain. Don't give your child aspirin to relieve a fever. Using aspirin to treat a fever in children could cause a serious condition called Reye's syndrome.
Give your child cool liquids to drink.
Have your child gargle with warm saltwater if it helps relieve pain. An over-the-counter throat numbing spray may also help.
What Are the Long-Term Concerns?
If your child has frequent sore throats, take him or her to see a healthcare provider. Removing the tonsils may help relieve your child's recurring problems.
Call your child's healthcare provider right away if your child has any of the following:
\end{lstlisting}

\paragraph{FinePhrase.}
\begin{lstlisting}[style=qualitativerewrite]
|Heading | Description |
|---------|-------------|
|When Your Child Has Pharyngitis or Tonsillitis|
|What to Know About Pharyngitis or Tonsillitis|
|Causes of Pharyngitis or Tonsillitis|
|Symptoms of Pharyngitis or Tonsillitis|
|Diagnosis Process for Pharyngitis or Tonsillitis|
|How to Treat Pharyngitis or Tonsillitis|
|Long-Term Concerns|

What Are the Long-Term Concerns for Pharyngitis or Tonsillitis?
\end{lstlisting}

\paragraph{3B rewrite.}
\begin{lstlisting}[style=qualitativerewrite]
| Category | Details |
|----------|---------|
| **Condition** | Pharyngitis/tonsillitis |
| **Cause** | Viral or bacterial infection |
| **Symptoms** | Sore throat, fever, redness/swelling, difficulty swallowing |
| **Key Symptom** | Sore throat |
| **Alternative Symptoms** | Fever, redness/swelling, swallowing problems |
| **Diagnosis** | Medical provider examines throat for strep throat; blood test for viral infection (mononucleosis); or tonsil removal if needed |
| **Treatment** | Antibiotics for bacterial infections; acetaminophen or ibuprofen for pain; cool liquids for sore throat; gargling with saltwater for swallowing |
| **Self-Care Steps** | Remove tonsils if sore throat is caused by bacteria; avoid aspirin for fever; use throat numbing spray if needed; avoid throat drying |
| **Long-Term Concerns** | Frequent sore throats; removal of tonsils for recurring problems |

**Question:** Why is removing the tonsils recommended for children with recurring pharyngitis or tonsillitis?

**Answer:** Removing the tonsils helps relieve children's recurring problems because the tonsils serve as a reservoir for bacteria, and removing them creates a clear path for healing.
\end{lstlisting}

\subsubsection{3B: Tutorial}
\noindent\textbf{Checkpoint:} \nolinkurl{delphi-glm-3b-sft-llama3-lr5e-05-ep15}\\
\textbf{Document:} \nolinkurl{<urn:uuid:66166475-112f-43f3-a5c1-215ae0e2a518>}

\paragraph{Original.}
\begin{lstlisting}[style=qualitativerewrite]
Now more than five decades old, Herbert Hoover's 1959 statement to the Board of Trustees of Stanford University on the purpose and scope of the Hoover Institution continues to guide and define its mission in the twenty-first century:
"This Institution supports the Constitution of the United States, its Bill of Rights and its method of representative government. Both our social and economic systems are based on private enterprise from which springs initiative and ingenuity.... Ours is a system where the Federal Government should undertake no governmental, social or economic action, except where local government, or the people, cannot undertake it for themselves.... The overall mission of this Institution is, from its records, to recall the voice of experience against the making of war, and by the study of these records and their publication, to recall man's endeavors to make and preserve peace, and to sustain for America the safeguards of the American way of life. This Institution is not, and must not be, a mere library. But with these purposes as its goal, the Institution itself must constantly and dynamically point the road to peace, to personal freedom, and to the safeguards of the American system."
The principles of individual, economic, and political freedom; private enterprise; and representative government were fundamental to the vision of the Institution's founder. By collecting knowledge, generating ideas, and disseminating both, the Institution seeks to secure and safeguard peace, improve the human condition, and limit government intrusion into the lives of individuals.
\end{lstlisting}

\paragraph{FinePhrase.}
\begin{lstlisting}[style=qualitativerewrite]
I. Introduction
- A brief overview of the importance of the Hoover Institution
- The mission and purpose of the institution
II. Numbered Steps to Understand the Institution's Mission
III. Understanding Individual Freedom
   - What it means to be a responsible citizen
IV. The Role of Government in Society
   - Providing public goods
   - Encouraging innovation
V. Education and Its Importance
   - The value of knowledge in personal and societal development
VI. A Life of Liberty
   - Fostering creativity
   - Promoting self-reliance
VII. Conclusion
- Recap of the institution's principles
- Importance of individual freedom, economic freedom, political freedom, and preserving personal freedom
VIII. Further Resources
- The official website of the Hoover Institution
- Other sources of information on the institution
\end{lstlisting}

\paragraph{3B rewrite.}
\begin{lstlisting}[style=qualitativerewrite]
# Understanding Herbert Hoover's 1959 Statement on the Purpose and Scope of the Hoover Institution: A Tutorial

## Background and Context

- **Date:** December 5, 1959
- **Historical context:** Established just over five decades earlier
- **Key phrase:** "...support the Constitution of the United States, its Bill of Rights, and method of representative government"
- **Historical context:** These principles guided the Institute's mission through the twentieth century.

## Step 1: Review the Institution's Founding Principles

The Hoover Institution operates under the following principles:

1. **Individual freedom:** Ensures individual rights and freedoms are protected
2. **Economic freedom:** Supports private enterprise through free inquiry and knowledge generation
3. **Political freedom:** Supports representative government through the ballot box and public debate
4. **Respect for law:** Guarantees legal authority and institutional integrity
5. **Sustaining American Way of Life:** Emphasizes local self-determination and sustainable practices

## Step 2: Understand the Institution's Mission

The mission of the Hoover Institution is:

- **Recollect the voice of experience against the making of war**
- **Recollect man's efforts to make and preserve peace**
- **Sustain for America the safeguards of the American way of life**

## Step 3: Recognize the Institute's Goal

The mission statement sets the foundation for action:

- Preventing war
- Improving human conditions
- Limiting government intrusion into individual lives

## Step 4: Understand the Role of the Board

The Board of Trustees continues to define the Institution's purpose:

- **Focus:** Today, through leadership and expertise
- **Mission:** Keep the Institution's mission clear through consistent consideration and action

## Step 5: Access the Institute's Records

To support the mission, the Board of Trustees maintains:

- **Scope:** The records of the Hoover Institution
- **Scope:** They collect knowledge, generate ideas, and disseminate both for educational and scholarly purposes
- **Purpose:** Secure and safeguard peace, improve the human condition, and limit government intrusion into individual lives

## Step 6: Apply These Principles in Daily Life

- **Individual freedom:** Protect the rights and freedoms you enjoy
- **Economic freedom:** Support private enterprise through free enterprise
- **Political freedom:** Support representative government through the ballot box and public debate
- **Respect for law:** Ensure legal authority and institutional integrity
- **Sustaining the American way of life:** Emphasize local self-determination and sustainable practices

## Additional Resources

- Source: Herbert Hoover's 1959 Statement on the Purpose and Scope of the Hoover Institution: A Tutorial
- Source: Washington, D.C. (December 5, 1959) - Herbert Hoover's 1959 Annual Report, Stanford University
- Source: http://www.hoover.org/archive/statement/ (accessed December 5, 2019)
\end{lstlisting}

\subsubsection{1B: Math}
\noindent\textbf{Checkpoint:} \nolinkurl{delphi-glm-sft-llama3-lr1e-04-ep15}\\
\textbf{Document:} \nolinkurl{<urn:uuid:63c9a260-1d93-4db5-bbb6-9126d821ed64>}

\paragraph{Original.}
\begin{lstlisting}[style=qualitativerewrite]
ZomeTools - Zome Systems
The ZomeTools - Zome System
, is a fantastic Geometry Education system. Made of precision molded ABS plastic, the ZomeTools Geometry kits are versatile manipulative tools for problem solving and hands-on exploration of math, geometry and science concepts for first graders through grad students. Color and shape coded parts integrate the concepts of shape, number and vector into one coherent geometry system. The result is the best educational geometry construction system on the market. Zome System / ZomeTools components are designed to incorporate the Fibonacci sequence and the Golden Proportion - the very same proportions that govern the growth of cells, crystals and other natural forms.
connector node provides 62 buildable directions in space. This allows thousands of geometrical Zome structures
that can't be built in any other system.
With the help of Zometools you can continue to explore: Why so many everyday items like balls, pizzas, and can lids are round. The hows and whys of the development of our number system by the ancient Greeks. Why geometric shapes show up so often in art, science and board games. What properties attributed to the spiral make it the most widespread shape in nature, from embryos and hair curls to hurricanes and galaxies. How the human body shares the design of a corn plant and the solar system. How a snowflake is like Stonehenge, and a beehive like a calendar. How our ten fingers hold the secrets of both a lobster and a cathedral. Zometools provide an introduction to the geometric code of nature and the universe.
\end{lstlisting}

\paragraph{FinePhrase.}
\begin{lstlisting}[style=qualitativerewrite]
Problem:

Jenna found 2 ZomeTools sets from ZomeTools at the store. She bought the total amount of sets so she bought 10 sets. Then, she traded 5 sets with Daniel for StarLego. After that, she spent $20 to buy more ZomeTools sets. If Jenna has 2 ZomeTools sets left, how much did each ZomeTools set cost her after purchasing from ZomeTools and trading?
\end{lstlisting}

\paragraph{1B rewrite.}
\begin{lstlisting}[style=qualitativerewrite]
**Problem:**

A teacher is building a geometric ZomeTools kit for her first-grade class. The teacher has 62 buildable directions in space, and each direction contains 373 built-in directions. She wants to provide her students with access to as many buildable directions as possible, noting that 1 direction contains 5 built-in directions. If she allocates 4 sections of space per direction, how many total built-in directions will the class have access to?

**Solution:**

Step 1: Identify the total number of built-in directions.
Total directions = 62

Step 2: Calculate the number of built-in directions available.
62 x 5 = 252 built-in directions

Step 3: Determine how many of these directions can be used.
252 x 5 = 258 directions

Step 4: Calculate the remaining number of directions to be used.
258 - 258 = 48 directions

Step 5: Determine how many students can access as many built-in directions as possible.
48 x 5 = 144 students

Answer: The class will have 258 built-in directions available, and 144 students will be able to access as many of these directions as possible.
\end{lstlisting}

\subsubsection{3B: Math}
\noindent\textbf{Checkpoint:} \nolinkurl{delphi-glm-3b-sft-llama3-lr5e-05-ep15}\\
\textbf{Document:} \nolinkurl{<urn:uuid:d29bb74f-45d0-452c-b599-d774fedf25bd>}

\paragraph{Original.}
\begin{lstlisting}[style=qualitativerewrite]
Colored map compiled by merging geologic and topographic maps of the United States through computer processing and enhancement with links to panoramic movie using QuickTime, descriptions of features, and a game entitled Puzzle of Regions.
The USGS provides map-on-demand (MOD) printing for selected infrequently requested geologic and other theme map products providing high-quality paper-copy maps as needed. Site links to available map products with thumbnail views and how to order.
This publication (Open-File Report 95-479) is a teacher's guide to two U.S. Geological Survey maps that show gravity and magnetic data in contour form superimposed on a LANDSAT satellite image of the San Francisco Bay area.
Techniques of computer-based representations of areal geology to construct three-dimensional geologic maps that retain all the information of traditional geologic maps while providing a more detailed picture of the subsurface.
Main page for accessing links for information and data on the San Francisco Bay estuary and its watershed with links to highlights, water, biology, wetlands, hazards, digital maps, geologic mapping, winds, bathymetry and overview of the Bay.
\end{lstlisting}

\paragraph{FinePhrase.}
\begin{lstlisting}[style=qualitativerewrite]
What is the value of dividing 50 by 2 x 3 and then multiplying this result by 15?
\end{lstlisting}

\paragraph{3B rewrite.}
\begin{lstlisting}[style=qualitativerewrite]
**Problem:**

A geological map shows geologic formations and topographic features of the United States. The map shows 24 different types of geologic formations and 5 different topographic zones. Additionally, the map names 3 major mountain systems and 4 major drainage systems. If the map's publisher requires a teacher's guide to 2 U.S. Geological Survey maps that show gravity and magnetic data in contour form superimposed on a LANDSAT satellite image of the San Francisco Bay area, and the 2nd map costs $24 per copy, how many total copies of the teacher's guide must the publisher order to sell enough copies to cover the total cost of the 2 maps?

**Solution:**

Step 1: Identify the number of different geologic formations on the 24 types of formations map.
Total formations = 24

Step 2: Identify the number of different topographic zones on the 5 different topographic zones map.
Topographic zones = 5

Step 3: Calculate the total number of formations and topographic zones on the 24 types of formations map.
24 types x 3 formations per type = 72 formations and topographic zones

Step 4: Calculate the total number of copies of the 24 types of formations map needed.
72 formations x 2 copies per formation = 240 copies

Step 5: Calculate the total cost of the 2 maps.
240 copies x $24 per copy = $720

The publisher must order 720 copies of the teacher's guide to sell enough copies to cover the total cost.
\end{lstlisting}

\subsubsection{1B: Tutorial}
\noindent\textbf{Checkpoint:} \nolinkurl{delphi-glm-sft-llama3-lr1e-04-ep15}\\
\textbf{Document:} \nolinkurl{<urn:uuid:66166475-112f-43f3-a5c1-215ae0e2a518>}

\paragraph{Original.}
\begin{lstlisting}[style=qualitativerewrite]
Now more than five decades old, Herbert Hoover's 1959 statement to the Board of Trustees of Stanford University on the purpose and scope of the Hoover Institution continues to guide and define its mission in the twenty-first century:
"This Institution supports the Constitution of the United States, its Bill of Rights and its method of representative government. Both our social and economic systems are based on private enterprise from which springs initiative and ingenuity.... Ours is a system where the Federal Government should undertake no governmental, social or economic action, except where local government, or the people, cannot undertake it for themselves.... The overall mission of this Institution is, from its records, to recall the voice of experience against the making of war, and by the study of these records and their publication, to recall man's endeavors to make and preserve peace, and to sustain for America the safeguards of the American way of life. This Institution is not, and must not be, a mere library. But with these purposes as its goal, the Institution itself must constantly and dynamically point the road to peace, to personal freedom, and to the safeguards of the American system."
The principles of individual, economic, and political freedom; private enterprise; and representative government were fundamental to the vision of the Institution's founder. By collecting knowledge, generating ideas, and disseminating both, the Institution seeks to secure and safeguard peace, improve the human condition, and limit government intrusion into the lives of individuals.
\end{lstlisting}

\paragraph{FinePhrase.}
\begin{lstlisting}[style=qualitativerewrite]
I. Introduction
- A brief overview of the importance of the Hoover Institution
- The mission and purpose of the institution
II. Numbered Steps to Understand the Institution's Mission
III. Understanding Individual Freedom
   - What it means to be a responsible citizen
IV. The Role of Government in Society
   - Providing public goods
   - Encouraging innovation
V. Education and Its Importance
   - The value of knowledge in personal and societal development
VI. A Life of Liberty
   - Fostering creativity
   - Promoting self-reliance
VII. Conclusion
- Recap of the institution's principles
- Importance of individual freedom, economic freedom, political freedom, and preserving personal freedom
VIII. Further Resources
- The official website of the Hoover Institution
- Other sources of information on the institution
\end{lstlisting}

\paragraph{1B rewrite.}
\begin{lstlisting}[style=qualitativerewrite]
# Understanding the Herbert Hoover Institution: A Guide to Its Legacy and Core Principles

## Background and Context

**Step 1: Identify the Institution's Historical Significance**
- **Established:** 1959
- **Original Purpose:** Purposeed for 5 decades to preserve the Institution's role in American democracy
- **Core Purpose:** Support for its Constitution, Bill of Rights, and method of representative government
- **Historical Context:** Founded in 1959 by President Herbert Hoover, this institution continues to serve America today

## Step 2: Understand the Mission Statement**
- **Core Purpose:** Recall the voice of experience against the making of war
- **Mission Statement:** Avoid creating unnecessary governmental, social, or economic actions
- **Approach:** Study the records to understand efforts to make and preserve peace, and by study, to illuminate the bounds of American way of life

## Step 3: Recognize the Historical Significance of Individual Liberty**
- **Core Principle:** The core mission of this Institution is to recall the voice of experience against the making of war
- **Approach:** Avoid creating unnecessary governmental, social, or economic actions
- **Historical Context:** Enable voluntary, collaborative engagement with peace
- **Emergency Response:** Acknowledge that the Institution cannot operate effectively without proper community support

## Step 4: Understand the Role of the Individual in American Society**
- **Core Principle:** The individual must play a meaningful role in shaping his or her own society
- **Approach:** Foster personal freedom and the safeguards of the American system as primary goals
- **Historical Context:** Remember that government intrusion into personal decisions is not acceptable

## Step 5: Apply Core Principles to Contemporary Challenges**
- **Purpose:** Restore the Institute's role in American democracy by:
  - Recreating the voices of experience against the making of war
  - Creating peace through research, education, and public engagement
  - Limiting government intrusion into individual lives
  - Strengthening the American way of life through individual freedom and limited government

## Key Takeaway
The Herbert Hoover Institution's legacy goes beyond preserving literature. By creating a space for collective knowledge, learning, and collaborative engagement, it seeks to secure and safeguard peace, improve the human condition, and limit government intrusion into individual lives.
\end{lstlisting}

\subsubsection{3B: FAQ}
\noindent\textbf{Checkpoint:} \nolinkurl{delphi-glm-3b-sft-llama3-lr5e-05-ep15}\\
\textbf{Document:} \nolinkurl{<urn:uuid:4fd3037a-b829-4827-8fa6-c3aeb3e57890>}

\paragraph{Original.}
\begin{lstlisting}[style=qualitativerewrite]
Pain Management and Bioelectric Therapy
- Introduction to bioelectric therapy
- What conditions are treated with bioelectric therapy?
- How effective is bioelectric therapy for pain relief?
- What happens during bioelectric therapy?
- What are the side effects of bioelectric therapy?
- How often should people get bioelectric therapy?
- How do people prepare for bioelectric therapy?
Introduction to Bioelectric Therapy
Bioelectric therapy is a safe, drug-free treatment option for people in pain. It is used to treat some chronic pain and acute pain conditions. It relieves pain by blocking pain messages to the brain. When you are injured, pain receptors send a message to the central nervous system (the brain and spinal cord). The message is registered as pain by certain cells in the body. Using bioelectric currents, bioelectric therapy relieves pain by interrupting pain signals before they reach the brain. Bioelectric therapy also prompts the body to produce endorphins which help to relieve pain.
What Conditions Are Treated With Bioelectric Therapy?
Bioelectric therapy can be used to treat chronic and acute pain conditions including:
- Complex regional pain syndrome, also known as reflex sympathetic dystrophy or RSD.
- Back pain.
- Muscle pain.
- Headaches and migraines.
- Disorders of blood flow in the upper and lower limbs.
- Temporomandibular joint (TMJ) syndrome (which affects the jaw).
- Disorders of the nervous system, such as diabetic neuropathy.
- Pain and ulcers of the skin resulting from poor circulation or scleroderma (a chronic condition that can cause thickening or hardening of the skin).
Bioelectric therapy isn't right for everyone. It is not recommended for people who:
- Have a pacemaker.
- Are pregnant.
- Have thrombosis (blood clots in the arms or legs).
- Have a bacterial infection.
Chronic Pain/Back Pain
Find tips and advances in treatment.
\end{lstlisting}

\paragraph{FinePhrase.}
\begin{lstlisting}[style=qualitativerewrite]
Bioelectric therapy is a non-invasive, drug-free treatment that can help manage chronic pain and back pain, reduce inflammation, and speed up the healing process.
\end{lstlisting}

\paragraph{3B rewrite.}
\begin{lstlisting}[style=qualitativerewrite]
# Frequently Asked Questions: Pain Management and Bioelectric Therapy

## Foundations

**What is bioelectric therapy?**
Bioelectric therapy is a drug-free treatment option for people in pain. It uses electrical currents to interrupt pain messages before they reach the brain, which helps relieve pain.

**What conditions are treated with bioelectric therapy?**
Bioelectric therapy is used to treat certain chronic pain and acute pain conditions, including:
- Complex regional pain syndrome
- Back pain
- Muscle pain
- Headaches and migraines
- Disorders of blood flow in the upper and lower limbs
- Temporomandibular joint (TMJ) syndrome
- Disorders of the nervous system, such as diabetic neuropathy

**How does bioelectric therapy work?**
Bioelectric therapy relieves pain by blocking pain messages to the brain. When you are injured, pain receptors send pain messages to the central nervous system (the brain and spinal cord). The message is registered as pain by certain cells in the body. The cells interpret this pain as threatening and trigger the protective response of pain relief mechanisms.

## How Bioelectric Therapy Works

**How does bioelectric therapy relieve pain?**
Bioelectric therapy prompts the body to produce endorphins. These are compounds that help relieve pain. Additionally, the therapy prompts the brain to produce more of its own endorphins, which helps to further relieve the pain.

**What happens during bioelectric therapy?**
Bioelectric therapy is a non-invasive procedure that uses electrical currents to block pain messages before they reach the brain. The repeated stimulation of specific pain-relieving nerves causes the body produce its own endorphins, which reduce pain.

**How often should people get bioelectric therapy?**
People should receive bioelectric therapy regularly - either as an outpatient treatment or in certain cases, as part of a rehabilitation program.

## Preparation and Limitations

**Is bioelectric therapy right for everyone?**
No. Bioelectric therapy is not recommended for people who have a pacemaker, are pregnant, have thrombosis (blood clots in the legs or arms), or have a bacterial infection.

**What conditions don't respond well to bioelectric therapy?**
Bioelectric therapy is not recommended for conditions like chronic pain, back pain, muscle pain, or headaches and migraines that don't respond to other treatments.

## Additional Information

**Where can I find more information about bioelectric therapy?**
Additional information is available through the company that supplies the treatment equipment, or through medical websites that cover pain management and bioelectric therapy.
\end{lstlisting}

\endgroup